%% file: main.tex
\documentclass{article}
\usepackage{jfrExamplee}

\usepackage{apalike}
\usepackage{setspace}
\usepackage{makecell}
\usepackage{hhline}
\usepackage[usenames, dvipsnames]{color}
\usepackage[cmex10]{amsmath}
\usepackage{eufrak}
\usepackage{multirow}
\usepackage{fixltx2e}
\usepackage{pbox}
\usepackage{textcomp}
\usepackage[caption=false,font=footnotesize]{subfig}
\usepackage[pdftex]{graphicx}
\usepackage[hidelinks]{hyperref}
\graphicspath{{Images/}}
\DeclareGraphicsExtensions{.pdf,.jpeg,.png,.jpg}
\usepackage{cite}
\usepackage{amssymb}

\def\figurename{Fig. }

\title{RTAB-Map as an Open-Source Lidar and Visual SLAM Library for Large-Scale and Long-Term Online Operation} 

\author{
Mathieu Labb\'e\\
Interdisciplinary Institute of Technological Innovation (3IT)\\
Department of Electrical Engineering and Computer Engineering\\
Universit\'{e} de Sherbrooke\\
Sherbrooke, Qu\'{e}bec, Canada\\
\texttt{Mathieu.M.Labbe@USherbrooke.ca} \\
\And
Fran{\c c}ois Michaud\\
Interdisciplinary Institute of Technological Innovation (3IT)\\
Department of Electrical Engineering and Computer Engineering\\
Universit\'{e} de Sherbrooke\\
Sherbrooke, Qu\'{e}bec, Canada\\
\texttt{Francois.Michaud@USherbrooke.ca} \\
}

\begin{document}

\hbox to0pt{\vbox{\baselineskip=10dd\hrule\hbox
to\hsize{\vrule\kern3pt\vbox{\kern3pt
\hbox{This is a preprint version of an article accepted in Journal of Field Robotics.}
\hbox{The final authenticated version will be available online at: \url{https://doi.org/10.1002/rob.21831}}
\kern3pt}\hfil\kern3pt\vrule}\hrule}%
\hss}

\maketitle

\begin{abstract}
\input{0_abstract.tex}
\end{abstract}

\section{Introduction}
\input{1_Introduction.tex}

\section{Popular SLAM Approaches Available on ROS}
\label{sec:related_work}
\input{2_related_work.tex}

\section{RTAB-Map Description}
\label{sec:system_description}
\input{3_system_description}

\input{4_results}

\section{Discussion}
\label{sec:discussion}

\input{5_discussion.tex}

\section{Conclusion}
\label{sec:conclusion}
\input{6_conclusion.tex}

\subsubsection*{Acknowledgments}
This work was supported by the Natural Sciences and Engineering Research Council of Canada (NSERC).

\bibliographystyle{apalike}
\bibliography{main} 

\end{document}

%% file: 0_abstract.tex
Distributed as an open source library since 2013, RTAB-Map started as an appearance-based loop closure detection approach with memory management to deal with large-scale and long-term online operation. 
It then grew to implement Simultaneous Localization and Mapping (SLAM) on various robots and mobile platforms. 
As each application brings its own set of contraints on sensors, processing capabilities and locomotion, it raises the question of which SLAM approach is the most appropriate to use in terms of cost, accuracy, computation power and ease of integration.
Since most of SLAM approaches are either visual or lidar-based, comparison is difficult. 
Therefore, we decided to extend RTAB-Map to support both visual and lidar SLAM, providing in one package a tool allowing users to implement and compare a variety of 3D and 2D solutions for a wide range of applications with different robots and sensors.
This paper presents this extended version of RTAB-Map and its use in comparing, both quantitatively and qualitatively, a large selection of popular real-world datasets (e.g., KITTI, EuRoC, TUM RGB-D, MIT Stata Center on PR2 robot), outlining strengths and limitations of visual and lidar SLAM configurations from a practical perspective for autonomous navigation applications.

%% file: 1_Introduction.tex
RTAB-Map, for Real-Time Appearance-Based Mapping\footnote{\url{http://introlab.github.io/rtabmap}} \cite{labbe13appearance,labbe2017}, is our open source library implementing loop closure detection with a memory management approach, limiting the size of the map so that loop closure detections are always processed under a fixed time limit, thus satisfying online requirements for long-term and large-scale environment mapping. 
Initiated in 2009 and released as an open source library in 2013, RTAB-Map has since be extended to a complete graph-based SLAM approach \cite{stachniss2016} to be used in various setups and applications \cite{laniel-letourneau-labbe-grondin-polgar-michaud2017,forestiemotive,chen2015kejia,iros14challengekinect}. 
As a result, RTAB-Map has evolved into a cross-platform standalone C++ library and a ROS package\footnote{\url{http://wiki.ros.org/rtabmap\_ros}}, driven by practical requirements such as: 

\begin{itemize}

\item Online processing: output of the SLAM module should be bounded to a maximum delay after receiving sensor data. 
For graph-based SLAM in particular, as the map grows, more processing time is required to detect loop closures, to optimize the graph and to assemble the map. 
Also, integration with other processing modules for control, navigation, obstacle avoidance, user interaction, object recognition, etc. may also limit the CPU time available for SLAM. 
Having the possibility to limit computation load is therefore beneficial to avoid lagging problems with other modules, and may even be necessary to prevent unsafe situations. 

\item Robust and low-drift odometry: while loop closure detection can correct most of the odometry drift, in real-world scenarios the robot often cannot properly localize itself on the map, either because it is exploring new areas or that there is a lack of discriminative features in the environment. 
During that time, odometry drift should be minimized so that accurate autonomous navigation is still possible until localization can occur, to avoid incorrectly overwriting mapped areas (e.g., incorrectly adding obstacles in the entrance of a room, making it a closed area for instance).
Estimating odometry with exterioceptive sensors such as cameras and lidars can be very accurate when there are enough features in the environment, but only using one sensing modality can be problematic and prone to localization failures if their tracked features in the environment are no longer visible. 
Using a mix of proprioceptive (e.g., wheel encoders, inertial measurement units (IMU)) and exterioceptive sensors would increase robustness to odometry estimation.

\item Robust localization: the SLAM approach must be able to recognize when it is revisiting past locations (for loop closure detection) to correct the map. 
Dynamic environments, illumination changes, geometry changes or even repetitive environments can lead to incorrect localization or failure to localize, and therefore the approach should be robust to false positives.

\item Practical map generation and exploitation: most popular navigation approaches are based on occupancy grid, and therefore it is beneficial to develop SLAM approaches that can provide 3D or 2D occupancy grid out-of-the-box for easy integration. 
Also, when the environment is mostly static, it is more practical to do a mapping session and then switch to localization, setting memory usage and saving map management time. 

\item Multi-session mapping (a.k.a. \textit{kidnapped robot problem} or \textit{initial state problem}): when turned on, a robot does not know its relative position to a previously created map, making it impossible to plan a path to a previously visited location.
To avoid having the robot restart the mapping process to zero or localize itself in a previously-built map before initiating mapping, multi-session mapping allows the SLAM approach to initialize a new map with its own referential on startup, and when a previously visited location is encountered, a transformation between the two maps can be computed. 
This brings the advantages of avoiding remapping the whole environment when only a small part should be remapped or a new area should be added.

\end{itemize}

With the diversity of available SLAM approaches, determining which one to use in relation to a specific platform and application is a difficult task, mostly because of the absence of comparative analyses between them. 
SLAM approaches are generally visual-based \cite{fuentes2015visual} or lidar-based only \cite{thrun2002robotic}, and are benchmarked often on datasets having only a camera or a lidar, but not both, making difficult to have a meaningful comparison between them.
It is even more the case when their implementation is either unavailable, only run offline or the required input formats on the robot platform are missing. The Robotic Operating System (ROS) \cite{quigley2009ros}, introduced in 2008, contributes greatly to standardize sensor data format, thus improving interoperability between robot platforms and making it possible to compare SLAM approaches. 
But still, visual SLAM approaches integrated in ROS are not often tested on autonomous robots: only SLAM by teleoperation or by a human moving the sensor \cite{murORB2,engel2015large,dai2017bundlefusion}. 
This avoids proper \textit{tf} (Transform Library) \cite{foote13_tf} handling to transform the outputs according to the robot base frame to satisfy ROS coordinate frame convention\footnote{http://www.ros.org/reps/rep-0105.html}.
It also avoids the need to have map outputs (e.g., 2D or 3D occupancy grid) compatible for the navigation algorithm to plan a path and avoid obstacles. 
Furthermore, some of the practical requirements outlined above are not always all addressed by the SLAM approaches, thus limiting comparison.

Therefore, since RTAB-Map evolved to handle these practical requirements, we decided to further extend RTAB-Map capabilities to compare visual and lidar SLAM configurations for autonomous robot navigation. 
RTAB-Map being a loop-closure approach with memory management as its core, it is independent of the odometry approach used, meaning that it can be fed with visual odometry, lidar odometry or even just wheel odometry. 
This means that RTAB-Map can be used to implement either a visual SLAM approach, a lidar SLAM approach or a mix of both, which makes it possible to compare different sensor configurations on a real robot. 
This paper describes the extended version of the RTAB-Map library and demonstrates its use to compare state-of-the-art visual and lidar SLAM approaches, and consequently outlining practical limitations between the two paradigms for autonomous navigation. 

The paper is organized as follows. 
Section \ref{sec:related_work} presents a brief overview of popular SLAM approaches currently available, compatible with ROS and that can be used on a robot for comparative evaluations. 
Section \ref{sec:system_description} presents the main components of the extended version of RTAB-Map. 
Section \ref{sec:slamperf} uses RTAB-Map to compare its visual and lidar SLAM configurations in terms of trajectory performance using standard offline and online datasets: the KITTI dataset for outdoor stereo and 3D lidar mapping by autonomous cars; the TUM RGB-D dataset for hand-held RGB-D mapping; the EuRoC dataset for stereo mapping on a drone; and the MIT Stata Center dataset comparing indoor stereo, RGB-D and 2D lidar SLAM configurations on a PR2 robot. 
Section \ref{sec:computationPerf} assesses map quality and computation performance variations according to the sensors used, and shows the effect of memory management for online mapping. 
Finally, Section \ref{sec:discussion} presents, based on the observed results, guidelines derived through the use of RTAB-Map regarding the choice of sensors for autonomous robot SLAM applications.

%% file: 2_related_work.tex
There are a great variety of open-source SLAM approaches available through ROS.
In this section, we review the most popular ones to outline their characteristics and to situate what RTAB-Map covers in terms of inputs and outputs to handle comparative studies of SLAM approaches.

Let us start with the following lidar approaches:
\begin{itemize}
\item GMapping \cite{grisetti2007improved} and TinySLAM \cite{steux2010tinyslam} are two approaches that use a particle filer to estimate the robot trajectory. 
As long as there are enough estimated particles and the real position error corresponds to the covariance of the input odometry, the particle filter converges to a solution which represents well the environment, particularly for GMapping when there are loop closures. 
GMapping, being ROS' default SLAM approach, has been widely used to derive a 2D occupancy grid map of the environment from 2D laser scans. 
Once the map is created, it can be used with Adaptive Monte Carlo Localization \cite{fox1999monte} for localization and autonomous navigation. 

\item Hector SLAM \cite{KohlbrecherHectorSLAM2011} can create fast 2D occupancy grid maps from a 2D lidar with low computation resources. 
It has proven to generate very low-drift localization while mapping in real-world autonomous navigation scenarios, like those in RoboCup Rescue Robot League competition \cite{kohlbrecher2016robocup}. 
It can also use external sensors like an IMU to estimate the robot position in 3D. 
However, Hector SLAM is not exactly a full SLAM approach as it does not detect loop closures, and thus the map cannot be corrected when visiting back a previous localization. 
Hector SLAM does not need external odometry, which can be an advantage when the robot does not have one, but can be a disadvantage when operating in an environment without a lot of geometry constraints, limiting laser scan matching performance. 

\item ETHZASL-ICP-Mapper\footnote{\url{http://wiki.ros.org/ethzasl_icp_mapper}}, based on \textit{libpointmatcher} library \cite{Pomerleau12comp}, can be used to create 2D occupancy grid maps from 2D lidar and an assembled point cloud from 2D or 3D lidars. 
But similarly to Hector SLAM, the approach does not detect loop closures, thus map errors over time cannot be corrected. 

\item Karto SLAM \cite{vincent2010comparison}, Lago SLAM\footnote{\url{https://github.com/rrg-polito/rrg-polito-ros-pkg}} \cite{carlone2012linear} and Google Cartographer \cite{hess2016real} are lidar graph-based SLAM approaches.  
They can generate 2D occupancy grid from their graph representation. 
Google Cartographer can be also used as backpack mapping platform as it supports 3D lidars, thus providing a 3D point cloud output. 
While mapping, they create sub-maps that are linked by constraints in the graph. 
When a loop closure is detected, the position of the sub-maps are re-optimized to correct errors introduced by noise of the sensor and scan matching accuracy. 
Unlike Hector SLAM, external odometry can be provided to get more robust scan matching in environments with low geometry complexity. 

\item BLAM\footnote{\url{https://github.com/erik-nelson/blam}} is a lidar graph-based SLAM that only supports 3D lidar for 3D point cloud generation of the environment. 
From the online documentation (which is the only documentation available), loop closures seem detected locally by scan matching when the robot visits previous locations, to then optimize the map using GTSAM \cite{dellaert2012factor}. 
This means that BLAM is not able to close large loops, for which local scan matching would not be able to appropriately register. 

\item SegMatch \cite{dube2016segmatch} is a 3D lidar-based loop closure detection approach that can be also used as 3D lidar graph-based SLAM. 
Loop closures are detected by matching 3D segments (e.g., parts of vehicles, buildings or trees) created from laser point clouds. 

\end{itemize}

In these lidar-based SLAM approaches, only SegMatch can be used for multi-session or multi-robot mapping \cite{dube2017online}.

Regarding visual SLAM, many open-source approaches exist but not many can be easily used on a robot (consult \cite{zollhofer2018state} for a review on 3D reconstruction focused approaches). For navigation, to avoid dealing with scale ambiguities, we limit our review to approaches able to estimate the real scale of the environment while mapping (e.g., with stereo and RGB-D cameras or with visual-inertial odometry), thus excluding structure from motion or monocular SLAM approaches like PTAM \cite{klein07parallel}, SVO \cite{Forster2014ICRA}, REMODE \cite{Pizzoli2014ICRA}, DT-SLAM \cite{herrera2014dt}, LSD-SLAM \cite{engel2014lsd} or ORB-SLAM \cite{mur2015orb}. The following visual SLAM approaches do not suffer from this scale drift over time.

\begin{itemize}
\item maplab \cite{schneider2018maplab} and VINS-Mono \cite{yi2017vinsmono} have recently been released as visual-inertial graph-based SLAM systems. Using only an IMU and a camera, they can provide visual maps for localization. maplab workflow is done in two steps: the data is recorded during an open loop phase using only visual-inertial odometry; then map management (i.e., loop closure detection, graph optimization, multi-session, dense map reconstruction) is done offline. The resulting visual map can be then used in localization mode afterward. In contrast, VINS-Mono's map management process is done online. For navigation, a local TSDF volume map computed on GPU can be provided for obstacle avoidance and path planning. To keep processing time bounded for large-scale environments, VINS-Mono limits the size of the graph, removing nodes without loop closures first, then removing others depending on the density of the graph. 

\item ORB-SLAM2 \cite{murORB2} and S-PTAM \cite{pire2017sptam} are currently two of the best state-of-the-art feature-based visual SLAM approaches that can be used with a stereo camera. 
More recently, ProSLAM \cite{schlegel2017proslam} has been released (only benchmark tools available at this time) to provide a comprehensive open source package using well know visual SLAM techniques.
For ORB-SLAM2, it can be also used with a RGB-D camera. 
They are all graph-based SLAM approaches. For ORB-SLAM2 and S-PTAM, when a loop closure is detected using DBoW2 \cite{GalvezTRO12}, the map is optimized using bundle adjustment. Graph optimization after loop closure is done in a separate thread to avoid influencing camera tracking frame rate performance.  For ProSLAM, loop closures are detected by direct comparison of the descriptors in the map, instead of using a bag-of-words approach.
For all these approaches, loop closure detection and graph optimization processing time increases as the map grows, which can make loop closure correction happening with a significant delay after being detected. 
The approaches maintain a sparse feature map. 
Without occupancy grid or dense point cloud outputs available out-of-the-box like lidar approaches, they can be then difficult to use on a real robot. 

\item DVO-SLAM \cite{kerl2013dense}, RGBiD-SLAM \cite{gutierrez2016dense} and MPR \cite{della2017general}, instead of using local visual features to estimate motion, use photometric and depth errors over all pixels of the RGB-D images. 
They can generate dense point clouds of the environment. MPR can be also used with a lidar but it is only an odometry approach. DVO-SLAM lacks of a loop closure detection approach independent of the pose estimate, which makes it less suitable for large-scale mapping. 

\item ElasticFusion \cite{whelan2016elasticfusion}, Kintinuous \cite{whelan2015real}, BundleFusion \cite{dai2017bundlefusion} and InfiniTAM \cite{InfiniTAM_ECCV_2016} are based on truncated signed distance field (TSDF) volume for RGB-D cameras. 
They can reconstruct online very appealing surfel-based maps, but a powerful computer with a recent Nvidia GPU is required. 
For ElasticFusion, while being able to process camera frames in real-time for small environments, processing time per frame increases according to the number of surfels in the map. 
For BundleFusion, global dense optimization time on loop closure detection increases according to the size of the environment. 
InfiniTAM seems faster to close loops, though processing time for loop closure detection and correction still increases with the size of the environment. 
While being open-source, these algorithms do not support ROS because they rely on extremely fast and tight coupling between the mapping and tracking on the GPU. 

\end{itemize}

All these previous visual SLAM approaches assume that the camera is never obstructed or that images always have enough visual features to track. 
Such assumptions cannot be satisfied practically on an autonomous robot where the camera can be fully obstructed from people passing by or when the robot is facing a surface without visual features (e.g., white wall) during navigation. 
The following visual SLAM approaches are designed to be more robust to these events:

\begin{itemize}
\item MCPTAM \cite{harmat2015multi} uses multiple cameras to increase the field of view of the system. 
If visual features can be perceived through at least one camera, MCPTAM is able to track the position. 

\item
RGBDSLAMv2 \cite{endres20143} can use external odometry as motion estimation. 
ROS packages like \textit{robot\_localization} \cite{MooreStouchKeneralizedEkf2014} can be then used to do sensor fusion (with an extended Kalman filter) of multiple odometry sources for a more robust odometry. 
RGBDSLAMv2 can generate a 3D occupancy grid (OctoMap\cite{hornung2013octomap}) and a dense point cloud of the environment. 
\end{itemize}

\begin{table*}[!t]
\caption{Popular ROS-compatible lidar and visual SLAM approaches with their supported inputs and online outputs.}
\label{slams}
\centering
\begin{tabular}{l||cccc|cc|c||c|cc|c}
\cline{2-12}
&  \multicolumn{7}{c||}{Inputs} &  \multicolumn{4}{c}{Online Outputs} \\
& \multicolumn{4}{c|}{Camera} & \multicolumn{2}{c|}{Lidar} & Odom  & Pose & \multicolumn{2}{c|}{Occupancy} & Point\\
& Stereo & RGB-D & Multi & IMU &  2D & 3D & & & 2D & 3D & Cloud  \\
\hline
GMapping 		& & 	& 	& 	& \checkmark	& 	& \checkmark	&  \checkmark 	& \checkmark 	& 	& \\
TinySLAM 		& & 	& 	& 	& \checkmark	& 	& \checkmark	&  \checkmark 	& \checkmark 	& 	& \\
Hector SLAM 		& & 	& 	& 	& \checkmark 	&  	&	& \checkmark 	& \checkmark 	& 	&  \\
ETHZASL-ICP 		& & 	& 	& 	& \checkmark 	& \checkmark 	& \checkmark	& \checkmark 	& \checkmark 	& 	& Dense\\
Karto SLAM	 	& & 	& 	& 	& \checkmark 	&  	& \checkmark	& \checkmark 	& \checkmark 	& 	& \\
Lago SLAM	 	& & 	& 	& 	& \checkmark 	&  	& \checkmark	& \checkmark 	& \checkmark 	& 	& \\
Cartographer 		& & 	& 	& 	& \checkmark 	& \checkmark  	& \checkmark	& \checkmark 	& \checkmark 	&  	& Dense\\
BLAM 			& & 	& 	& 	& 	& \checkmark 	&	& \checkmark 	& 	& 	& Dense\\
SegMatch 		& & 	& 	& 	& 	& \checkmark 	&	&  	& 	& 	& Dense\\
VINS-Mono 		&	&	&	&  \checkmark 	&	&	&	& \checkmark	& 	&  &  \\
ORB-SLAM2 		& \checkmark 	& \checkmark 	&  	& & 	& 	&	& 	& 	& 	& \\
S-PTAM 			& \checkmark 	& 	& 	& & 	& 	&	& \checkmark 	& 	& 	& Sparse\\
DVO-SLAM  		& 	& \checkmark 	& 	& & 	& 	&	& \checkmark 	& 	& 	& \\
RGBiD-SLAM 		& 	& \checkmark 	& 	& & 	& 	&	& 	& 	& 	& \\
MCPTAM 			& \checkmark 	& 	& \checkmark 	& &	& 	&	& \checkmark 	& 	& 	& Sparse \\
RGBDSLAMv2 		& 	& \checkmark 	& 	& & 	& 	& \checkmark	& \checkmark 	& 	&  \checkmark	& Dense \\
RTAB-Map 		& \checkmark 	& \checkmark 	& \checkmark 	&  & \checkmark	& \checkmark 	& \checkmark	& \checkmark 	& \checkmark	& \checkmark 	& Dense\\
\hline
\end{tabular}
\end{table*}

Table \ref{slams} provides a summary of the open-source ROS-compatible SLAM approaches in relation to their inputs and outputs. 
The Lidar 3D category includes all point cloud types, including those derived from depth images of a RGB-D camera. 
Odom refers to odometry input that can be used to help the SLAM approach compute motion estimation.
3D occupancy grid map refers to OctoMap \cite{hornung2013octomap}. 
Note that ORB-SLAM2 and RGBiD-SLAM do not have any online outputs: they do have a visualizer to see the pose and point cloud, but they do not provide them as ROS topics to other modules out-of-the-box. VINS-Mono does provide current point cloud of odometry but not the map and the TSDF map output is not available through the current project page.
The last entry in Table \ref{slams} situates what inputs can be used and outputs that are provided in the extended version of RTAB-Map presented in this paper.
Beside RTAB-Map and RGBDSLAMv2, no visual SLAM approaches provide out-of-the-box occupancy grid outputs required for autonomous navigation.
RGBDSLAMv2 \cite{endres20143} is probably the visual SLAM approach sharing the most similarities with RTAB-Map, since both can use external odometry as motion estimation. While they do not combine IMU with camera, they can still use visual-intertial odometry approach with their external odometry input.
They can also generate a 3D occupancy grid (OctoMap\cite{hornung2013octomap}) and a dense point cloud for depending modules. 
However, RTAB-Map can also provide 2D occupancy grid like lidar-based SLAM approaches.

%% file: 3_system_description.tex
\begin{figure*}[!t] 
\centering 
\includegraphics[width=\textwidth]{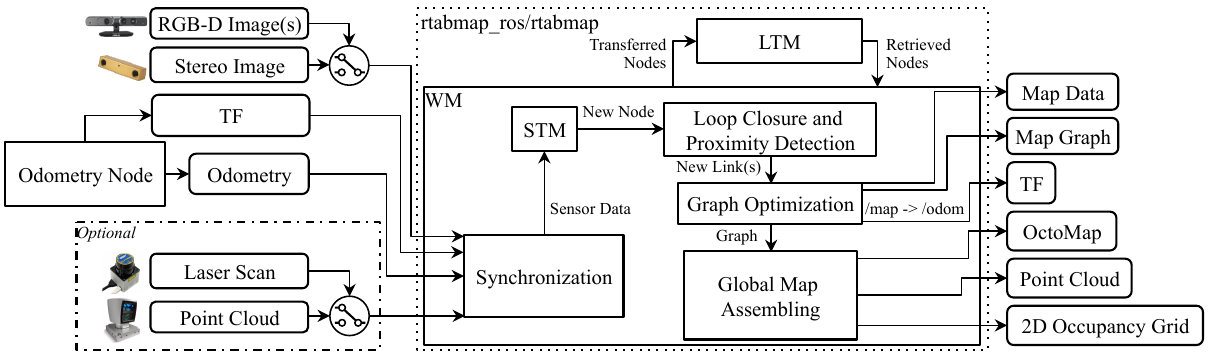} 
\caption{The required inputs are: TF to define the position of the sensors in relation to the base of the robot; Odometry from any source (which can be 3DoF or 6DoF); one of the camera inputs (one or multiple RGB-D images, or a stereo image) with corresponding calibration messages. Optional inputs are either a laser scan from a 2D lidar or a point cloud from a 3D lidar. All  messages from these inputs are then synchronized and passed to the graph-SLAM algorithm. The outputs are: Map Data containing the latest added node with compressed sensor data and the graph; Map Graph without any data; odometry correction published on TF; an optional OctoMap (3D occupancy grid); an optional dense Point Cloud; an optional 2D Occupancy Grid.}
\label{fig:rtabmap} 
\end{figure*}

RTAB-Map is a graph-based SLAM approach that has been integrated in ROS as the \textit{rtabmap\_ros}\footnote{\url{http://wiki.ros.org/rtabmap_ros}} package since 2013. 
Figure \ref{fig:rtabmap} shows its main ROS node called \textit{rtabmap}. 
The odometry is an external input to RTAB-Map, which means that SLAM can also be done using any kind of odometry to use what is appropriate for a given application and robot. 
The structure of the map is a graph with nodes and links. 
After sensor synchronization, the Short-Term Memory (STM) module creates a node memorizing the odometry pose, sensor's raw data and additional information useful for next modules (e.g., visual words for Loop Closure and Proximity Detection, and local occupancy grid for Global Map Assembling).
Nodes are created at a fixed rate ``Rtabmap/DetectionRate" set in milliseconds according to how much data created from nodes should overlap each other. 
For example, if the robot is moving fast and sensor range is small, the detection rate should be increased to make sure that data of successive nodes overlap, but setting it too high would unnecessary increase memory usage and computation time. A link contains a rigid transformation between two nodes. 
There are three kind of links: Neighbor, Loop Closure and Proximity links. 
Neighbor links are added in the STM between consecutive nodes with odometry transformation. 
Loop Closure and Proximity links are added through loop closure detection or proximity detection, respectively. 
All the links are used as constraints for graph optimization.
When there is a new loop closure or proximity link added to the graph, graph optimization propagates the computed error to the whole graph, to decrease odometry drift. 
With the graph optimized, OctoMap, Point Cloud and 2D Occupancy Grid outputs can be assembled and published to external modules. 
Odometry correction to derive the robot localization in map frame is also available through \textit{tf} \cite{foote13_tf} /map\textrightarrow /odom. 

RTAB-Map' memory management approach \cite{labbe13appearance} runs on top of graph management modules.
It is used to limit the size of the graph so that long-term online SLAM can be achieved in large environments. 
Without memory management, as the graph grows, processing time for modules like Loop Closure and Proximity Detection, Graph Optimization and Global Map Assembling can eventually exceed  real-time constraints, i.e., processing time can become greater than the node acquisition cycle time. 
Basically, RTAB-Map's memory is divided into a Working Memory (WM) and a Long-Term Memory (LTM). 
When a node is transferred to LTM, it is not available anymore for modules inside the WM. 
When RTAB-Map's update time exceeds the fixed time threshold ``Rtabmap/TimeThr", some nodes in WM are transferred to LTM to limit the size of the WM and decrease the update time. 
Similarly to the fixed time threshold, there is also a memory threshold ``Rtabmap/MemoryThr" that can be used to set the maximum number of nodes that WM can hold. 
To determine which nodes to transfer to LTM, a weighting mechanism identifies locations that are more important than others, using heuristics such as the longer a location has been observed, the more important it is and therefore should be left in the WM. To do so, when creating a new node, STM initializes the node's weight to 0 and compares it visually (deriving a percentage of corresponding visual words) with the last node in the graph. 
If they are similar (with the percentage of corresponding visual words over the similarity threshold ``Mem/RehearsalSimilarity"), the weight of the new node is increased by one plus the weight of the last node.
The weight of the last node is reset to 0, and the last node is discarded if the robot is not moving to avoid increasing uselessly the graph size.
When the time or the memory thresholds are reached, the oldest of the smallest weighted nodes are transferred to LTM first. 
When a loop closure happens with a location in the WM, neighbor nodes of this location can be brought back from LTM to WM for more loop closure and proximity detections. 
As the robot is moving in a previously visited area, it can then remember the past locations incrementally to extend the current assembled map and localize using past locations \cite{labbe2017}. 

The next sections explain in more details RTAB-Map's pipeline, starting from Odometry Node to Global Map Assembling.
Definition of key parameters to configure and use RTAB-Map are provided.

\subsection{Odometry Node}
\label{sec:odometry}

Odometry Node can implement any kind of odometry approaches from simpler ones derived from wheel encoders and IMU to more complex ones using camera and lidar. Independently of the sensor used, it should provide to RTAB-Map at least the pose of the robot estimated so far in form of an Odometry message\footnote{\url{http://docs.ros.org/api/nav_msgs/html/msg/Odometry.html}} with the corresponding \textit{tf}'s transform (e.g., /odom \textrightarrow /base\_link). When proprioceptive odometry is not already available on the robot or when it is not accurate enough, visual or lidar-based odometry must be used.
For visual odometry, RTAB-Map implements two standard odometry approaches \cite{scaramuzza2011visual} called Frame-To-Map (F2M) and Frame-To-Frame (F2F). 
The main difference between these approaches is that F2F registers the new frame against the last keyframe, and F2M registers the new frame against a local map of features created from past keyframes. 
These two approaches are also implemented for lidars and are referred to as Scan-To-Map (S2M) and Scan-To-Scan (S2S), following the same idea than F2M and F2F but using point clouds instead of 3D visual features. The following sections show how Odometry Node is implemented when one of these visual or lidar odometry approaches is chosen.

\subsubsection{Visual Odometry}
\label{sec:vo}

\begin{figure}[!t] 
\centering 
\includegraphics[width=5in]{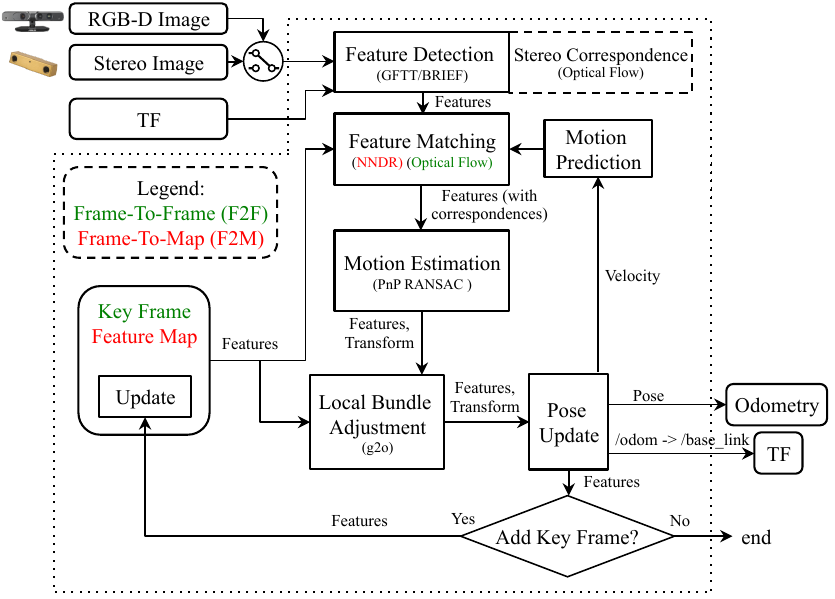} 
\caption{Block diagram of \textit{rgbd\_odometry} and \textit{stereo\_odometry} ROS nodes. TF defines the position of the camera in relation to the base of the robot and as output to publish the odometry transform of the base of the robot. The pipeline is the same for a RGB-D camera or a stereo camera, except that stereo correspondences are computed for the later to determine the depth of the detected features. Two odometry approaches can be used: a Frame-To-Frame (F2F) approach in green, and a Frame-To-Map (F2M) approach in red.} 
\label{fig:visual_odometry} 
\end{figure}

Figure \ref{fig:visual_odometry} presents RTAB-Map's visual odometry using two colors to differentiate between F2F (green) and F2M (red). 
It can use RGB-D or stereo cameras as inputs. 
\textit{tf} is required to know where the camera is placed on the robot so that output odometry can be transformed into the robot base frame (e.g., /base\_link). 
If the camera is on the robot's head and the head turns, it does not influence the odometry of the robot base as long as \textit{tf} between the robot's body and the robot's head is also updated. 
The process works as follows.

\begin{itemize}

\item Feature Detection: When a frame is captured, GoodFeaturesToTrack \cite{shi1994good} (GFTT) features are detected with a maximum number fixed by ``Vis/MaxFeatures" parameter. 
RTAB-Map supports all feature types available in OpenCV\footnote{\url{https://opencv.org}}, but GFTT has been chosen to ease parameter tuning and get uniformly detected features across different image size and light intensity. 
For stereo images, stereo correspondences are computed by optical flow using the iterative Lucas–Kanade method \cite{lucas1981iterative}, to derive disparity per feature between left and right images. 
For RGB-D images, the depth image is used as a mask for GFTT to avoid extracting features with invalid depth. 

\item Feature Matching: For F2M, matching is done by nearest neighbor search \cite{muja_flann_2009} with nearest neighbor distance ratio (NNDR) test \cite{Lowe04}, using BRIEF descriptors \cite{calonder2010brief} of the extracted features against those in the Feature Map. 
The Feature Map contains 3D features with descriptors from last key frames. 
NNDR is defined by parameter ``Vis/CorNNDR". 
For F2F, optical flow is done directly on GFTT features without having to extract descriptors, providing faster feature correspondences against the Key Frame.

\item Motion Prediction: A motion model is used to predict where the features of the Key Frame (F2F) or the Feature Map (F2M) should be in the current frame, based on the previous motion transformation. 
This limits the search window for Feature Matching to provide better matches, particularly in environments with dynamic objects and repetitive textures. 
The search window radius is defined by parameter ``Vis/CorGuessWinSize", and a constant velocity motion model is used. 

\item Motion Estimation: When correspondences are computed, the Perspective-n-Point (PnP) RANSAC implementation of OpenCV \cite{bradski2008learning} is used to compute the transformation of the current frame accordingly to features in Key Frame (F2F) or Feature Map (F2M). A minimum of inliers ``Vis/MinInliers" is required to accept the transformation.

\item Local Bundle Adjustment: The resulting transformation is refined using local bundle adjustment \cite{kummerle11g2o} on features of all key frames in the Feature Map (F2M) or only those of the last Key Frame (F2F). 

\item Pose Update: With the estimated transformation, the output odometry is then updated as well as \textit{tf}'s /odom \textrightarrow /base\_link transform. 
Covariance is computed using the median absolute deviation (MAD) approach \cite{Rusu_ICRA2011_PCL} between 3D feature correspondences.

\item Key Frame and Feature Map update: If the number of inliers computed during Motion Estimation is below the fixed threshold ``Odom/KeyFrameThr", the Key Frame or Feature Map is updated. 
For F2F, the Key Frame is simply replaced by the current frame. 
For F2M, the Feature Map is updated by adding the unmatched features of the new frame and updating the position of matched features that were refined by the Local Bundle Adjustment module. 
Feature Map has a fixed maximum of features kept temporary (consequently a maximum of Key-Frames). 
When the size of Feature Map is over the fixed threshold ``OdomF2M/MaxSize", the oldest features not matched with the current frame are removed. 
If a key frame does not have features in Feature Map anymore, it is discarded. 

\end{itemize}

If for some reasons the current motion of the camera is very different than the predicted one, a valid transformation may not be found (after Motion Estimation or Local Bundle Adjustment boxes), thus features are matched again but without motion prediction. 
For F2M, features in the current frame are compared to all features in Feature Map, then another transformation is computed. 
For F2F, to be more robust to invalid correspondences, feature matching with NNDR is done instead of optical flow, and thus BRIEF descriptors have to be extracted. 
If transformation still cannot be computed, odometry is considered lost and the next frame is compared without motion prediction. 
The output odometry pose is set to null with very high variance (i.e., 9999). 
Modules subscribing to this visual odometry node can then know when odometry cannot be computed. 

Note that as odometry in RTAB-Map is independent of the mapping process, other visual odometry approaches have been integrated in RTAB-Map for convenience and ease of comparison between them. 
The approaches chosen are open-source or provide an application programming interface (API) and can be used as odometry-only. 
Complete visual SLAM approaches in which it is difficult to split the odometry from the mapping processes cannot be integrated because RTAB-Map take care of the mapping process. 
Seven approaches have been integrated in RTAB-Map: FOVIS \cite{huang2011visual}, Viso2 \cite{Geiger2011IV}, DVO \cite{kerl2013dense}, OKVIS \cite{leutenegger2015keyframe}, ORB-SLAM2 \cite{murORB2}, MSCKF \cite{sun2018robust} and Google Project Tango. 
FOVIS, Viso2, DVO, OKVIS and MSCKF are visual or visual-inertial odometry-only approaches, which make them straightforward to integrate by connecting their odometry output to RTAB-Map. 
ORB-SLAM2 is a full SLAM approach, thus to integrate in RTAB-Map, loop closure detection inside ORB-SLAM2 is disabled. 
Local bundle adjustment of ORB-SLAM2 is still working, which makes the modified module similar to F2M. 
The big difference is the kind of features extracted (ORB \cite{rublee2011orb}) and how they are matched together (direct descriptor comparison instead of NNDR). 
Similarly to F2M, the size of the feature map is limited so that constant time visual odometry can be achieved (without limiting the feature map size, ORB-SLAM2 computation time increases over time). 
As ORB-SLAM2 has not been designed (at least in the code available at the time of writing this paper) to remove or forget features in its map, memory is not freed when features are removed, which results in an increasing RAM usage over time (a.k.a. memory leak). 
To integrate Google Project Tango in RTAB-Map library, \textit{area learning} feature is disabled and its visual inertial odometry is used directly.

\subsubsection{Lidar Odometry}
\label{sec:icp_odometry}

\begin{figure}[!t] 
\centering 
\includegraphics[width=5in]{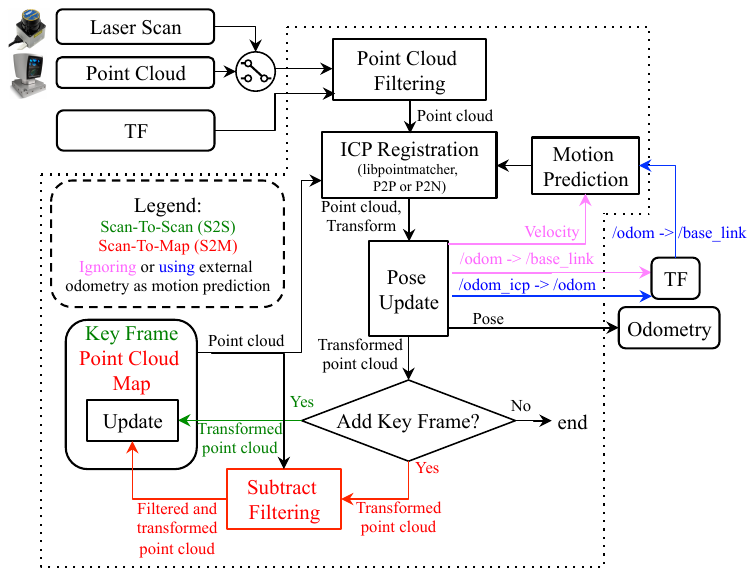} 
\caption{Block diagram of \textit{icp\_odometry} ROS node. TF defines the position of the lidar in relation to the base of the robot and as output to publish the odometry transform of the base of the robot. Two odometry approaches can be used: a Scan-To-Scan (S2S) approach in green, and a Scan-To-Map (S2M) approach in red. The approaches also have the choice of using a constant velocity model (pink) or another source of odometry (blue) for motion prediction. For the later, the correction of the input odometry is published on TF.} 
\label{fig:icp_odometry} 
\end{figure}

Figure \ref{fig:icp_odometry} provides the block diagram of lidar odometry, also using two colors to differentiate between S2S (green) and S2M (red). 
Using a terminology similar to visual odometry, a key frame refers to a point cloud or a laser scan. 
The laser scan input is 2D as the point cloud input can be either 2D or 3D. Laser scans can have some motion distortions when the robot is moving during the scan. It is assumed here that such distortions are corrected prior to feed the scan to RTAB-Map. Note that if the laser scanner rotation frequency is high comparatively to robot velocity, laser scans would have very low motion distortion and thus correction can be ignored without significant loss of registration accuracy. The process is described as follows: 

\begin{itemize}

\item Point Cloud Filtering: The input point cloud is downsampled and normals are computed. 
\textit{tf} is used to transform the point cloud into robot base frame so that odometry is computed accordingly (e.g., /base\_link).

\item ICP Registration: To register the new point cloud to Point Cloud Map (S2M) or the last Key Frame (S2S), iterative-closest-point (ICP) \cite{besl1992method} is done using implementation of \textit{libpointmatcher} \cite{Pomerleau12comp}. 
The Point Cloud Map is a cloud assembled by past key frames.
Registration can be done using Point to Point (P2P) or Point to Plane (P2N) correspondences. 
P2N is preferred in human-made environments with a lot of plane surfaces. 

\item Motion Prediction: As ICP is dealing with unknown correspondences, this module requires a valid motion prediction before estimating the transform, either from a previous registration or from external odometry approach (e.g., wheel odometry) though \textit{tf} (shown in blue and purple, respectively). 
Identity transform is provided as motion prediction only at initialization when processing the two first frames. 
If an external odometry is not used as an initial guess, motion prediction is done according to a constant velocity model based on the previous transformation. 
A problem with this technique is that if the environment is not complex enough (like in a corridor), odometry may drift a lot if there are no constraints on the direction of the robot. 
Using an external initial guess in this case can help estimate the motion in the direction in which the environment is lacking features. 
For example, a robot with a short-range lidar moving in a long corridor in which there are no doors (i.e., not distinguishable geometry) would only see two parallel lines. 
If the robot accelerates or decelerates in the direction of the corridor, ICP would be able to correct orientation but it would not be able to detect any changes in velocity in the direction of the corridor. 
In such case, using external odometry can help estimate velocity in the direction in which ICP cannot. 
If the structural complexity of the current point cloud is lower than the fixed threshold ``Icp/PointToPlaneMinComplexity", only orientation is estimated with ICP and position (along the problematic direction) is derived from the external odometry. 
The structural complexity of a 2D point cloud is defined as the second Eigenvalue of the Principal Component Analysis (PCA) of the point cloud's normals, multiplied by two. 
For 3D point cloud, the third Eigenvalue is used multiplied by three. 

\item Pose Update: After successful registration, odometry pose is then updated. 
When external odometry is used, \textit{tf} output is the correction of the external odometry \textit{tf} so that both transforms can be in the same \textit{tf} tree (i.e., /odom\_icp\textrightarrow /odom\textrightarrow /base\_link). 
Like visual odometry, covariance is computed using the MAD approach \cite{Rusu_ICRA2011_PCL} between 3D point correspondences.

\item Key Frame and Point Cloud Map Update: If the correspondence ratio is under the fixed threshold ``Odom/ScanKeyFrameThr", the new frame becomes the Key Frame for S2S. 
For S2M, an extra step is done before integrating the new point cloud to Point Cloud Map. 
The map is subtracted from the new point cloud (using a maximum radius of ``OdomF2M/ScanSubtractRadius"), then the remaining points are added to the Point Cloud Map. 
When the Point Cloud Map has reached the fixed maximum threshold ``OdomF2M/ScanMaxSize", oldest points are removed. 

\end{itemize}

In case ICP cannot find a transformation, odometry is lost. 
In contrast to visual odometry, lidar odometry cannot recover from being lost when the motion prediction is null, in order to avoid large odometry errors. 
Lidar odometry must then be reset. 
However, as long as the lidar can perceive environmental structures, the robot would rarely get lost. 
Note that if external odometry is used, motion prediction would still give a valid estimation, so ICP registration could recover from being lost if the robot comes back where it has lost tracking. 
Finally, similarly to the third party visual odometry approaches integrated, an open source version\footnote{\url{https://github.com/laboshinl/loam_velodyne}} of the lidar odometry approach called LOAM \cite{zhang2017low} has been integrated to RTAB-Map for comparison.

\subsection{Synchronization}
\begin{figure}[!t] 
\centering 
\includegraphics[width= 4in]{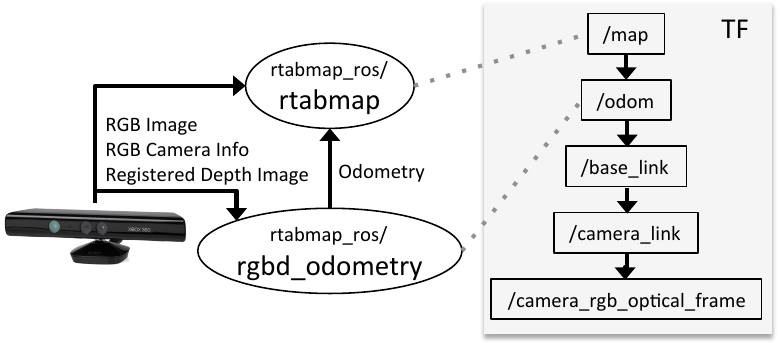} 
\caption{Visual SLAM with a RGB-D camera like the Kinect for Xbox 360. The \textit{rgbd\_odometry} ROS node is used to compute odometry for \textit{rtabmap} ROS node. 
On the right is a standard resulting TF tree for this sensor configuration (with transforms linked by a dotted line to corresponding publishing ROS nodes).} 
\label{fig:rgbdslam_setup} 
\end{figure}

\begin{figure}[!t] 
\centering 
\includegraphics[width= 5in]{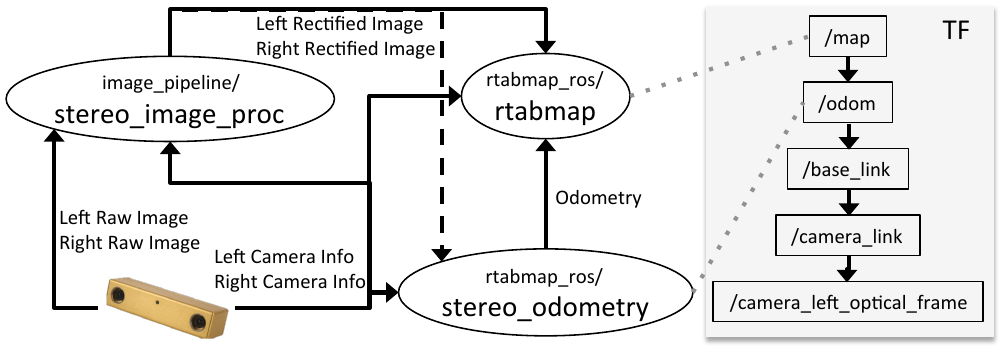} 
\caption{Visual SLAM with a stereo camera like the BumbleBee2. The \textit{stereo\_odometry} ROS node is used to compute odometry for \textit{rtabmap} ROS node. 
RTAB-Map's ROS nodes require rectified stereo images, thus the standard \textit{stereo\_image\_proc} ROS node is used to rectify them. On the right is a standard resulting TF tree for this sensor configuration (with transforms linked by a dotted line to corresponding publishing ROS nodes).} 
\label{fig:stereoslam_setup} 
\end{figure}

RTAB-Map has a variety of input topics (e.g., RGB-D images, stereo images, odometry, 2D laser scan, 3D point cloud and user data) that can be used depending on the sensors available. 
The minimum topics that are required to make the \textit{rtabmap} ROS node work are registered RGB-D or calibrated stereo images with odometry, provided through a topic or by \textit{tf} (e.g., /odom\textrightarrow /base\_link).
RTAB-Map also supports multiple RGB-D cameras as long as they have all the same image size. 
Accurate \textit{tf} of the sensors used are required (e.g., /base\_link\textrightarrow /camera\_link). 
Figure \ref{fig:rgbdslam_setup} and Figure \ref{fig:stereoslam_setup} illustrate two visual SLAM examples with corresponding \textit{tf} trees. 
RTAB-Map's visual odometry nodes can be replaced by any other odometry approaches (i.e., wheel odometry, other visual odometry package, lidar odometry, etc.). 
The dotted links show which node is publishing the corresponding \textit{tf}. 
For other \textit{tf} frames describing the position of the sensors on the robot, they are usually published by the camera driver, by some \textit{static\_transform\_publisher}\footnote{\url{http://wiki.ros.org/tf\#static_transform_publisher}} or by a \textit{robot\_state\_publisher}\footnote{\url{http://wiki.ros.org/robot_state_publisher}} using Unified Robot Description Format of the robot.

Once subscribed to basic sensors, there are two other topics that can be optionally synchronized: a 2D laser scan (e.g., Hokuyo and SICK lidars) or a 3D point cloud (e.g., Velodyne lidar) to generate 2D and 3D occupancy grids, respectively. 
They can be also used to refine the links in the graph using ICP. 

\begin{figure}[!t] 
\centering 
\includegraphics[width=5in]{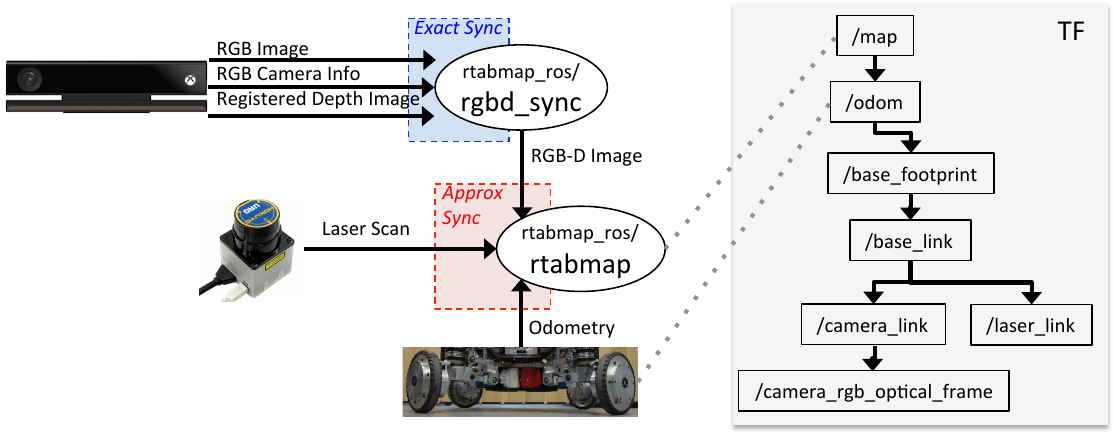} 
\caption{Synchronization example of a RGB-D camera (Kinect for Xbox One) with laser scan (URG-04LX) and odometry. In this case, odometry is computed through wheel encoders.
Camera messages are synchronized together using \textit{rgbd\_sync} ROS node before synchronizing the resulting RGB-D image message with the other sensors (which can have different publishing rates). On the right is an example of the resulting TF tree for this sensor configuration (with transforms linked by a dotted line to corresponding publishing ROS nodes).} 
\label{fig:sync1} 
\end{figure}

As sensors do not always publish data at the same rate and at the same exact time, good synchronization is important to avoid bad registration of the data. 
ROS provides two kind of synchronization: exact and approximate. 
An exact synchronization requires that input topics have exactly the same timestamp, i.e., for topics coming from the same sensor (e.g., left and right images of a stereo camera). 
An approximate synchronization compares timestamps of the incoming topics and try to synchronize all topics with a minimum delay error. 
It is used for topics coming from different sensors. 
Synchronization can then become a little tricky if a subset of input topics (i.e., camera topics) must be synchronized with the exact time policy while being approximately synchronized with other sensors. 
To do so, the \textit{rtabmap ros/rgbd\_sync} ROS nodelet can be used to synchronize camera topics into a single topic of type \textit{rtabmap\_ros/RGBDImage}\footnote{\url{http://docs.ros.org/api/rtabmap_ros/html/msg/RGBDImage.html}} before the \textit{rtabmap} node. 
Figure \ref{fig:sync1} presents a synchronization example with a RGB-D camera and a lidar. 
For RGB-D camera, ROS packages do not always provide exact same timestamps for RGB and depth images, and \textit{rgbd\_sync} can be also used with approximate synchronization to synchronize images at the camera frame rate (e.g., 30 Hz) independently of the rate of the other inputs (e.g., laser scan, odometry).

\subsection{STM}
\label{sec:local_grid}
When a new node is created in STM, in complement to the information described in \cite{labbe2017}, a local occupancy grid is now computed from the depth image, the laser scan or the point cloud. 
In case of stereo images, a dense disparity image is computed using a block matching algorithm \cite{konolige1998small}, and converted to a point cloud. 
A local occupancy grid is referenced in the robot frame and it contains empty, ground and obstacle cells at the fixed grid cell size ``Grid/CellSize". 
The total size of a local occupancy grid is defined by the range and field of view of the sensor used to create it. 
These local occupancy grids are used to generate a global occupancy grid by transforming them in map referential using the poses of the map's graph. 
While pre-computing the local occupancy grids requires more memory for each node, it greatly decreases the regeneration time of the global occupancy grid when the graph has been optimized. 
For example, in previous work \cite{labbe14online}, when the global occupancy grid had to be generated, all the laser scans had to be ray traced again. 

\begin{figure}[!t] 
\centering 
\includegraphics[width= 4in]{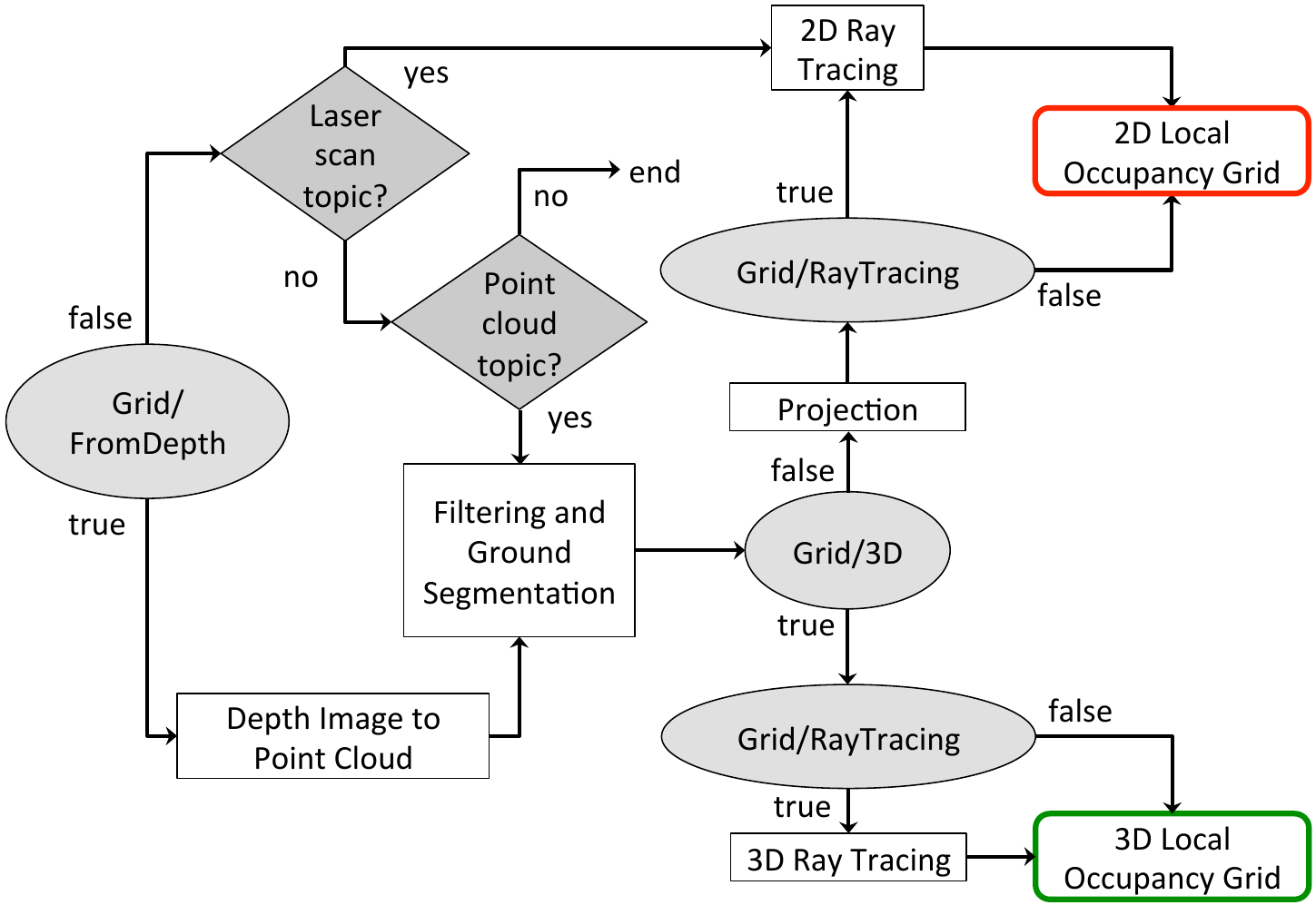} 
\caption{STM's local occupancy grid creation. Depending on the parameters (shown by ellipses) and the availability of the optional laser scan and point cloud inputs (shown by diamonds), the local occupancy grid can either be 2D or 3D.} 
\label{fig:local_grid_creation} 
\end{figure}

Depending on the parameters ``Grid/FromDepth", ``Grid/3D" and the input topics set, the local occupancy grid is generated differently and the result is either 2D or 3D, as shown in Figure \ref{fig:local_grid_creation}. 
For example, if parameter ``Grid/FromDepth" is false and \textit{rtabmap} node is subscribed to a laser scan topic, a local 2D occupancy grid is created. 
A 2D local occupancy grid requires less memory than the 3D one because there is one less dimension to save (e.g., $z$) and superposed obstacles can be reduced to only one obstacle cell. 
However, local 2D occupancy grids cannot be used to generate a 3D global occupancy grid, while local 3D occupancy grids can be used to generate 2D and 3D global occupancy grids. 
The choice depends on what kind of global map is required for the application and on the processing power available. 
Note that if ``Grid/FromDepth" is false and no laser scan and point cloud topics are subscribed, no grids are computed.
The rectangular boxes in Figure \ref{fig:local_grid_creation} are described as follows:

\begin{itemize}
\item 2D Ray Tracing: For each ray of the laser rangefinder, a line is traced on the grid to fill empty cells between the sensor and the obstacle hit by the ray. 
It is assumed that the rays are parallel to the ground. 
This approach can generate 2D local occupancy grids very fast and is done by default for 2D lidar-based mapping.

\item Depth Image to Point Cloud: The input depth image (or disparity image in case of stereo images) is projected in 3D space according to sensor frame and camera calibration. 
The cloud is then transformed in the robot base frame.

\item Filtering and Ground Segmentation: The point cloud 
is downsampled by a voxel grid filter \cite{Rusu_ICRA2011_PCL} with voxel size equals to fixed grid cell size. 
The ground plane is then segmented from the point cloud: 
the normals of the point cloud are computed, then all points with their normal parallel to $z$-axis (upward) within the fixed maximum angle ``Grid/MaxGroundAngle" are libelled as ground, others are obstacles.

\item Projection: If ``Grid/3D" is false, the 3D ground and obstacle point clouds are projected on ground plane (e.g., $x$-$y$ plane). 
The voxel grid filter is applied again to merge points projected in the same cell. 
2D ray tracing can be done to fill empty space between obstacles and the camera. 
If 2D ray tracing is not used and if the point cloud does not have any points segmented as ground, no empty cells are set in the occupancy grid between the sensor and the obstacles. 

\item 3D Ray Tracing: An OctoMap is created from the single local occupancy grid in the robot referential. 
OctoMap does 3D ray tracing and detects empty cells between the camera and occupied cells. 
The OctoMap is converted back to local occupancy grid format with empty, ground and obstacle cells.
\end{itemize}

\subsection{Loop Closure and Proximity Detection}
\label{loopclosure}

Loop closure detection is done using the bag-of-words approach described in \cite{labbe13appearance}. 
Basically, when creating a new node, STM extracts visual features from the RGB image and quantizes them to an incremental visual word vocabulary. 
Features can be any of the types included in OpenCV like SURF \cite{Bay08}, SIFT \cite{Lowe04}, ORB \cite{rublee2011orb} or BRIEF \cite{calonder2010brief}. 
When visual odometry F2F or F2M is used, it is possible to re-use features already extracted for odometry for loop closure detection. 
This eliminates extracting twice the same features. 
As loop closure detection does not need as many features than odometry to detect loop closures and to reduce computation load, only a subset (maximum of ``Kp/MaxFeatures") of the odometry features with highest response are quantized to visual word vocabulary. 
The other features are still kept in the node when loop closure transformation has to be computed. 
The created node is then compared to nodes in WM to detect a loop closure. 
STM contains the last nodes added to map, and therefore these nodes are not used for loop closure detection. 
Locations in STM would be very similar to last location and would bias loop closure hypotheses on them. 
STM can be seen as a buffer of a fixed size ``Mem/STMSize" before a node is moved WM. 
To compute likelihood between the created node and all those in WM, Tf-IDF approach \cite{sivic2003video} is used to update a Bayes filter estimating the loop closure hypotheses. 
The filter estimates if the new node is from a previously visited location or a new location. 
When a loop closure hypothesis reaches the fixed threshold ``Rtabmap/LoopThr", a loop closure is detected and transformation is computed. 
The transformation is computed using the same Motion Estimation approach used by visual odometry (Section \ref{sec:vo}), and if accepted, the new link is added to the graph. 
When a laser scan or a point cloud is available, link's transformation is refined using the same ICP Registration approach than with lidar odometry (described in Section \ref{sec:icp_odometry}).

Introduced in \cite{labbe2017}, proximity detection is used to localize nodes close to the current position with laser scans (when available). 
For example, with proximity detection, it is possible to do localization when traversing back the same corridor in a different direction, during which the camera can not be used to find loop closures. 
In contrast to loop closure detection where complexity depends on the WM size, the complexity of proximity detection is bounded to nodes close to the robot. 
These nodes must be close in the graph, i.e., the number of links between them and the latest node should be less than the fixed threshold ``RGBD/ProximityMaxGraphDepth". 
When odometry drifts over large distance, the robot may move to a previously mapped area that differs from the current real location, so using this threshold, proximity detection do not make comparison with nodes from the previously mapped area to avoid invalid proximity detections. 
If odometry does not drift too much or that the map update rate is higher, the threshold can be set higher, otherwise it should be lowered. 

\subsection{Graph Optimization}
\label{sec:optimization}
When a loop closure or a proximity detection are detected or some nodes are retrieved or transferred because of memory management, a graph optimization approach is applied to minimize errors in the map. 
RTAB-Map integrates three graph optimization approaches: TORO \cite{grisetti2010tutorial}, g2o \cite{kummerle11g2o} and GTSAM \cite{dellaert2012factor}. 
g2o and GTSAM converge faster than TORO, 
but are less robust to multi-session mapping when multiple independent graphes have to be merged together. 
TORO is also less sensitive to poorly estimated odometry covariance. 
However, for single map, based on empirical data, g2o and GTSAM optimization quality is better than TORO, particularly for 6DoF maps. 
GTSAM is slightly more robust to multi-session than g2o, and thus is the strategy now used by default in RTAB-Map contrarily to our previous works using TORO. 

Visual loop closure detection is not error-free, and very similar places can trigger invalid loop closure detections, which would add more errors to the map rather than reducing them. 
To detect invalid loop closure or proximity detections, RTAB-Map now uses a new parameter. 
If a link's transformation in the graph after optimization has changed more than than the factor ``RGBD/OptimizeMaxError" of its translational variance, all loop closure and proximity links added by the new node are rejected, keeping the optimized graph as if no loop closure happened.

\subsection{Global Map Assembling}
\label{sec:global_grid}

\begin{figure}[!t] 
\centering 
\includegraphics[width= 4in]{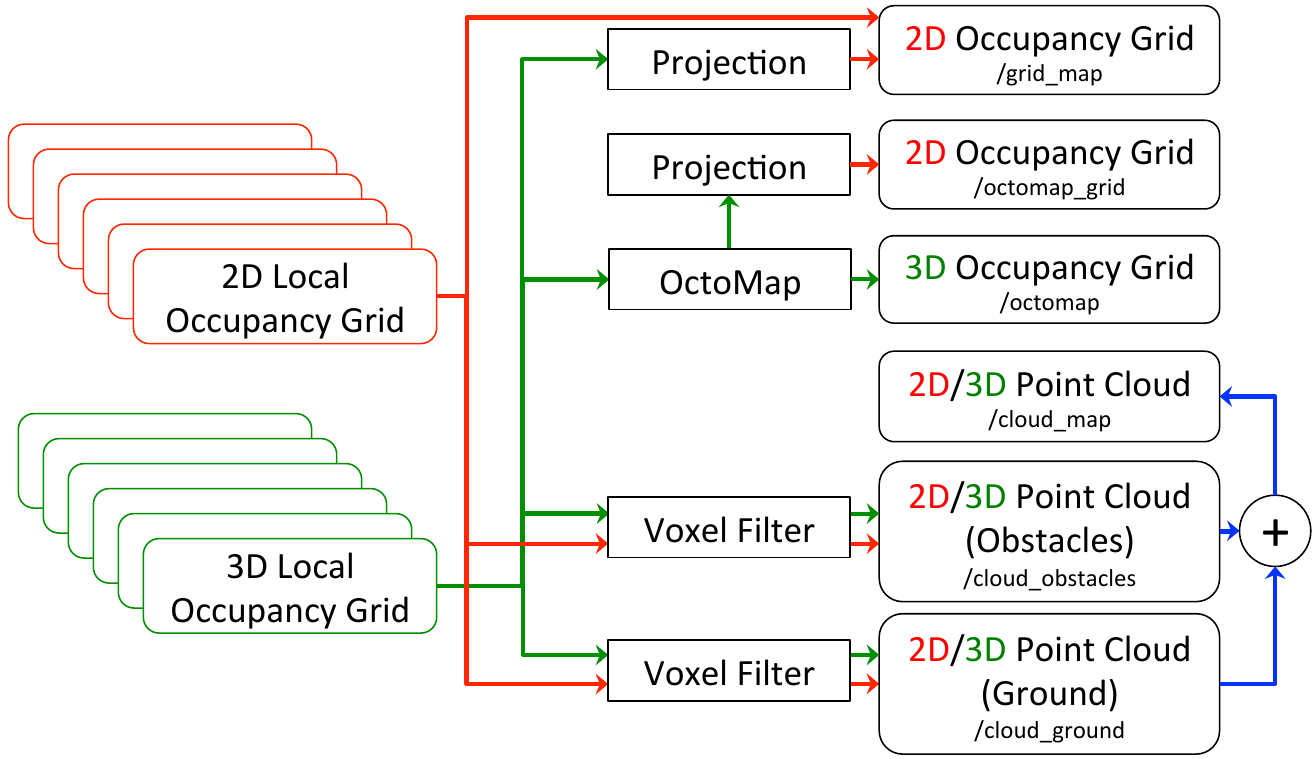} 
\caption{Global map assembling. Depending on the type of local maps created in the map's graph (see Figure \ref{fig:local_grid_creation}), the available output global maps will differ. Only 3D local occupancy grids can be used to generate the 3D occupancy grid (OctoMap) and its projection in 2D.} 
\label{fig:global_grid_creation} 
\end{figure}

Figure \ref{fig:global_grid_creation} illustrates the global map outputs that can be assembled from the local occupancy grids of Figure \ref{fig:local_grid_creation}. 
Saving 3D local occupancy grids in nodes gives to most flexibility, as they can be used to generate all types of map. 
However, if only a 2D global occupancy grid map is needed, saving already projected local grids in the nodes saves memory (two numbers per point instead of three) and time (points are already projected to 2D) when assembling the local maps. 
Using the map's graph, each local occupancy grid are transformed into its corresponding pose. 
When a new node is added to map, the new local occupancy grid is combined with the global occupancy grid, clearing and adding obstacles. 
When a loop closure occurs, 
the global map should be re-assembled according to all new optimized poses for all nodes in the map's graph. 
This process is required so that obstacles that have been incorrectly cleared before the loop closure can be reincluded. 
The point cloud outputs assemble all points of the local maps and publish them in the standard \textit{sensor\_msgs/PointCloud2}\footnote{\url{http://docs.ros.org/api/sensor_msgs/html/msg/PointCloud2.html}} ROS format. 
Voxel grid filtering is done to merge overlapping surfaces. 
The resulting point cloud is a convenient format for visualization and debugging, and ease integration with third party applications.

%% file: 4_results.tex
\section{Evaluating Trajectory Performance of RTAB-Map Using Different Sensor Configurations}
\label{sec:slamperf}

Performance of RTAB-Map has been evaluated on four datasets having a ground truth: 
KITTI \cite{geiger2012we}, TUM RGB-D \cite{sturm2012benchmark}, EuRoC \cite{burri2016euroc} and PR2 MIT Stata Center \cite{fallon2013stata}. 
These datasets have a variety of sensors (i.e., stereo and RGB-D cameras, 2D and 3D lidars, combined wheel and IMU odometry).
Using RTAB-Map as the common evaluation framework makes it possible to outline performance differences between all sensor and odometry configurations.
The metric used for trajectory accuracy is the absolute trajectory (ATE) root-mean-square error, derived from the TUM RGB-D benchmark \cite{sturm2012benchmark}. 
All errors are computed with the map's graph including loop closures. 

\begin{table*}[!t]
\caption{RTAB-Map (version $0.16.3$) Default Parameters}
\label{parameters}
\centering
\begin{tabular}{llll}
\hline
GFTT/MinDistance & 3 pixels  & RGBD/OptimizeMaxError & 1\\
GFTT/QualityLevel & 0.001 & RGBD/ProximityMaxGraphDepth & 50 nodes \\
Kp/MaxFeatures & 500 features & Rtabmap/DetectionRate & 2 Hz\\
Odom/KeyFrameThr (F2M)& 0.3 & Rtabmap/TimeThr & 0 ms \\
Odom/KeyFrameThr (F2F) & 0.6 & Rtabmap/MemoryThr & 0 nodes\\
Odom/ScanKeyFrameThr & 0.9 & Rtabmap/LoopThr & 0.11\\
OdomF2M/MaxSize & 2000 features & Vis/CorNNDR & 0.6\\
OdomF2M/ScanMaxSize & 10000 points & Vis/CorGuessWinSize & 20 pixels\\
OdomF2M/ScanSubtractRadius & 0.05 & Vis/MaxFeatures & 1000 features\\
Mem/RehearsalSimilarity & 0.2  & Vis/MinInliers & 20\\
Mem/STMSize & 30 nodes \\
\hline
\end{tabular}
\end{table*}

Experiments were conducted on a desktop computer using an Intel® Core™ i7-3770 Processor (four cores), 6 GB of RAM and 512 GB SSD running Ubuntu 16.04. 
To compare trajectory accuracy of the different SLAM configurations independently of the computation power available, KITTI, TUM and EuRoC datasets have been processed offline, 
so that every frames are processed by odometry even if in some cases odometry processing time is higher than the camera frame rate. 
To compare the overall computation load in relation to computation time, a single core is used for offline experiments. 
For the PR2 MIT Stata Center dataset, experiments were conducted online in ROS and are not limited to a particular core of the computer. Table \ref{parameters} shows default parameters (unless explicitly specified) used for all datasets. 
To provide a fair comparison between results from RTAB-Map and other SLAM approaches, RTAB-Map's memory management has been disabled (``Rtabmap/TimeTheshold" and ``Rtabmap/MemThreshold" set both to 0) in these experiments.

With the offline datasets, the different stereo odometry approaches tested are using their own stereo correspondence approaches, which can generate slightly different disparity values. We also observed that camera calibration are not always accurate: rectified images still contain some visible distortions. 
These problems can influence the scale of the created map, thus bias trajectory accuracy when comparing with the ground truth (for which we assume there is no scale error). 
As observed in \cite{murORB2}, depth image values from some TUM RGB-D sequences can be slightly off, causing also a scale problem. 
For a fair comparison between all visual odometry approaches, as we cannot know if the scale problem is caused by the camera calibration and/or the disparity computation approach, results are presented with the map scaled to minimize the error against the ground truth. 
When ORB-SLAM2 is used as odometry input to RTAB-Map, it is referred to as ORB2-RTAB to differentiate from results of ORB-SLAM2 full SLAM version. 
Similarly, when LOAM is used as odometry input to RTAB-Map, it is referred to as LOAM-RTAB.

Note that the focus of the paper is on pose estimation evaluation to compare SLAM approaches. 
Localization robustness is not addressed explicitly but in all the results presented, no wrong loop closures were accepted. 
For similar places, either they were rejected by the criteria of not having enough visual inliers (see Motion Estimation of Section \ref{sec:vo}) or because of large graph optimization errors (see Section \ref{sec:optimization}), making the evaluations robust to invalid localization for the datasets tested.

\subsection{KITTI}

In the KITTI dataset, stereo images were recorded from two synchronized monochrome PointGrey cameras installed on the top of a car. 
The dataset provides rectified stereo images of size 1241$\times$376 pixels with baseline of $0.54$ m at 10 Hz. 
The dataset contains also 3D point clouds coming from a Velodyne 64E installed on the top of the car. 
The point clouds are synchronized with stereo images at 10 Hz. 
For S2M and S2S approaches using the lidar, to reduce computation load and memory usage, the raw point clouds are downsampled using a voxel filter of 50 cm in the Point Cloud Filtering step. 
Based on preliminary tests, Point-to-Point ICP Registration has been chosen because it gives slightly better results than Point-to-Plane on this dataset, in particular in sequences where there are lot of trees and not many planes.
For S2M, ``OdomF2M/ScanSubtractRadius" is set to 50 cm to match the voxel filter. 
As Velodyne data is 360$^{\circ}$ with far range, ``Odom/ScanKeyFrameThr" is set to $0.8$ instead of $0.9$ to trigger new key frames less often. 
In comparison to other datasets, KITTI has larger images.
Therefore, parameters affected by image size are modified: ``GFTT/MinDistance" is set to 7, ``GFTT/QualityLevel" is set to 0.01, ``Vis/MaxFeatures" is set to 1500, ``Kp/MaxFeatures" is set accordingly to half of ``Vis/MaxFeatures" at 750 and ``OdomF2M/MaxSize" is set to 3000.

\begin{figure*}[!t]
\centering
\begin{tabular}{ccc}
\subfloat[00 - F2M]{\includegraphics[width=0.3\textwidth]{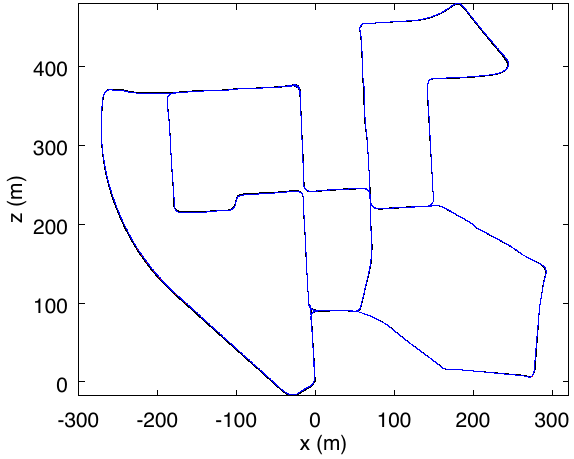}\label{}}
\subfloat[01 - F2M]{\includegraphics[width=0.3\textwidth]{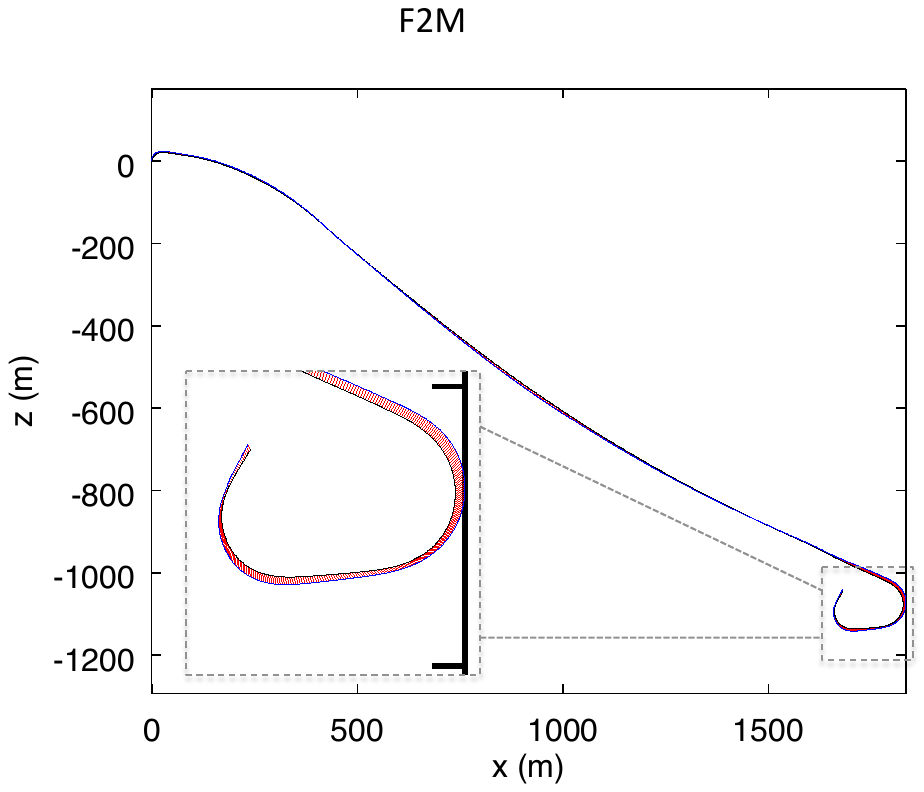}\label{}}
\subfloat[08 - F2M]{\includegraphics[width=0.3\textwidth]{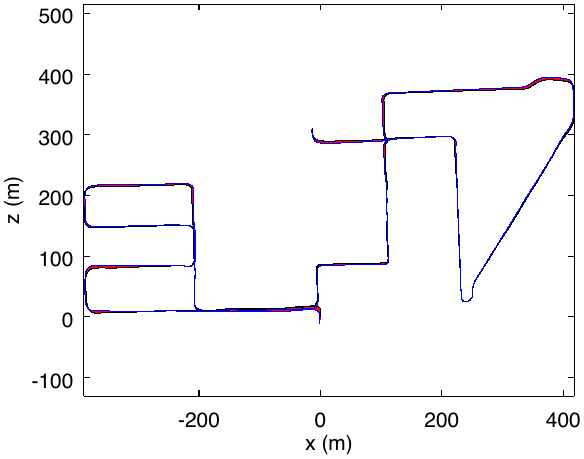}\label{}} \\
\subfloat[00 - S2S]{\includegraphics[width=0.3\textwidth]{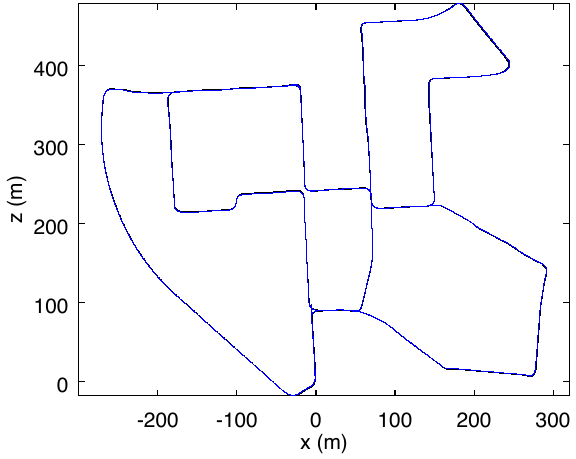}\label{}}
\subfloat[01 - S2S]{\includegraphics[width=0.3\textwidth]{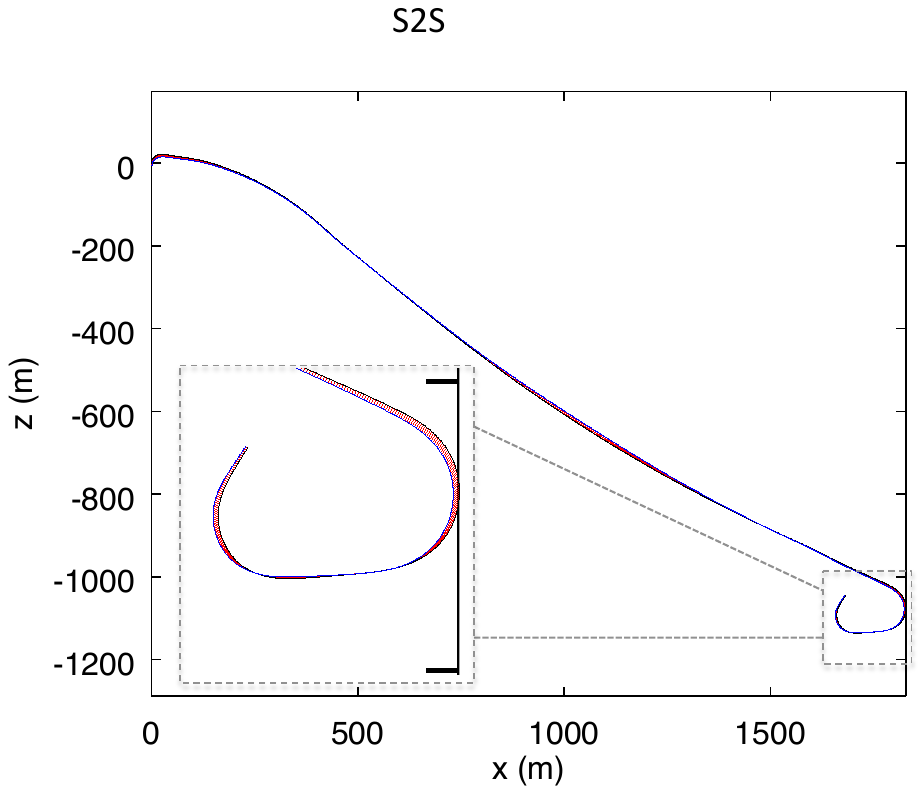}\label{}}
\subfloat[08 - S2S]{\includegraphics[width=0.3\textwidth]{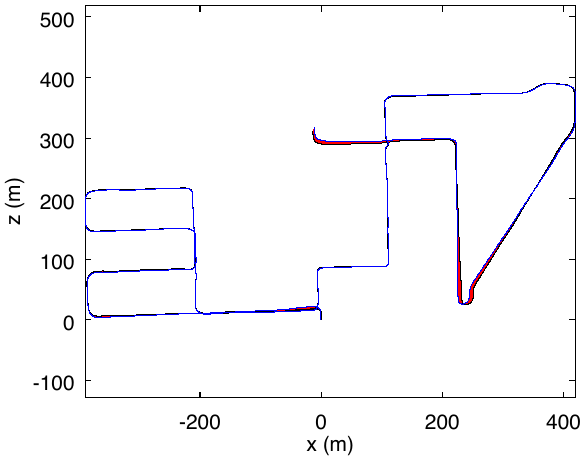}\label{}} 
\end{tabular}
\caption{Trajectories using RTAB-Map with stereo odometry F2M (top) and lidar odometry S2S (bottom) against ground truths for the three KITTI sequences 00, 01 and 08. Errors between poses estimated by RTAB-Map (blue) and the ground truths (black) are shown in red.}
\label{fig:kitti_paths}
\end{figure*}

\begin{table*}[!t]
\caption{ATE (m) results for the KITTI sequences in relation to the odometry approach and the sensor used using a single CPU core}
\label{kitti_rmse}
\centering
\begin{tabular}{|l|l|c|c|c|c|c|c|c|c|c|c|c||c|}
\hline
& RTAB-Map & \multicolumn{11}{c||}{KITTI Sequence} & $o_{avg}$  \\
Sensor& Odometry & 00 & 01 & 02 & 03 & 04 & 05 & 06 & 07 & 08 & 09 & 10 & (msec)  \\
\hline
  \multirow{3}{*}{Lidar}&S2S & \textbf{1.0} & 24.0 & 3.1 & 0.7 & 0.4 & 0.6 & \textbf{0.5} & \textbf{0.3} & 4.2 & \textbf{1.1} & 1.8 & 62\\
 &S2M & 1.1 & 17.2 & 2.9 & 0.7 & 0.5 & 0.7 & \textbf{0.5} & \textbf{0.3} & 7.7 & \textbf{1.1} & 1.7 & 82\\
  & LOAM-RTAB & 1.8 & 23.3 & 47 & 1.1 & 0.3 & 1.1 & 1 & 0.4 & 10.1 & 1.3 & 3 & 330 \\
  \hline
\multirow{5}{*}{\pbox{20cm}{Stereo \\ Camera}} &F2F & 1.4 & 14.5 & 4.7 & 0.4 & \textbf{0.2} & 0.72 & 1.8 & 0.6 & 5.8 & 2.2 & 3.0 & 61\\
 &F2M &  \textbf{1.0} & 4.7 & 4.7 & 0.3 & \textbf{0.2} & \textbf{0.5} & 0.8 & 0.5 & 3.8 & 2.8 & \textbf{0.8} & 82\\
 &Fovis &  28.8 & $\times$ & $\times$ & 2.0 & 1.9 & 11.4 & 24.5 & 3.8 & 25 & 33.3 & 13.6 & \textbf{17}\\
 &ORB2-RTAB & \textbf{1.0} & 5.3 & 4.4 & \textbf{0.2} & \textbf{0.2} & \textbf{0.5} & 0.6 & 0.5 & 3.0 & 1.5 & 0.9 & 175\\
 &Viso2 & 4.2 & 19.7 & 37.9 & 1.2 & 0.3 & 3.4 & 6.3 & 1.1 & 22.4 & 4.1 & 4.8 & 66\\
\hline
\hline
\multicolumn{2}{|l|}{LSD-SLAM (stereo)} & \textbf{1.0} & 9.0 & \textbf{2.6} & 1.2 & \textbf{0.2} & 1.5 & 1.3 & 0.5 & 3.9 & 5.6 & 1.5 & -\\
\multicolumn{2}{|l|}{ORB-SLAM2 (stereo)}  & 1.3 & 10.4 & 5.7 & 0.6 & \textbf{0.2} & 0.8 & 0.8 & 0.5 & 3.6 & 3.2 & 1.0 & -\\
\multicolumn{2}{|l|}{SOFT-SLAM (stereo)}  & 1.2 & \textbf{3} & 5.1 & 0.5 & 0.4 & 0.8 & \textbf{0.5} & \textbf{0.3} & \textbf{2.3} & 1.3 & 0.9 & -\\
\hline
\end{tabular}
\end{table*}

Figure \ref{fig:kitti_paths} presents a comparison of visual and lidar trajectories derived using RTAB-Map with stereo odometry (F2M) and lidar odometry (S2S), in relation to ground truths of three sequences in the KITTI dataset: sequence 00 has loop closures; sequence 01 has been taken on a highway at a speed of 90 km/h and higher; sequence 08 does not have loop closures (e.g., the camera is oriented in a different direction when traversing back the same segments).
When comparing the trajectories at that scale, there is not so much difference between visual and lidar approaches and both follow well the ground truth.
Table \ref{kitti_rmse} summarizes trajectory accuracy in terms of ATE for all odometry configurations available in RTAB-Map, along with performance reported for ORB-SLAM2 \cite{murORB2}, LSD-SLAM \cite{engel2015large} and SOFT-SLAM \cite{cvivsic2018soft}.
$o_{avg}$ is the average odometry time across all sequences when limiting the approach to a single CPU Core.
For three sequences (06, 07 and 09), using the Velodyne provides better performance compared to the visual-based approaches. 
However, for sequence 01, which is a highway sequence not geometrically complex, lidar-based approaches perform quite worst: a lot of error along $y$ axis (in KITTI coordinates) are caused by bad pitch estimation. 
In contrast, visual-based approaches can use features farther to lidar range to better estimate the pitch orientation. 
S2M and S2S give relatively the same errors, so choosing between the two could be based on computation time $o_{avg}$, with S2S always being the fastest. 
LOAM-RTAB is better only for the fourth sequence in comparison to other lidar-based approaches.
Comparing only visual-based approaches, ORB2-RTAB is the best for nine of the 11 sequences and F2M in six sequences (with ties for four sequences). 
However, ORB2-RTAB cannot satisfy real-time constraints ($o_{avg}$ is always over 100 ms). 
In the ORB-SLAM2 paper \cite{murORB2}, the sequences can be processed under 100 ms using multiple cores, while here only a single core is used. 
On less powerful computers using a stereo camera, F2M odometry seems the better choice. 
Fovis is the fastest configuration, but it gets lost quite easily: there are not enough visual inliers to compute transformation, and odometry is automatically reset inside its library, causing some missing motions in the map or very bad transformations (where ``$\times$" means a very high drift). 

\begin{table*}[!t]
\caption{Average translational error (\%) results for the KITTI sequences in relation to the odometry approach and the sensor used}
\label{kitti_rel_error}
\centering
\begin{tabular}{|l|l|c|c|c|c|c|c|c|c|c|c|c|}
\hline
 & RTAB-Map &  \multicolumn{11}{c|}{KITTI Sequence}  \\
Sensor&Odometry & 00 & 01 & 02 & 03 & 04 & 05 & 06 & 07 & 08 & 09 & 10 \\
\hline
 \multirow{3}{*}{Lidar}&S2S & 0.82 & 3.17 & 1.26 & 1.02 & 1.21 & 0.51 & 0.58 & 0.58 & 1.11 & 0.90 & 1.64\\
 &S2M & 0.86 & 2.52 & 1.14 & 1.03 & 1.18 & 0.56 & 0.58 & 0.65 & 1.25 & 0.90 & 1.52\\
 & LOAM-RTAB & 1.2 & 2.9 & 4.4 & 1.1 & 1.5 & 0.8 & 0.9 & 0.6 & 1.5 & 1.2 & 1.7 \\
 \hline
\multirow{5}{*}{\pbox{20cm}{Stereo \\ Camera}} &F2F & 0.85 & 2.38 & 1.01 & 0.90 & \textbf{0.35} & 0.49 & 1.25 & 0.62 & 1.56 & 1.24 & 1.71\\
 &F2M & 0.68 & 2.04 & 0.97 & 0.77 & 0.45 & 0.38 & 0.57 & 0.56 & 1.17 & 1.38 & \textbf{0.49}\\
 &Fovis &  9.09 & $\times$ & $\times$ & 1.79 & 2.22 & 4.26 & 6.95 & 3.65 & 5.39 & 14.8 & 10.6\\
 &ORB2-RTAB & 0.67 & 0.96 & \textbf{0.75} & \textbf{0.62} & 0.50 & \textbf{0.35} & 0.48 & 0.53 & 1.06 & 0.87 & 0.54\\
 &Viso2 & 2.38 & 5.92 & 4.19 & 1.94 & 0.66 & 1.85 & 4.60 & 1.04 & 2.82 & 1.68 & 1.93\\
\hline
\hline
\multicolumn{2}{|l|}{LOAM (lidar)} & 0.78 & 1.43 & 0.92 & 0.86 & 0.71 & 0.57 & 0.65 & 0.63 & 1.12 & 0.77 & 0.79\\
\multicolumn{2}{|l|}{LSD-SLAM (stereo)} & \textbf{0.63} & 2.36 & 0.79 & 1.01 & 0.38 & 0.64 & 0.71 & 0.56 & 1.11 & 1.14 & 0.72\\
\multicolumn{2}{|l|}{ORB-SLAM2 (stereo)}  & 0.70 & 1.39 & 0.76 & 0.71 & 0.48 & 0.40 & 0.51 & 0.50 & 1.05 & 0.87 & 0.60\\
\multicolumn{2}{|l|}{SOFT-SLAM (stereo)} & 0.66 & \textbf{0.86} & 1.36 & 0.70 & 0.50 & 0.43 & \textbf{0.41} & \textbf{0.36} & \textbf{0.78} & \textbf{0.59} & 0.68\\
\hline
\end{tabular}
\end{table*}

To facilitate comparison with other papers using the KITTI dataset, Table \ref{kitti_rel_error} presents the average translational error metric reported with the KITTI dataset \cite{geiger2012we}. 
Results of LOAM \cite{zhang2017low}, a lidar-based approach, are also shown. 
Translational error (in percent) have been computed for all possible subsequences of length (100 m, ..., 800 m) of each sequence, then averaged together. 
ORB2-RTAB approach performs best in 10 out of 11 sequences. 
The difference between ORB2-RTAB and ORB-SLAM2 can be explained by the difference between the loop closure detection and graph optimization approaches, and also by the map scaling process explained at the beginning of Section \ref{sec:slamperf}. 
LOAM, using a more complex scan matching approach, scores better on seven out of the 11 sequences when compared only to lidar odometry approaches. 
This suggests that extracting geometric features from the point clouds can help get better motion estimation than by using ICP only. 
When comparing LOAM-RTAB and LOAM against each other, either the open source version used (with default parameters for this kind of Velodyne) seems to not reproduce the original algorithm perfectly, or some parameters tuning is required, as we expected that results would be more similar. 
Finally, results of RTAB-Map with F2M odometry approach have been submitted to KITTI's odometry benchmark\footnote{\url{http://www.cvlibs.net/datasets/kitti/eval_odometry.php}} for the testing sequences 11 to 21. 
Table \ref{kitti_leaderboard} presents a snapshot of the current ranking with other approaches in Table \ref{kitti_rmse}. 
Compared to popular ORB-SLAM2 and LSD-SLAM approaches, RTAB-Map's translation error is very close, even slightly better in terms of rotation performance. 
Note that LOAM, SOFT-SLAM and LSD-SLAM (stereo version) are not currently available as a C++ library or with ROS, which make them difficult to use on a real robot.

\begin{table*}[!t]
\caption{Current state of KITTI's odometry leaderboard}
\label{kitti_leaderboard}
\centering
\begin{tabular}{|l|l|c|c|c|c|c|c|c|c|c|c|c|}
\hline
Rank & Method &  Setting & Translation & Rotation  \\
\hline
2 & LOAM & Lidar & 0.61 \% & 0.0014 [deg/m]  \\
5 & SOFT-SLAM & Stereo & 0.65 \% & 0.0014 [deg/m]  \\
30 & ORB-SLAM2 & Stereo & 1.15 \% & 0.0027 [deg/m]  \\
38 & LSD-SLAM & Stereo & 1.20 \% & 0.0033 [deg/m]  \\
47 & RTAB-Map (F2M) & Stereo & 1.26 \% & 0.0026 [deg/m]  \\
73 & Viso2 & Stereo & 2.44 \% & 0.0114 [deg/m]  \\
\hline
\end{tabular}
\end{table*}

\subsection{TUM}

The TUM RGB-D dataset was recorded using a handheld Kinect v1 in small office-like environments. 
RGB and depth Images were recorded at 30 Hz and are synchronized using the tool provided with the dataset.

\begin{figure*}[!t]
\centering
\begin{tabular}{ccc}
\subfloat[fr1-room]{\includegraphics[width=0.3\textwidth]{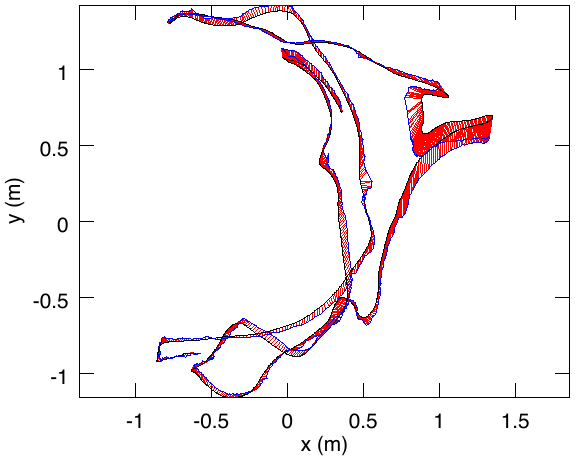}\label{tum_room}}
\subfloat[fr2-xyz]{\includegraphics[width=0.3\textwidth]{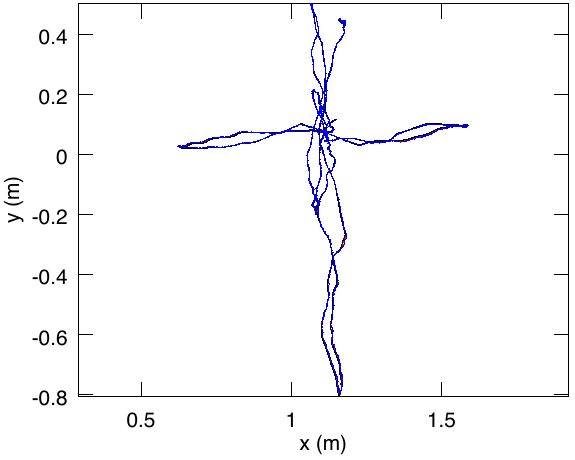}\label{}}
\subfloat[fr3-office]{\includegraphics[width=0.3\textwidth]{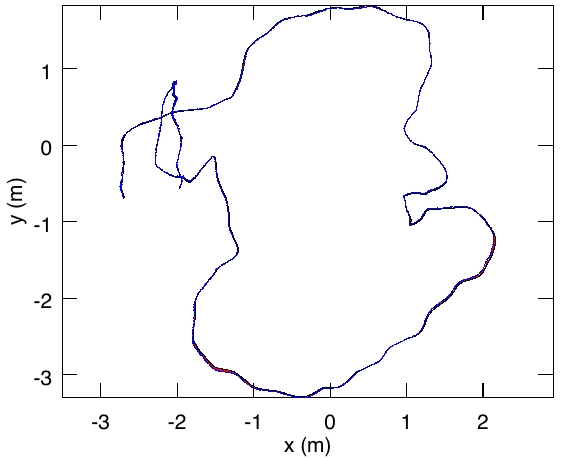}\label{}}
\end{tabular}
\caption{Trajectories using RTAB-Map with RGB-D odometry F2M (blue) against ground truths (black) for three TUM sequences. Errors between poses estimated by RTAB-Map and the ground truth are shown in red.}
\label{fig:tum_paths}
\end{figure*}

Figure \ref{fig:tum_paths} illustrates the trajectories for F2M compared to ground truths for three TUM sequences. 
In the fr1 sequence, the camera is moving and rotating faster than in other sequences, resulting in an estimated trajectory diverging more from the ground truth. 
When moving fast with this kind of camera, synchronization between RGB and depth images is poor (i.e., RGB pixels do not always match with the right depth pixels), causing bad motion estimations.
Table \ref{tum_rmse} presents ATE results with additional ones from other approaches like Elastic Fusion \cite{whelan2016elasticfusion}, Kintinuous  \cite{whelan2015real}, DVO SLAM \cite{kerl2013dense}, RGBDSLAMv2 \cite{endres20143}, RGBiD-SLAM \cite{gutierrez2016dense} BundleFusion\cite{dai2017bundlefusion} for comparison. 
Comparing RTAB-Map approaches together, ORB2-RTAB scores best in six out of the seven TUM sequences, followed by F2M. 
Higher errors observed with ORB2-RTAB compared to the original ORB-SLAM2 are because in the ORB-SLAM2 approach a global bundle adjustment is done after a loop closure: as there are a lot of visual features shared between many key frames, global bundle adjustment can indeed provide, with more computation time, better optimization than only optimizing the links between the graph's nodes as in RTAB-Map. 
In comparison to RTAB-Map's odometry approaches F2F and F2M, ORB2-RTAB seems less sensitive to large depth error at distance greater than 4 m with this kind of sensor, and to bad synchronization between RGB and depth cameras. 
In other words, ORB-SLAM2 refines the 3D position of the features in the feature map when new frames arrive, providing better triangulation of the features even if the initial depth taken form the depth image is erroneous. 
This could explain why ORB2-RTAB outperforms F2F and F2M. 
F2M still perform well in comparison to other visual SLAM approaches. 
Also, because of the fast rotation motions in fr1 sequences, F2F has problems tracking the features with optical flow in comparison with feature matching used by F2M. 
Finally, Fovis is the fastest and the only real-time odometry approach (under 33 ms), followed by F2F and DVO.

\begin{table*}[!t]
\caption{ATE (cm) results for the TUM sequences in relation to the odometry approach}
\label{tum_rmse}
\centering
\begin{tabular}{|l|ccc|cc|cc||c|}
\hline
 RTAB-Map & \multicolumn{3}{c|}{TUM fr1} &  \multicolumn{2}{c|}{TUM fr2} &  \multicolumn{2}{c||}{TUM fr3} & $o_{avg}$  \\
Odometry & desk & desk2 & room & desk & xyz & office & nst & (msec)  \\
\hline
 F2F & 7.2&  10.1 & 8.8 & 2.2 & 0.5 & 2.6 & 7.4 & 37 \\
 F2M & 2.9 & 4.4 & 6.6 & 2.4 & 0.5 & 2.1 & 1.7 & 70 \\
 DVO & 5.9 & 6.7 & 10.7 & 6.0 & 0.8 & 10.8 & 3.5 & 37\\
 Fovis & 4.8 & 8.8 & 11.9 & 4.7 & 0.7 & 5.1 & 10.6 & \textbf{21} \\
 ORB2-RTAB & 1.9 & 4.3 & 10.3 & 1.2 & \textbf{0.4} & 1.7 & 1.3 & 54 \\
\hline
\hline
BundleFusion & \textbf{1.6} & - & - & - & 1.1 & 2.2 & \textbf{1.2} & - \\
DVO SLAM & 2.1 & 4.6 & 5.3 & 1.7 & - & 3.5 & - & -\\
Elastic Fusion & 2.0 & 4.8 & 6.8 & 7.1 & 1.1 & 1.7 &1.6 & -\\
Kintinuous  & 3.7 & 7.1 & 7.5 & 3.4 & 2.9 & 3.0 & 3.1& -\\
ORB-SLAM2  & \textbf{1.6} & \textbf{2.2} & \textbf{4.7} & \textbf{0.9} & \textbf{0.4} & \textbf{1.0} & 1.9 & -\\
RGBiD-SLAM & 3.2 & 6.6 & 8.7 & 7.5 & - & 6.4 & - & - \\
RGBDSLAMv2  & 2.6 & - & 8.7 & 5.7 & - & - & - & -\\
\hline
\end{tabular}
\end{table*}

\subsection{EuRoC}

\begin{figure*}[!t]
\centering
\begin{tabular}{ccc}
\subfloat[V1-02-medium, F2M]{\includegraphics[width=0.29\textwidth]{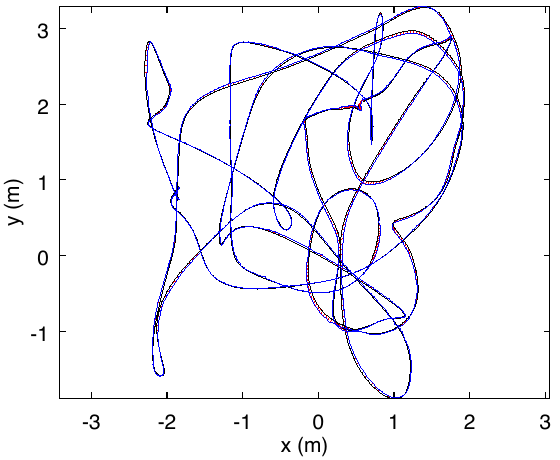}\label{}}
\subfloat[V2-03-difficult, F2M]{\includegraphics[width=0.29\textwidth]{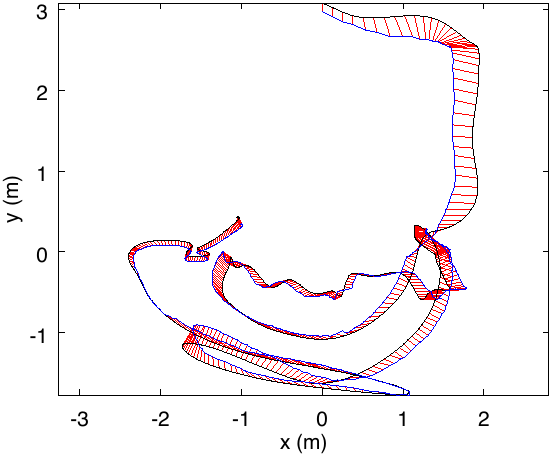}\label{}}
\subfloat[MH-04-difficult, F2M]{\includegraphics[width=0.29\textwidth]{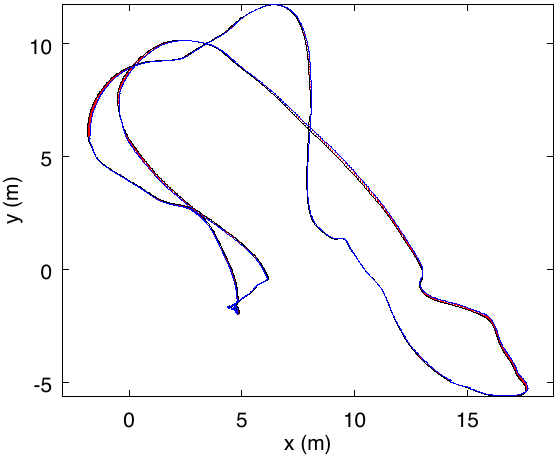}\label{}} \\
\subfloat[V1-02-medium, OKVIS]{\includegraphics[width=0.29\textwidth]{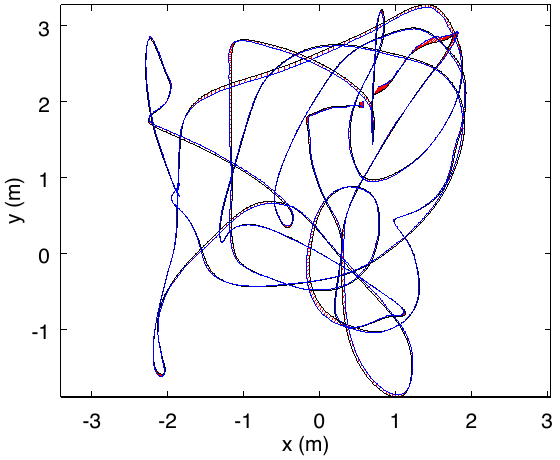}\label{}}
\subfloat[V2-03-difficult, OKVIS]{\includegraphics[width=0.29\textwidth]{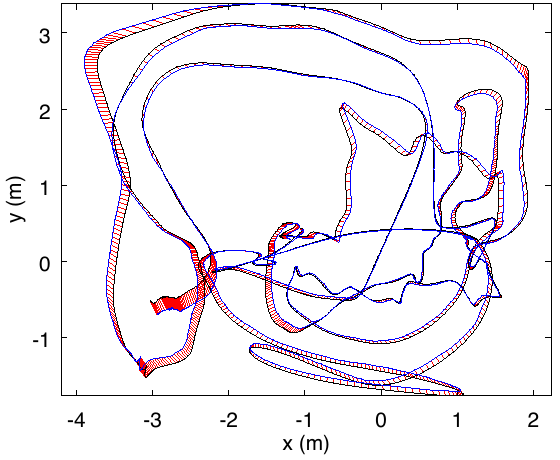}\label{}}
\subfloat[MH-04-difficult, OKVIS]{\includegraphics[width=0.29\textwidth]{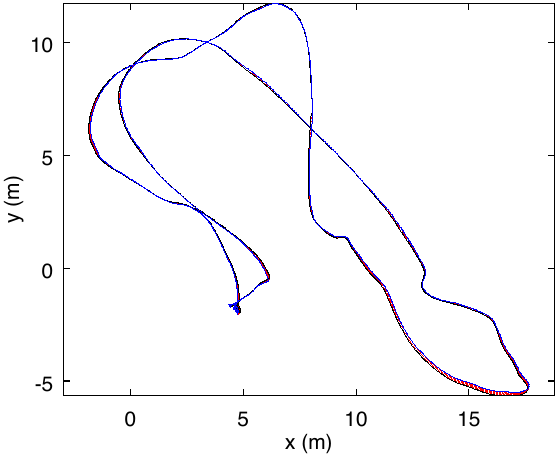}\label{}} 
\end{tabular}
\caption{Trajectories using RTAB-Map with stereo odometry F2M (top) and visual-inertial odometry OKVIS (bottom) against ground truths for three EuRoC sequences. Errors between poses estimated by RTAB-Map (blue) and the ground truths (black) are shown in red.}
\label{fig:euroc_paths}
\end{figure*}

\begin{table*}[!t]
\caption{ATE (cm) results for the EuRoC sequences in relation to the odometry approach}
\label{euroc_rmse}
\centering
\begin{tabular}{|l|ccc|ccc|ccccc||c|}
\hline
RTAB-Map &\multicolumn{3}{c|}{EuRoC V1} & \multicolumn{3}{c|}{EuRoC V2} &  \multicolumn{5}{c||}{EuRoC MH} & $o_{avg}$  \\
Odometry & 01 & 02 & 03 & 01 & 02 & 03 & 01 & 02 & 03 & 04 & 05 & (msec)  \\
\hline
 F2F & 8.4  & 6.9 & x & 17 & 89 & $\times$ & 3.1 & 4.2 & 12 & 12 & 10.1 & 40\\  
 F2M & 7.1 & 4.0 & 9.7 & 8.2 & 12 & $\times$ & \textbf{1.7} & 2.5 & 6.8 & 16 & 7.6 & 71\\
 Fovis & 10 & 26 & $\times$ & 44 & 80 & $\times$ & 4.4 & 10 & 9.9 & 32 & 36 & \textbf{16}\\ 
 ORB2-RTAB & 7.8 & 2.4 & 18 & 11 & 5.5 & $\times$ & 1.8 & \textbf{1.5} & \textbf{2.6} & 11 & \textbf{5.3} & 100\\ 
 Viso2 & 11 & 7.3 & 20 & 13 & 45 & $\times$ & 7.8 & 6.3 & 23 & 25 & 23 & 80\\ 
 OKVIS (IMU+stereo) & 4.2 & 3.2 & 7.2 & 16.2 & 11.7 & \textbf{14.4} & 4.0 & 3.3 & 7.6 & 10.0 & 10.2 & 272\\ 
 MSCKF (IMU+stereo) & 6.0 & 4.8 & 13 & 13.5 & 11.9 & 15.7 & 8.8 & 8.7 & 9 & 16 & 12 & 9\\ 
\hline
\hline
LSD-SLAM (stereo) & 6.6 & 7.4 & 8.9 & - & - & - & - & - & - & - & - & -\\
ORB-SLAM2 (stereo) & \textbf{3.5} & \textbf{2.0} & \textbf{4.8} & \textbf{3.7} & \textbf{3.5} & $\times$ & 3.5 & 1.8 & 2.8 & 12 & 6.0 & -\\
SOFT-SLAM (stereo) & 4.2 & 3.4 & 5.7 & 7.2 & 6.9 & 17.3 & 2.8 & 4.2 & 3.8 & \textbf{9.6} & 5.8 & -\\
\hline
\end{tabular}
\end{table*}

The EuRoC dataset has 11 stereo image sequences at 20 Hz taken on a drone in small indoor rooms (V1 and V2) and in a machine room (MH). 
Synchronized IMU data with camera are also available. 
While the images are time synchronized, exposure level between the cameras is not (e.g., right image can be darker with lower contrast than the left one). 
This increases the difficulty to find good stereo correspondences between left and right images when computing the disparity. 
To mitigate this problem, exposure compensation \cite{xu2010performance} is done between left and right images before processing them by the odometry approaches. 
For OKVIS odometry, the IMU is used along the stereo images.

Figure \ref{fig:euroc_paths} shows the paths computed for three EuRoC sequences using a stereo odometry approach (F2M) and the visual inertial odometry approach of OKVIS, compared against the ground truths. 
Except for the V2-03-difficult sequence where F2M fails to estimate the whole trajectory, the results between stereo visual odometry and visual inertial odometry are similar.
Table \ref{euroc_rmse} presents ATE results for all sequences. 
Overall, ORB2-RTAB performs better on six of the 11 sequences when compared to other RTAB-Map's odometry approaches, but is the second most computationally expensive approach. 
OKVIS and MSCKF are the only approaches able to track the whole V2-03-difficult sequence. 
Other approaches fail in this sequence when there is fast motion with a lot of motion blur, making difficult to track features: a visual inertial odometry approach is more robust to these kind of events. 
Fovis, F2F and MSCKF are the only real-time approaches (under 50 msec). 
For OKVIS and MSCKF, the processing time (on a single CPU core) is the average of image updates, excluding IMU updates that are done under 1 ms.
For V1 and V2 sequences, ORB2-RTAB performs worst than the results reported in the ORB-SLAM2 paper \cite{murORB2}: global bundle adjustment performed by ORB-SLAM2 on loop closures would then give better optimization than only using graph optimization done by RTAB-Map for these sequences. The opposite is observed however for the MH sequences.
LSD-SLAM has only been tested on V1 sequence, and results are slightly better than F2M on two out of three sequences.

\subsection{MIT Stata Center}
\label{sec:mit}

MIT Stata Center dataset is a collection of ROS bags recorded on a PR2 robot teleoperated in an office environment. 
The two sequences we use are 2012-01-25-12-14-25 and 2012-01-25-12-33-29. 
These sequences were chosen because they have many sensors: 2D lidar data, stereo images, RGB-D images and combined wheel and IMU odometry, making them perfect to compare different configurations of RTAB-Map. 
When replaying the ROS bags, the sensor data are published at the same rate as on the robot, allowing to test online capabilities of SLAM algorithms. 
Also, these two sequences overlap, allowing to test multi-session SLAM. 
The laser scans recorded in these bags are from a 2D long-range lidar of 30 m (UTM30) at 40 Hz. 
To test with a short-range lidar, ROS \textit{laser\_filters} package\footnote{\url{http://wiki.ros.org/laser_filters}} was used to filter the scans up to a maximum distance of 5.6 m to emulate a lower cost short-range lidar (like an URG04LX). 
For S2M and S2S approaches using the lidar, laser scans are downsampled using a voxel filter of 5 cm, then normals are computed during the Point Cloud Filtering step. 
Point to plane ICP registration is done with ``Icp/PointToPlaneMinComplexity" set to $0.02$.

To make comparison possible using the MIT Stata Center dataset, the following issues had to be addressed:

\begin{itemize}
\item Because these ROS bags are large (30 to 50 GB) and a lot of data have to be streamed in real-time, the hard-drive on the computer had some difficulties to correctly replay the bags, sometimes causing lags and message dropping.  
This is particularly annoying with visual odometry approaches for which a constant stream of images should be received to avoid getting lost;  lidar approaches are less affected because the large field of view of the laser scans let them recover from almost any orientation.
For this reason, a stereo and a RGB-D bags were created for each sequence with images at 15 Hz instead of 30 Hz, allowing the computer to stream images without lagging. 
Visual odometry approaches are tested using these 15 Hz bags, and lidar approaches are tested using the original bags.  
For visual odometry approaches able to process images faster than 15 Hz on the target computer, this could indeed impact negatively their performance in comparison than using images at 30 Hz. 
However, based on our experiments, 15 Hz is a good trade-off between able to process all images online while not getting lost (at the speed of the PR2), thus getting a better comparison of their performance independently of the frame rate.

\item We observed that the scale of stereo or RGB-D data are slightly off in comparison to lidar data: a factor of $1.091664$ has been applied to stereo baseline for stereo camera setup, and a factor of $1.043$ has been applied to scale the depth image for RGB-D camera setup. 

\item The lidar is located on the base and the cameras on the head of the robot, looking directly forward. 
The ground truth included in the bag refers to /laser\_link frame.
Therefore, a special node has been then created to transform it back in /base\_footprint frame so that all approaches (e.g., lidar-based or visual-based) refer to same base frame on the robot. 

\item We also observed that there is an offset with the timestamps between the ground truth frame and the topics, which is probably a technical error caused when the ground truth was recorded in the rosbag. 
Therefore, republishing the ground truth in /base\_footprint frame is done with offset of 82.2 sec to synchronize it with the other topics. 
\end{itemize}

\begin{figure*}[!t]
\centering
\begin{tabular}{cc}
\subfloat[Stereo Camera - F2M]{\includegraphics[width=0.4\textwidth]{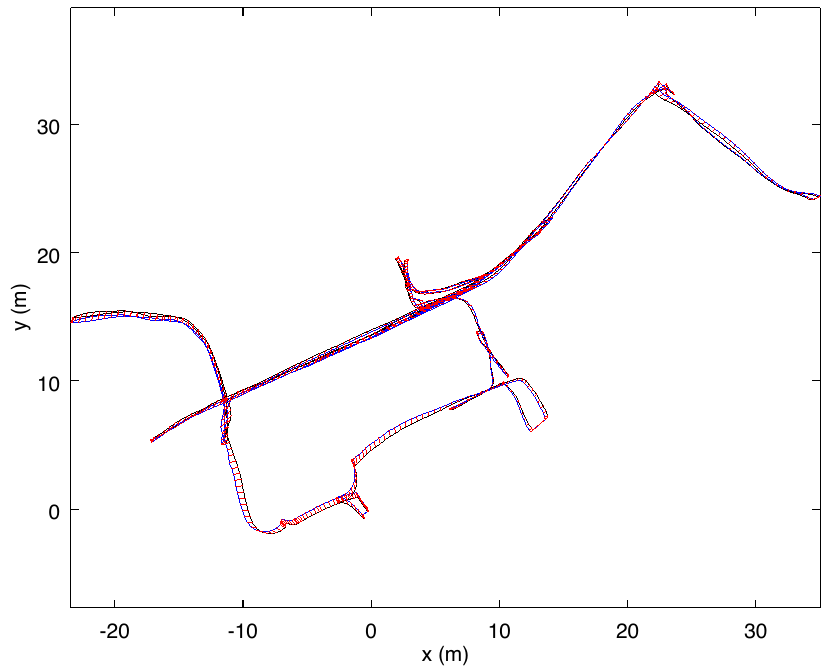}\label{}}
\subfloat[Stereo Camera - F2M]{\includegraphics[width=0.405\textwidth]{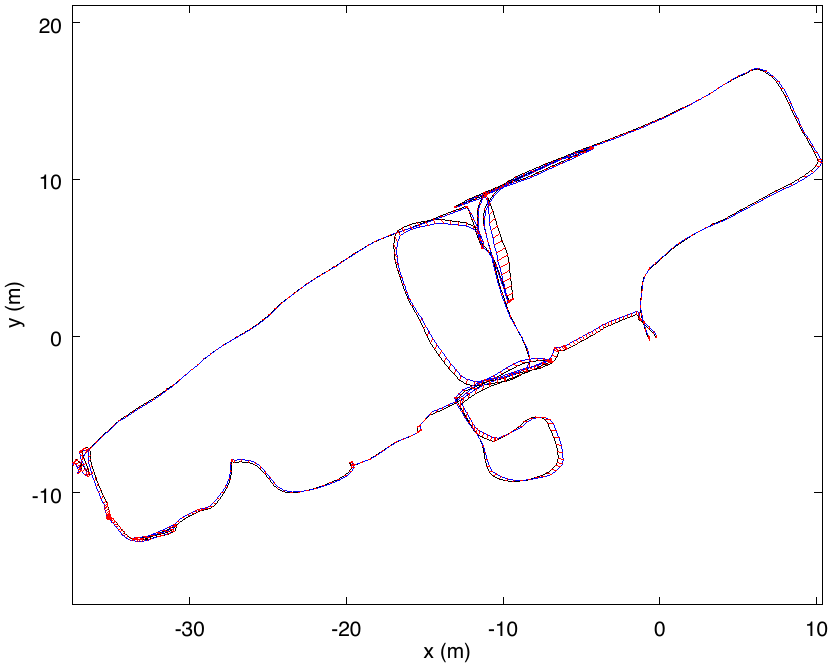}\label{}} \\
\subfloat[Long-Range Lidar - WheelIMU\textrightarrow S2M]{\includegraphics[width=0.4\textwidth]{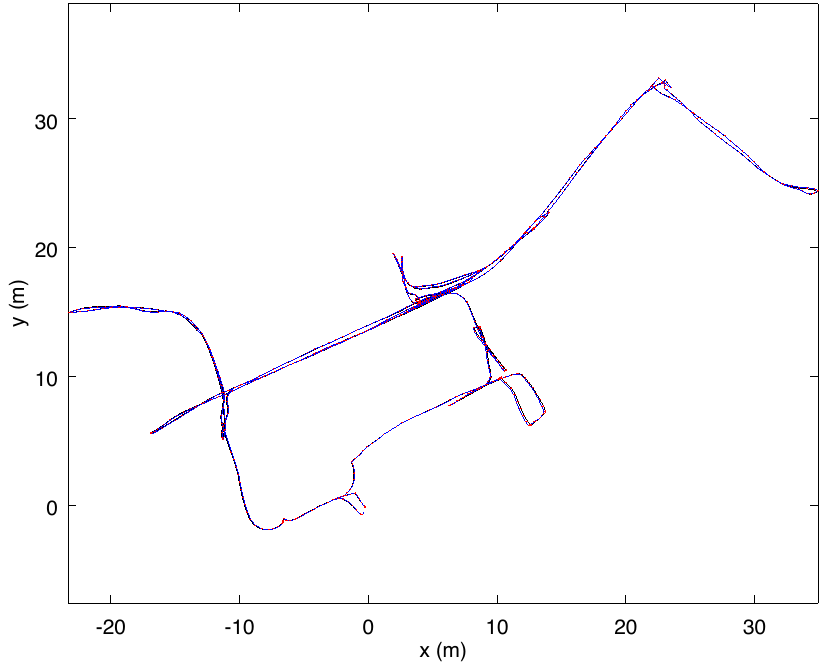}\label{}}
\subfloat[Long-Range Lidar - WheelIMU\textrightarrow S2M]{\includegraphics[width=0.405\textwidth]{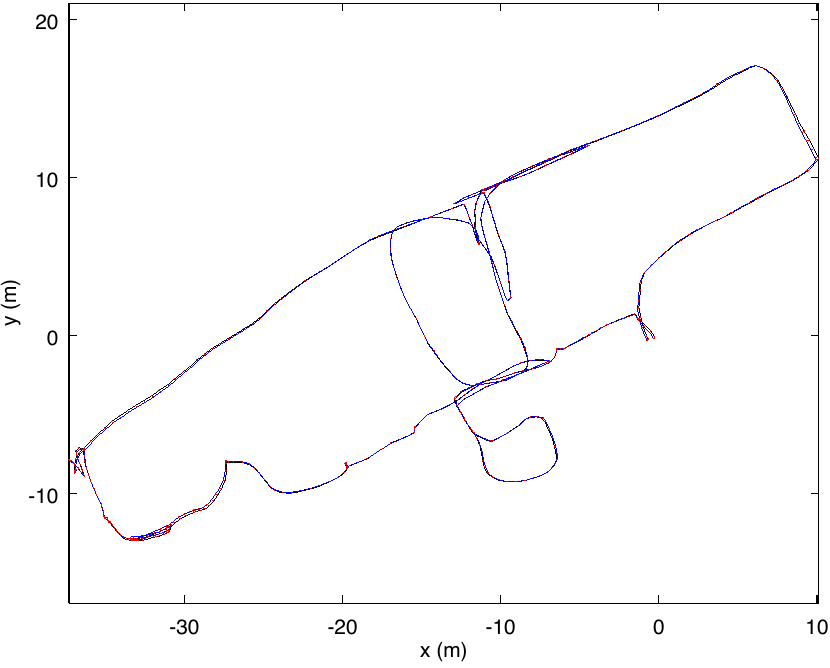}\label{}} 
\end{tabular}
\caption{Trajectories using RTAB-Map (blue) against ground truths (black) for the 2012-01-25-12-14-25 (left) and 2012-01-25-12-33-29 (right) Stata Center sequences using stereo camera (top) or long-range lidar (bottom). Errors between poses estimated by RTAB-Map and the ground truths are shown in red.}
\label{fig:stata_paths}
\end{figure*}

\begin{table*}[!t]
\caption{Online results for the MIT Stata Center 2012-01-25-xx-xx-xx sequences in relation to the sensor used and the odometry approach}
\label{stata_rmse}
\centering
\begin{tabular}{l|l|cc|cc|c}
\hline
& & \multicolumn{2}{c|}{12-14-25} &  \multicolumn{2}{c|}{12-33-29} & \\
& & ATE\textsubscript{end} & ATE\textsubscript{max} & ATE\textsubscript{end} & ATE\textsubscript{max}  & $o_{avg}$  \\
Sensor&Odometry & (m) & (m) & (m) & (m)  & (msec)  \\
\hline
\multirow{6}{*}{\pbox{20cm}{Long-Range \\ Lidar}}&WheelIMU\textrightarrow S2S & 0.06 & 0.08 &  0.08 & \textbf{0.09}  & \textbf{15} \\
&WheelIMU\textrightarrow S2M & \textbf{0.05} & \textbf{0.05} & 0.08 & \textbf{0.09}  & 25 \\
&S2S & \textbf{0.05} & 0.08 & \textbf{0.07} & 0.10 & \textbf{15} \\
&S2M  & \textbf{0.05} & 0.06 & \textbf{0.07} & 0.10 & 25 \\
&WheelIMU\textsubscript{refined} & 0.07 & 0.10 & 0.11 & 0.11  & -  \\
&WheelIMU & 0.09 & 0.26 & 0.10 & 1.13 & -\\
\hline
\multirow{6}{*}{\pbox{20cm}{Short-Range \\ Lidar}}&WheelIMU\textrightarrow S2S & 0.26 & 0.27 & 0.63 & 0.70  & \textbf{15} \\
&WheelIMU\textrightarrow S2M & \textbf{0.07} & \textbf{0.08} & \textbf{0.09} & \textbf{0.10}& 22  \\
&S2S & 11 & 11 & 4.47 & 4.47 & \textbf{15}\\
&S2M\  & 4.61 & 4.64 & 1.27 & 1.28 & 22 \\
&WheelIMU\textsubscript{refined}  & \textbf{0.07} & 0.12 &  0.11 & 0.14 & - \\
&WheelIMU & 0.12 & 0.39 & 0.12 & 2.23 & -\\
\hline
\multirow{6}{*}{Stereo Camera}&F2M  & 0.30 & 0.47 & 0.23 & \textbf{0.47} & 60 \\
&F2F  & 0.28 & 0.61 & 0.34 & 1.09 & \textbf{40}\\
&Fovis &  $\times$ & $\times$ & $\times$ & $\times$ & $\times$\\  
&ORB2-RTAB  & 0.227 & \textbf{0.31} & 0.37 & 0.57 & 45 \\
&Viso2 & 0.88 & 3.64 & 0.71 & 1.4 & 100 \\
&WheelIMU & \textbf{0.12} & 1.10 & \textbf{0.16} & 3.95  & -  \\
\hline
\multirow{6}{*}{RGB-D Camera}&F2M  & 0.37 & 0.91 & 0.38 & \textbf{0.80} & 54 \\
&F2F  & 0.38 & 0.69 & 0.42 & 0.94 & \textbf{32}\\
&Fovis & $\times$ & $\times$ & $\times$ & $\times$ & $\times$ \\
&ORB2-RTAB  & 0.28 & \textbf{0.64} & 0.49 & 0.85 & 44 \\
&DVO  & 0.56 & 0.77 & 0.55 & 1.62 & 45\\
&WheelIMU  & \textbf{0.11} & 1.30 &  \textbf{0.19} & 3.60  & - \\
\hline
\end{tabular}
\end{table*}

ATE values are computed at each frame to see their evolution over time. 
ATE\textsubscript{max} is the maximum error during the experiment, and ATE\textsubscript{end} is the error at the end of the experiment.
ATE\textsubscript{max} is a indication of which approach is better for autonomous navigation minimizing odometry drift, and ATE\textsubscript{end} is a indication about how well the final map represents the environment. 
As the robot is moving relatively slow and sequences are long, ``Rtabmap/DetectionRate" is set to 1 Hz to minimize memory usage, and ``Mem/STMSize" is set to half the size (15). 
WheelIMU is a new odometry type introduced in comparison with experiments done with the other datasets: WheelIMU is the odometry computed by combining odometry estimated by wheel encoders and the IMU using an Extended Kalman Filter \cite{officemarathon}, which is already available in the bags. 
When WheelIMU is set as prefix to S2M and S2S approaches, it means that WheelIMU is fed as external odometry estimation to S2M or S2S. 
WheelIMU\textsubscript{refined} indicates that neighbor links are refined using the laser scans when a new node is added to STM in the mapping module (``RGBD/NeighborLinkRefining" is true). 
For lidar-based odometry approaches, the RGB-D camera is used for loop closure detection in RTAB-Map. 
Stereo camera could also be used for loop closure, but computing motion estimation with RGB-D images is slightly faster than with stereo images (avoiding stereo correspondences computation).

Figure \ref{fig:stata_paths} illustrates trajectory comparison between a stereo-based approach (F2M) and a long-range lidar-based approach (WheelIMU\textrightarrow S2M). 
While the final results are roughly similar, the lidar-based approach follows the ground truth almost perfectly.
Table \ref{stata_rmse} presents the resulting ATE performance for each sequence.
Long-range lidar configurations are the most accurate (lowest ATE\textsubscript{end} and ATE\textsubscript{max}), and there are not so much differences between S2M and S2S approaches using WheelIMU or not. 
For short-range lidar, it is better to use WheelIMU\textrightarrow S2M over WheelIMU\textrightarrow S2S. 
The poor results of S2M and S2S with short-range lidar are caused by the corridors in the sequences: as explained in Section \ref{loopclosure}, when entering a corridor with a constant speed and a short-range lidar, the robot cannot know if it is accelerating or decelerating while seeing only two parallel lines, so with the constant motion assumption, it drifts along the corridor direction until it reaches the end of the corridor; for WheelIMU\textrightarrow S2M and WheelIMU\textrightarrow S2S approaches, this does not happen as the external odometry (using the wheel and IMU combined) can reveal that the robot is stopping at the middle of the corridor for example. 

In term of computation time, lidar odometry approaches are faster than visual odometry ones (lowest $o_{avg}$), with S2S approaches faster for all lidar experiments. 
Note that WheelIMU approaches do not have any computational cost ($o_{avg}$) because this is the odometry saved in ROS bags used directly as input to RTAB-Map. 
Since accuracy is similar, one may be tempted to choose WheelIMU\textsubscript{refined} over WheelIMU\textrightarrow S2M to minimize computation cost, but the advantage of the later is for navigation: the lidar odometry can be used as odometry input for other ROS modules which should run independently of the mapping module at higher frame rate (e.g., move\_base's local costmap can be more accurately updated using odometry from WheelIMU\textrightarrow S2M than WheelIMU). 

Therefore, lidar SLAM outperforms visual SLAM in this kind of environment. 
Comparing only visual approaches, stereo input gives better results than RGB-D input. 
This can be explained by the poor depth accuracy of the RGB-D sensor at range greater than 4 meters. 
In contrast, with a stereo camera, farther features have better depth estimation, which helps improve motion estimation. 
ORB2-RTAB and F2M perform better on the first and second sequences, respectively. 
With input images at 15 Hz, all visual odometry approaches are able to process frames in real-time (under 66 ms). 
Because of its memory leak explained in Section \ref{sec:vo}, ORB2-RTAB requires a lot more RAM with 1600 MB instead of 230 MB for other approaches. 
As Fovis gets lost very often, it is not able to complete the trajectories. 
WheelIMU performs better than visual odometry approaches if only the final error ATE\textsubscript{end} of the map is considered. 
This is mainly explained by the lack of visual features in some areas, where visual odometry can drift a lot more than WheelIMU which results in less consistent motion estimations, influencing the quality of the graph optimization. 
However, during mapping, WheelIMU reaches larger errors, which makes it less suitable for navigation.

\subsubsection{Lidar-Based SLAM Comparison}

\begin{figure*}[!t]
\centering
\begin{tabular}{cc}
\subfloat[2012-01-25-12-14-25 Short-Range Lidar]{\includegraphics[width=0.5\textwidth]{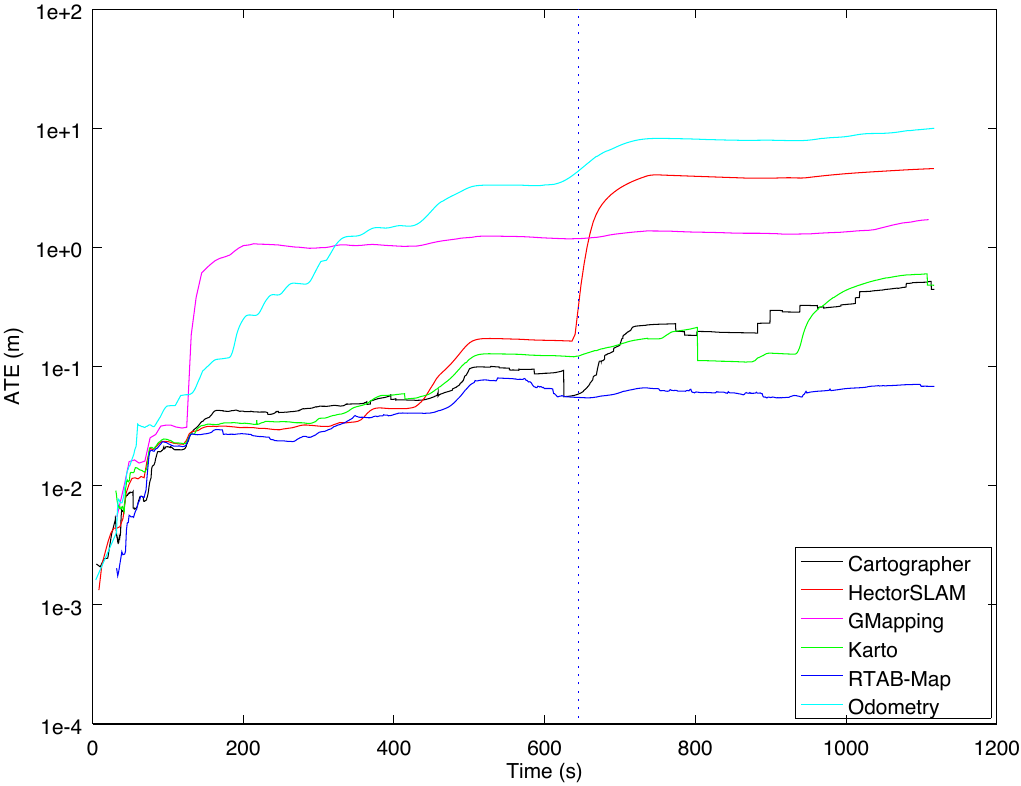}\label{}}
\subfloat[2012-01-25-12-33-29 Short-Range Lidar]{\includegraphics[width=0.5\textwidth]{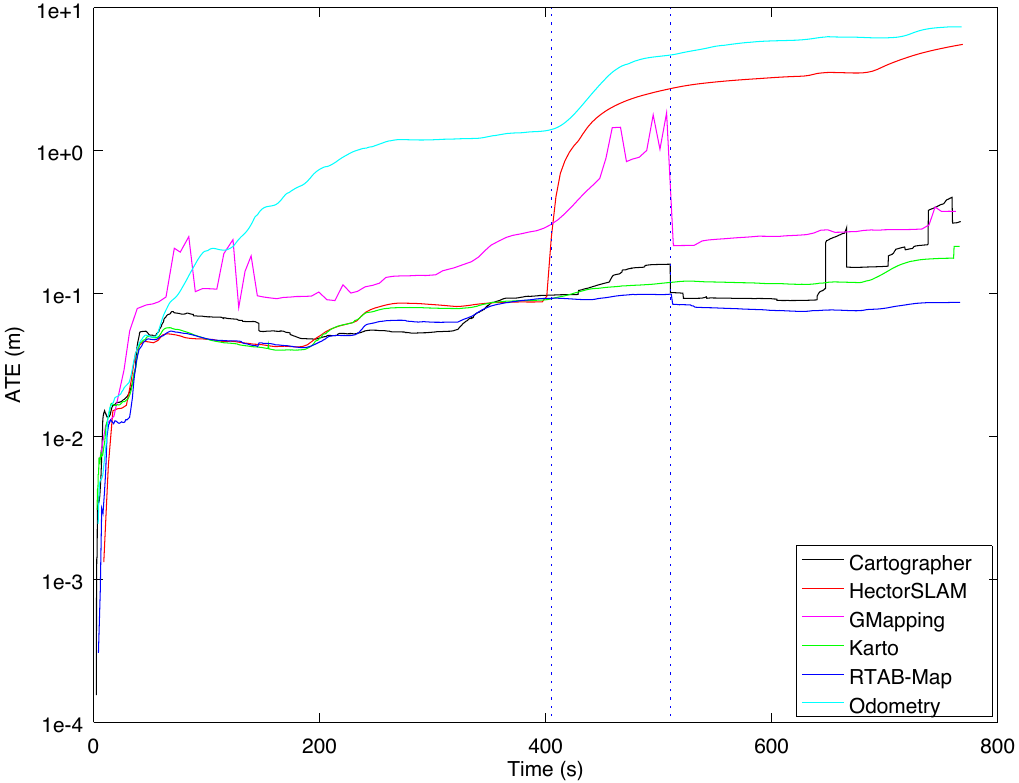}\label{}} \\
\subfloat[2012-01-25-12-14-25 Long-Range Lidar]{\includegraphics[width=0.5\textwidth]{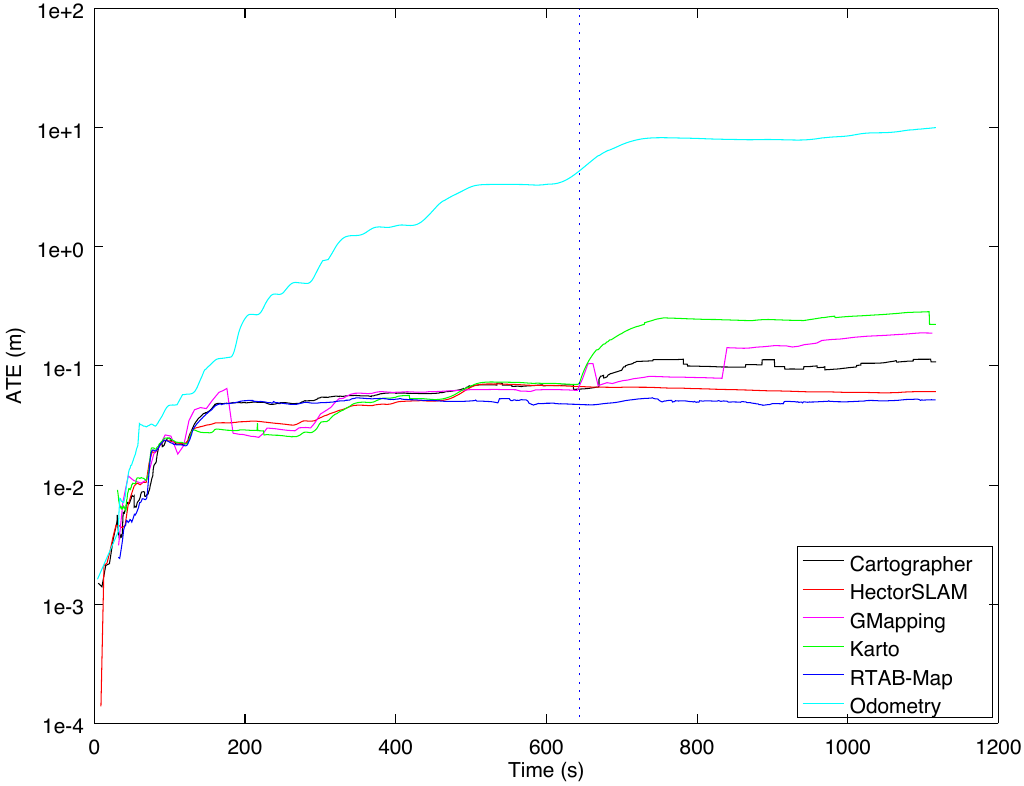}\label{}}
\subfloat[2012-01-25-12-33-29 Long-Range Lidar]{\includegraphics[width=0.5\textwidth]{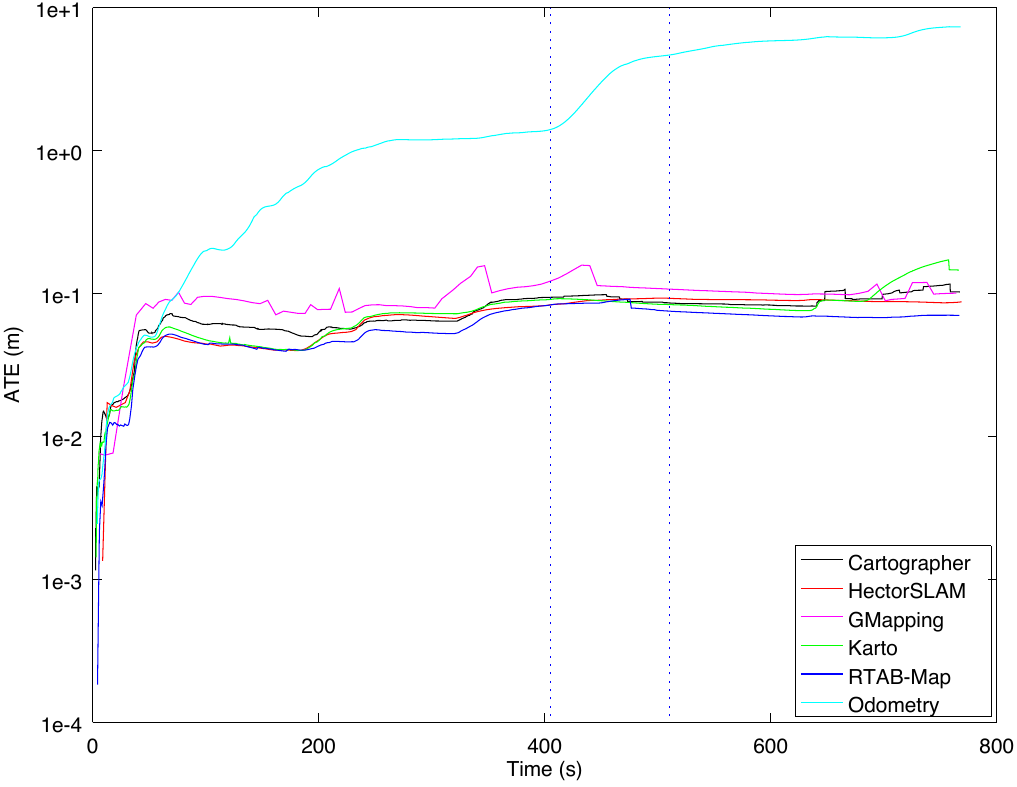}\label{}}
\end{tabular}
\caption{Comparison of RTAB-Map's WheelIMU\textrightarrow S2M with other lidar-based SLAM approaches}
\label{fig:compare_lidar}
\end{figure*}

To illustrate further the capabilities of the extended version of RTAB-Map, we conducted experiments comparing RTAB-Map lidar-based SLAM using the WheelIMU\textrightarrow S2M configuration against other popular open source lidar-based SLAM approaches, i.e., Google Cartographer \cite{hess2016real}, Karto SLAM \cite{vincent2010comparison}, Hector SLAM \cite{KohlbrecherHectorSLAM2011} and GMapping \cite{grisetti2007improved}. 
Their ROS implementations have been used with default parameters for long-range and short-range lidar data. 
For Google Cartographer, GMapping and Karto SLAM, combined WheelIMU odometry is also used as input. 
Figure \ref{fig:compare_lidar} and Table \ref{stata_community} summarize ATE results. 
For GMapping, as it is estimating multiple paths using its particle filter, ATE values are computed against the current best path published. 
An ATE of 1 m means that if at that moment the robot has to return to a specific location in the map, there is at least 1 m of error (without considering the accumulating error afterward if there is no localization) to reach it. 
Therefore, for autonomous navigation, ATE should be always as low as possible, and when ATE increases, it means that position estimated by the robot is drifting. 
If this value drifts too much, autonomous navigation to return to an known area may become impossible. 
To reduce the error, the robot has to localize itself on the map, which can be done by loop closure detection. 
The occurrence of loop closure detection can be observed when ATE decreases. 
For example, in Figure \ref{fig:compare_lidar}b, a large loop closure is found around 500 msec (on the second vertical dotted line). 
For long-range lidar, RTAB-Map's WheelIMU\textrightarrow S2M is the approach drifting the less, and is equal to Hector SLAM regarding ATE\textsubscript{max} on the second sequence. 
For short-range lidar, RTAB-Map's WheelIMU\textrightarrow S2M is the best for both sequences. 
Other approaches drifted a lot in corridor sections when using short-range lidar. 
In particular, Hector SLAM being the only approach not using external odometry to help scan matching, it diverges a lot more when traversing the corridors (identified by vertical lines just after 600 and 400 sec in each sequence, respectively). 

\begin{table*}[!t]
\caption{ATE (m) results of RTAB-Map's WheelIMU\textrightarrow S2M and popular lidar-based SLAM approaches on 2012-01-25 sequences.}
\label{stata_community}
\centering
\begin{tabular}{l|l|l|cc|cc}
\hline
& & & \multicolumn{2}{c|}{12-14-25} &  \multicolumn{2}{c}{12-33-29}\\
Sensor&Odometry&SLAM & ATE\textsubscript{end} & ATE\textsubscript{max} & ATE\textsubscript{end} & ATE\textsubscript{max}   \\
\hline
\multirow{5}{*}{\pbox{20cm}{Long-Range \\ Lidar}}& WheelIMU\textrightarrow S2M & RTAB-Map & \textbf{0.05} & \textbf{0.05} & \textbf{0.08} & \textbf{0.09}  \\
& WheelIMU &  Cartographer & 0.11 & 0.11 & 0.10 & 0.12 \\ 
& WheelIMU &  GMapping & 0.19 & 0.19 & 0.10 & 0.16\\
& WheelIMU &  Karto SLAM & 0.22 & 0.29 & 0.15 & 0.17\\
& - & Hector SLAM & 0.06 & 0.07 & 0.09 & \textbf{0.09} \\
\hline
\multirow{5}{*}{\pbox{20cm}{Short-Range \\ Lidar}}& WheelIMU\textrightarrow S2M & RTAB-Map  & \textbf{0.07} & \textbf{0.08} & \textbf{0.09} & \textbf{0.10} \\
& WheelIMU & Cartographer & 0.45 & 0.52 & 0.32 & 0.47 \\
& WheelIMU & GMapping & 1.71 & 1.71 & 0.38 & 1.84\\
& WheelIMU & Karto SLAM & 0.48 & 0.60 & 0.21 & 0.21\\
& - & Hector SLAM & 4.59 & 4.59 & 5.53 & 5.53 \\
\hline
\end{tabular}
\end{table*}

\section{Evaluating Computation Performance between Visual and Lidar SLAM Configurations with RTAB-Map}
\label{sec:computationPerf}

\begin{table*}[!t]
\caption{Occupancy grid performance using MIT Stata Center 2012-01-25-12-14-25 sequence after 860 nodes added to the graph (or 350 meters in 19 minutes)}
\label{stata_occupancy}
\centering
\begin{tabular}{l|l|c|l|cc}
\hline
 \multicolumn{3}{c|}{\multirow{2}{*}{Local Occupancy Grid}} & \multicolumn{3}{c}{Global Occupancy Grid} \\
\multicolumn{3}{c|}{}  & Type & \multicolumn{2}{c}{Time} \\
Type & Sensor & Time & & Update+Pub & With Loop \\
&  &  (msec)  & & (msec) & (msec)\\
\hline
GFD\textsubscript{0}\textrightarrow 2D & Long-Range Lidar & 4 &  2D & 2+0 & +600  \\
\hline
GFD\textsubscript{0}\textrightarrow 2D & Short-Range Lidar  & 1 & 2D & 1+0 & +200 \\
\hline
GFD\textsubscript{1}\textrightarrow G3D\textsubscript{0}  & RGB-D Camera & 13 & 2D & 1+0 & +40  \\
 \textrightarrow GRT\textsubscript{0}\textrightarrow 2D & Stereo Camera & 23 & 2D  & 1+0 & +25\\
\hline
GFD\textsubscript{1}\textrightarrow G3D\textsubscript{0}  & RGB-D Camera & 14 & 2D & 1+0 & +90 \\
\textrightarrow GRT\textsubscript{1}\textrightarrow 2D & Stereo Camera & 24 & 2D & 1+0 & +90 \\
\hline
GFD\textsubscript{1}\textrightarrow G3D\textsubscript{1}  & RGB-D Camera & 13 & OctoMap 3D & 2+100 & +1600\\ 
\textrightarrow GRT\textsubscript{0}\textrightarrow 3D & Stereo Camera & 23 & OctoMap 3D & 2+80 & +1100\\
\hline
\multirow{4}{*}{\pbox{20cm}{GFD\textsubscript{1}\textrightarrow G3D\textsubscript{1} \\  \textrightarrow GRT\textsubscript{1} \textrightarrow 3D}} & RGB-D Camera&  120 & OctoMap 2D & 15+540 & +11300  \\
 & Stereo Camera & 96 & OctoMap 2D & 20+670 & +14800  \\ 
 & RGB-D Camera & 120 & OctoMap 3D & 15+430 & +13500  \\ 
  & Stereo Camera & 96 & OctoMap 3D & 20+640 & +15500 \\ 
\hline
\end{tabular}
\end{table*}

As seen in Section \ref{sec:local_grid}, depending on the sensor and configurations chosen, some approaches are available to create local occupancy grids and to assemble a global occupancy grid described in Section \ref{sec:global_grid}. 
The choices have an impact on computation time, memory usage and map quality. 
Table \ref{stata_occupancy} presents all the configurations possible. 
For the local occupancy grid, the possibilities derived from Figure \ref{fig:local_grid_creation} involve GFD (Grid/FromDepth), G3D (Grid/3D) and GRT (Grid/RayTracing). 
The global occupancy grid presented in Figure \ref{fig:global_grid_creation} involves 2D (2D Global Occupancy Grid), OctoMap 2D (2D Global Occupancy Grid from OctoMap projection) and OctoMap 3D (3D Global Occupancy Grid). 
Results are presented using the 2012-01-25-12-14-25 sequence of the MIT Stata Center dataset, with ``Grid/CellSize" set to 5 cm and ``Grid/MaxGroundAngle" set to 45$^{\circ}$.
Time for the local occupancy grid refers to the time required by STM to create the local occupancy grid. 
The times for the global occupancy grid is the time needed to update the global occupancy grid: the Update time is the time required to assemble the new local occupancy grid to the global occupancy grid; the Pub time is the time required to serialize the global occupancy grid and to publish it as a ROS topic; and the With Loop time is the additional time required to re-assemble the whole global occupancy grid when the graph has been optimized after a loop closure. 
As expected, generating 2D local occupancy grids from 2D lidar data is faster than from depth or disparity images, as there are less points to process. 
When using stereo input, an additional 10 msec is required to compute the disparity image in comparison to RGB-D, for which the depth image can be used directly. 
Ray tracing in 2D for camera inputs adds 1 msec (e.g., 14 msec vs 13 msec for RGB-D camera). 

\begin{figure*}[!t]
\centering
\begin{tabular}{ccccc}
\subfloat[Lidar]{\includegraphics[width=0.35\textwidth]{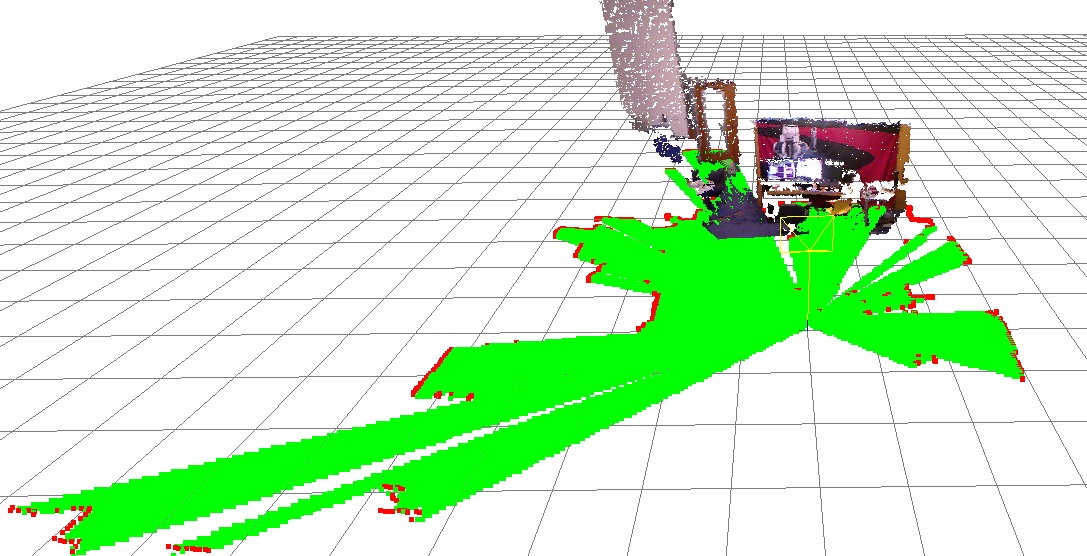}\label{}} \quad
\subfloat[Lidar (top view)]{\includegraphics[width=0.17\textwidth]{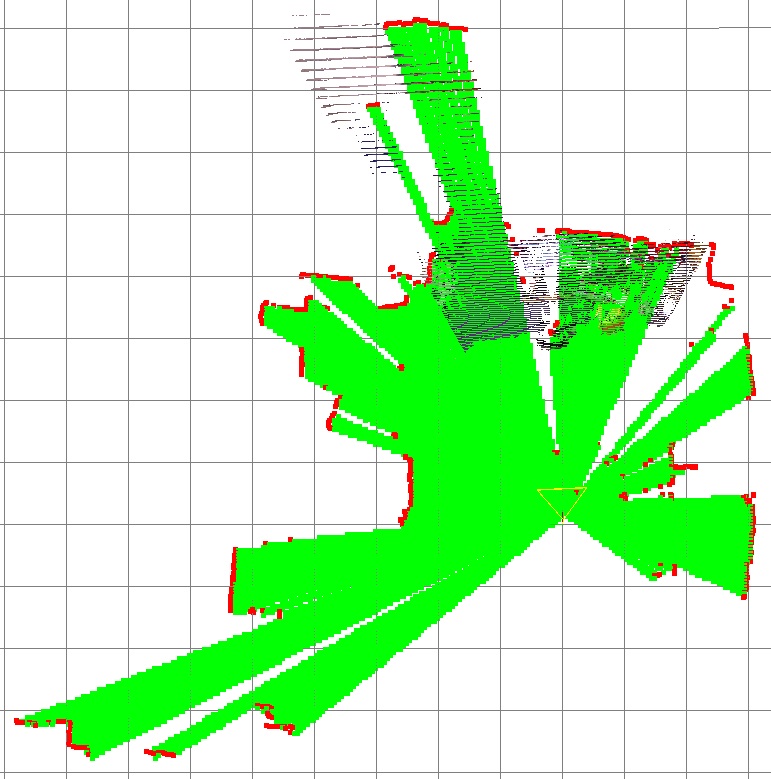}\label{}} \\
\subfloat[RGB-D camera view]{\includegraphics[width=0.2\textwidth]{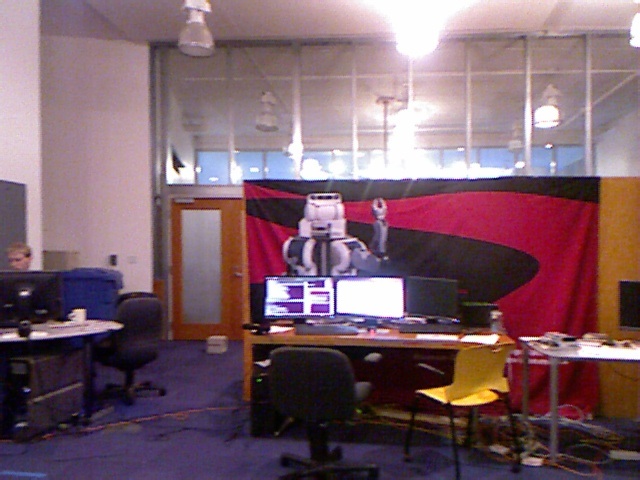}\label{}}  \quad
\subfloat[RGB-D segmentation]{\includegraphics[width=0.18\textwidth]{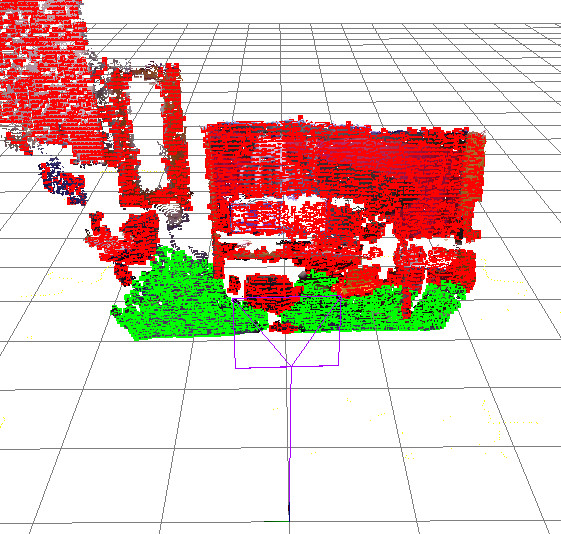}\label{fig:localgrids:seg}}  \quad
\subfloat[RGB-D projection]{\includegraphics[width=0.18\textwidth]{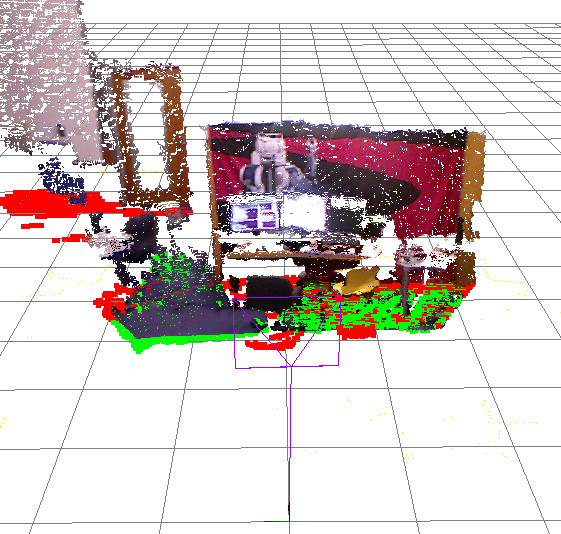}\label{}}  \quad
\subfloat[RGB-D projection with 2D ray tracing]{\includegraphics[width=0.18\textwidth]{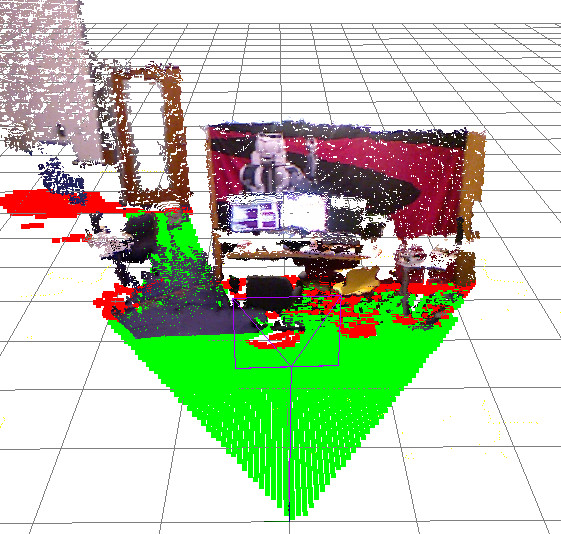}\label{}} \quad
\subfloat[RGB-D top view]{\includegraphics[width=0.15\textwidth]{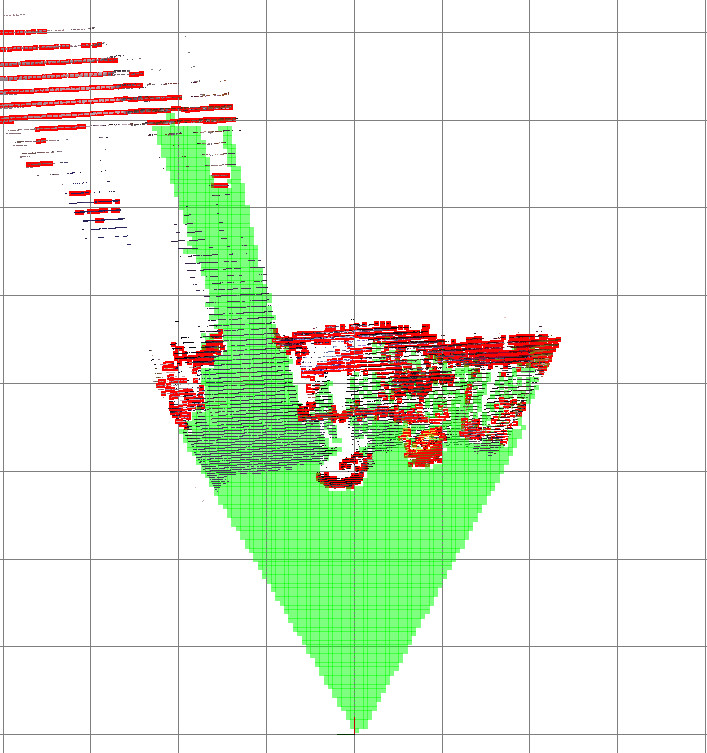}\label{fig:localgrids:rgbdtop}} \\
\subfloat[Stereo camera view]{\includegraphics[width=0.2\textwidth]{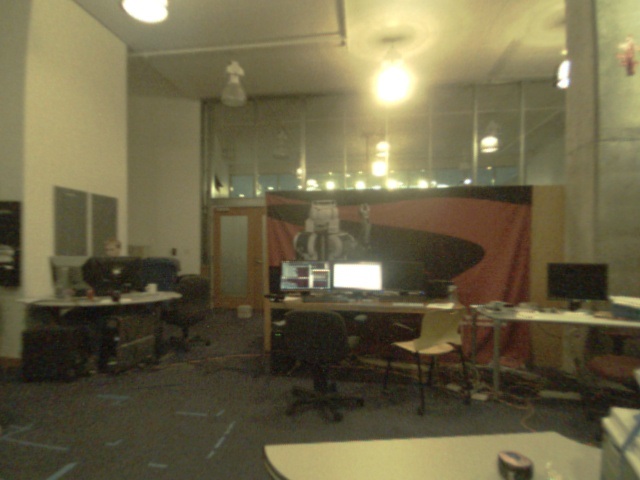}\label{}} \quad
\subfloat[Stereo segmentation]{\includegraphics[width=0.18\textwidth]{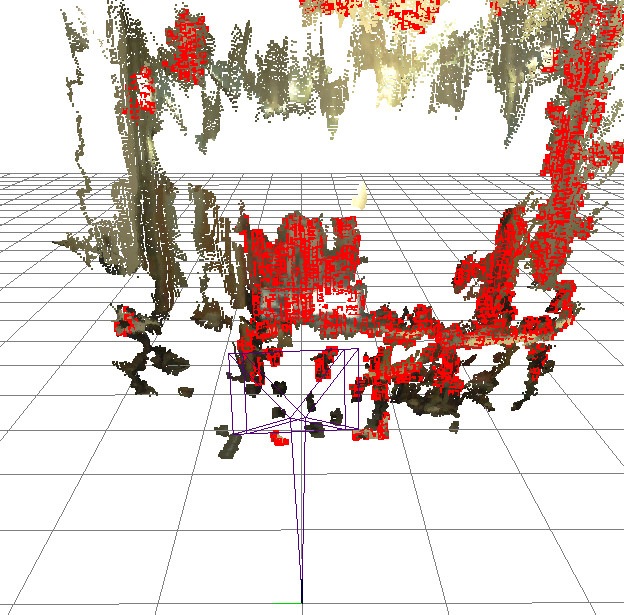}\label{fig:localgridsStereo:seg}} \quad
\subfloat[Stereo projection]{\includegraphics[width=0.18\textwidth]{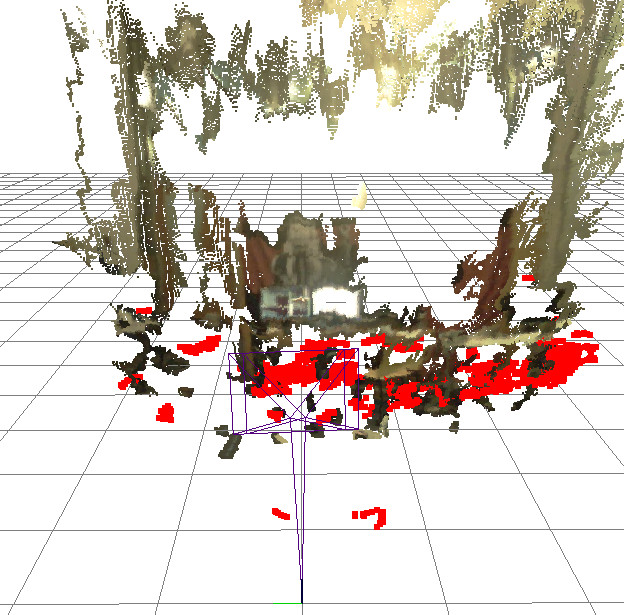}\label{}} \quad
\subfloat[Stereo projection with 2D ray tracing]{\includegraphics[width=0.18\textwidth]{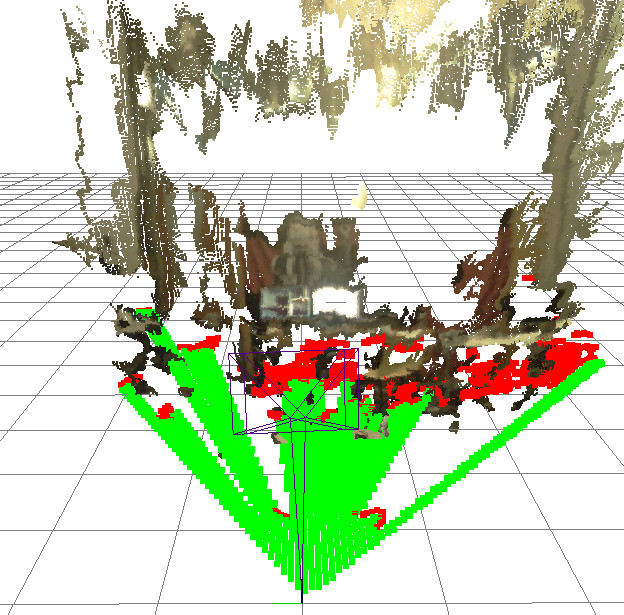}\label{}} \quad
\subfloat[Stereo top view]{\includegraphics[width=0.15\textwidth]{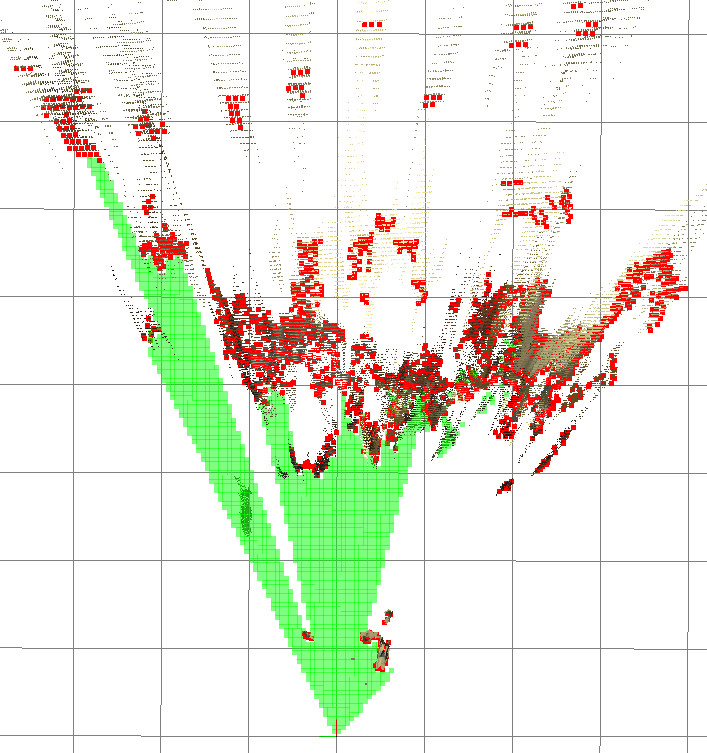}\label{}}
\end{tabular}
\caption{Local occupancy grid examples. For cameras, segmentation examples correspond to 3D local occupancy grids without ray tracing, then other ones are 2D local occupancy grids without or with 2D ray tracing. Obstacle cells are shown in red. Empty and ground cells are shown in green. The black grid is only a visual reference and has cell size of 1 m.}
\label{fig:localgrids}
\end{figure*}

\begin{figure*}[!t]
\centering
\begin{tabular}{cccc}
\subfloat[RGB-D camera view]{\includegraphics[width=0.17\textwidth]{41}\label{}}  \quad
\subfloat[Depth right side view]{\includegraphics[width=0.2\textwidth]{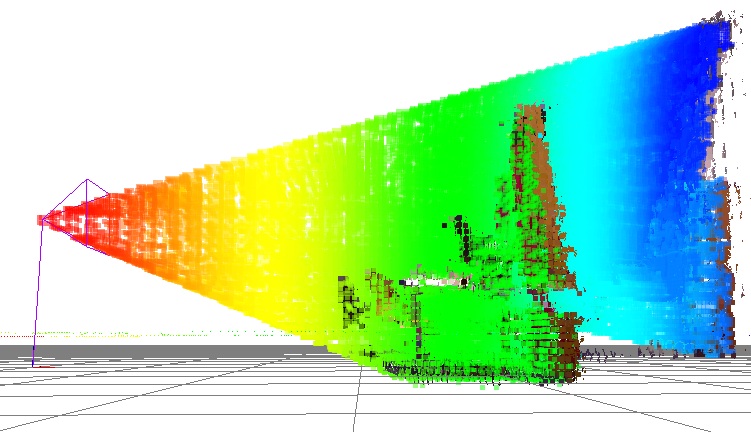}\label{}}  \quad
\subfloat[Depth top view]{\includegraphics[width=0.12\textwidth]{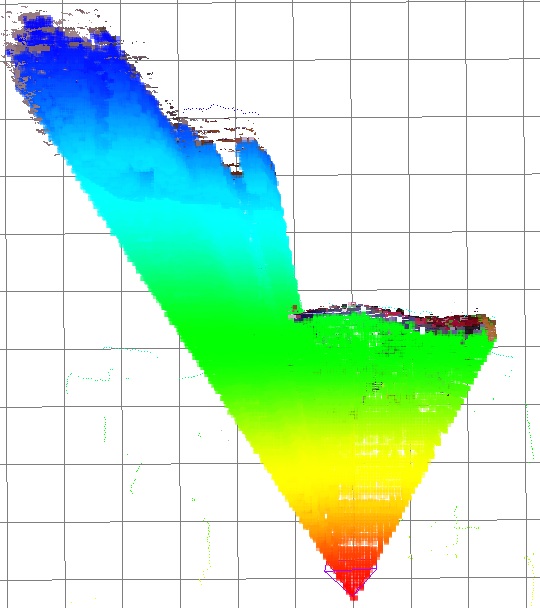}\label{}} \quad
\subfloat[Depth back view]{\includegraphics[width=0.2\textwidth]{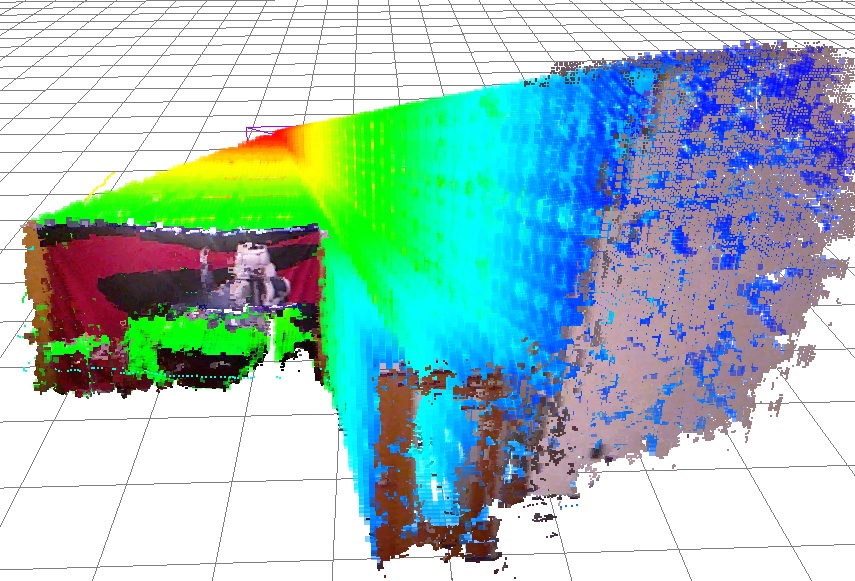}\label{}} \\
\subfloat[Stereo camera view]{\includegraphics[width=0.17\textwidth]{80}\label{}}  \quad
\subfloat[Stereo right side view]{\includegraphics[width=0.2\textwidth]{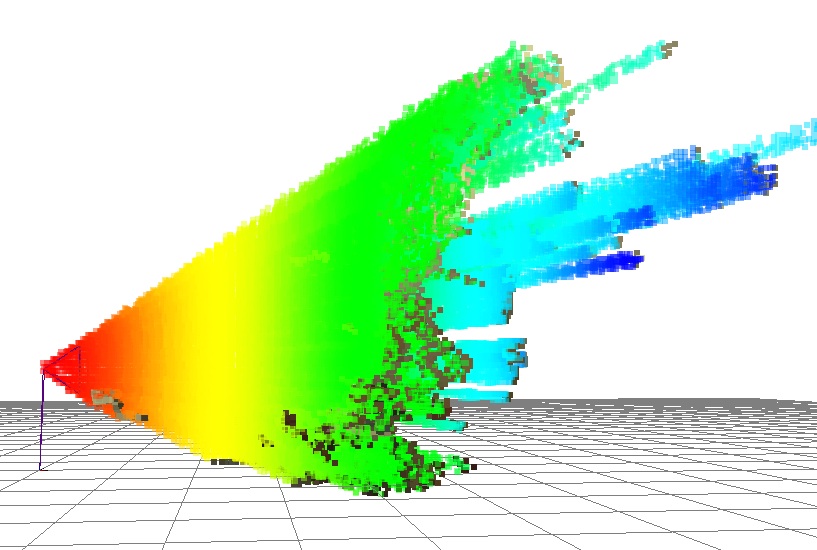}\label{}}  \quad
\subfloat[Stereo top view]{\includegraphics[width=0.12\textwidth]{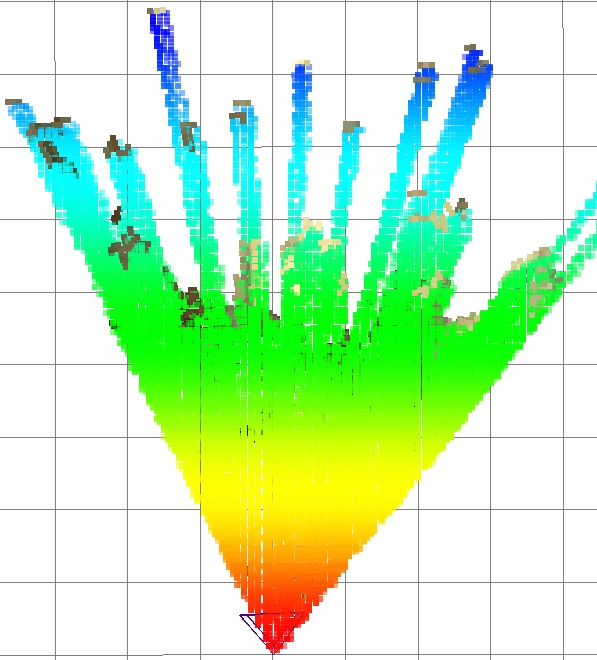}\label{}} \quad
\subfloat[Stereo back view]{\includegraphics[width=0.2\textwidth]{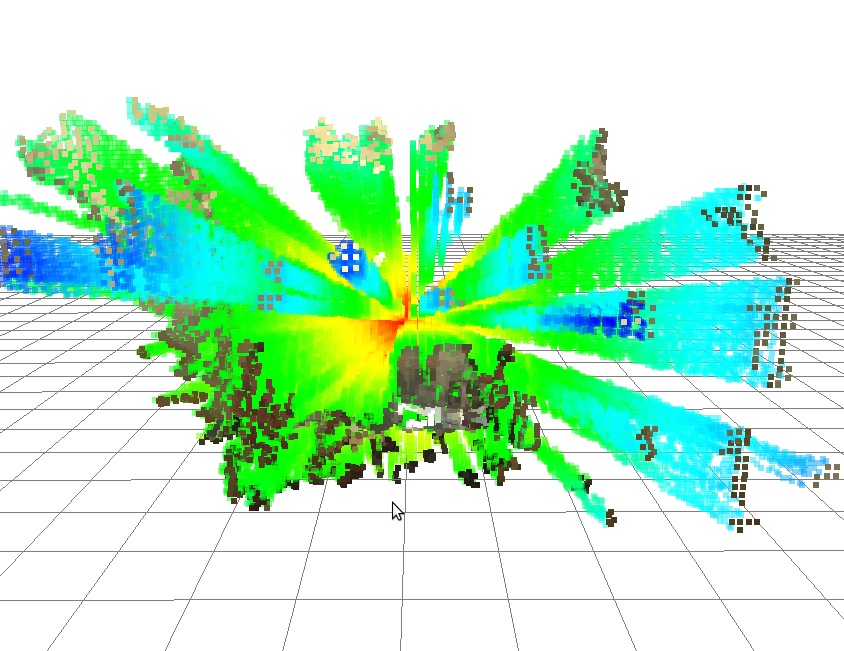}\label{}} \quad
\end{tabular}
\caption{3D local occupancy grid map with ray tracing examples}
\label{fig:local3dgrids}
\end{figure*}

\begin{figure*}[!t]
\centering
\begin{tabular}{cccc}
\subfloat[Short-range lidar]{\includegraphics[width=0.22\textwidth]{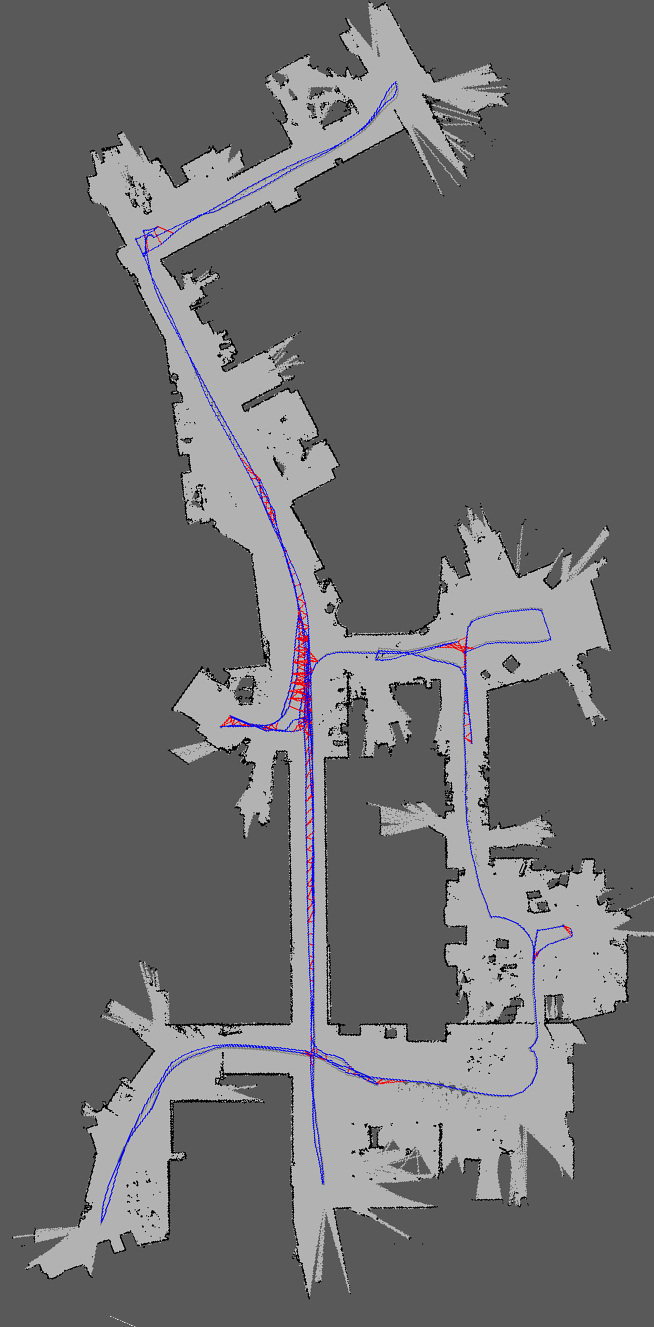}\label{global_grid_short}}  \hspace{0.1em}
\subfloat[RGB-D projection without ray tracing]{\includegraphics[width=0.22\textwidth]{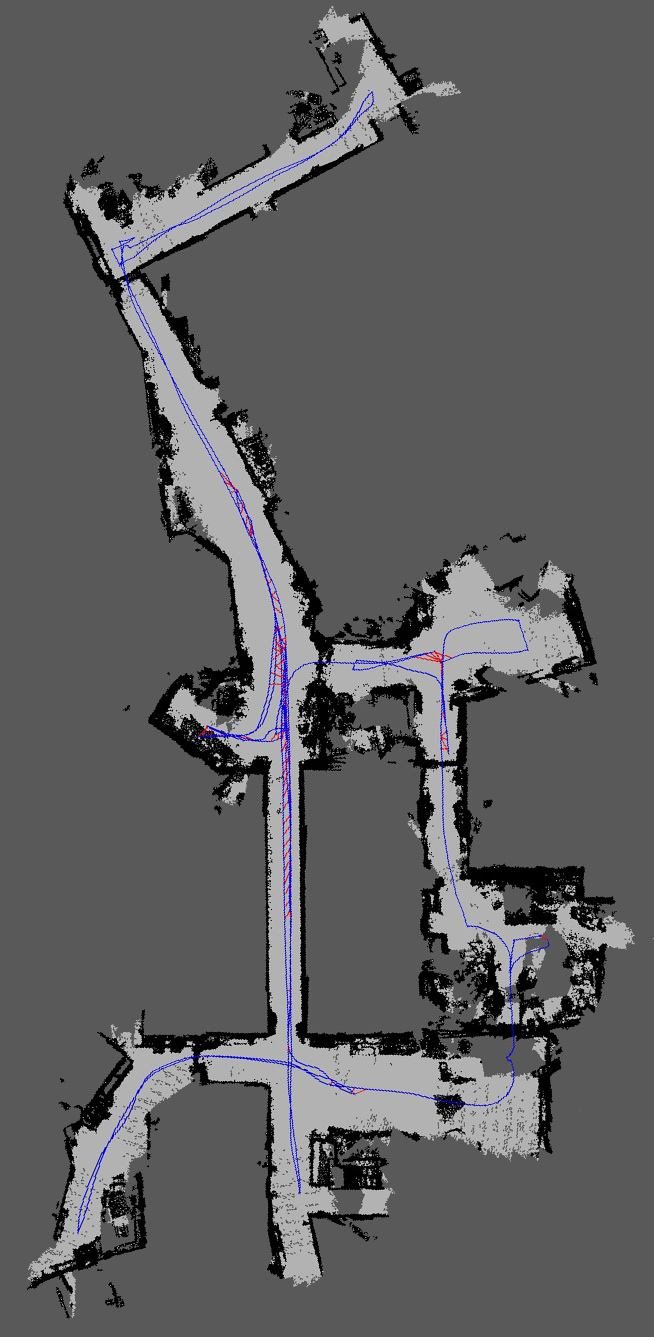}\label{}}  \hspace{0.1em}
\subfloat[RGB-D projection with 2D ray tracing]{\includegraphics[width=0.22\textwidth]{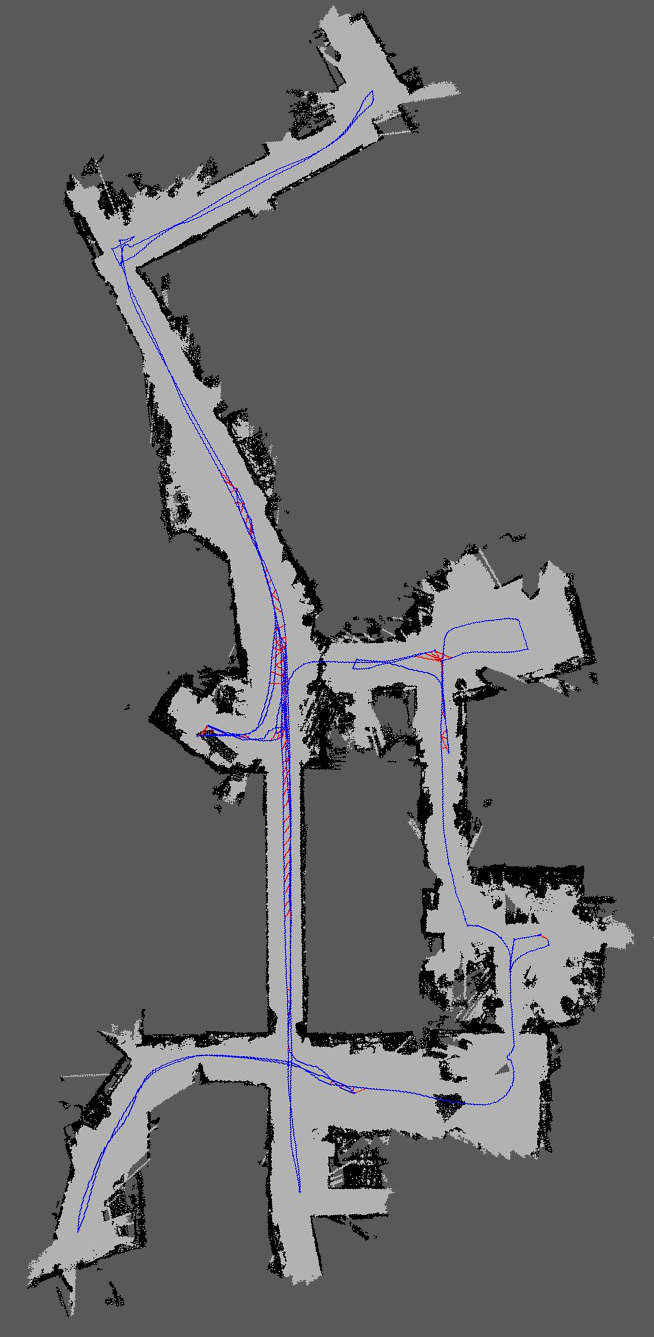}\label{}} \hspace{0.1em}
\subfloat[RGB-D OctoMap projection with 3D ray tracing]{\includegraphics[width=0.22\textwidth]{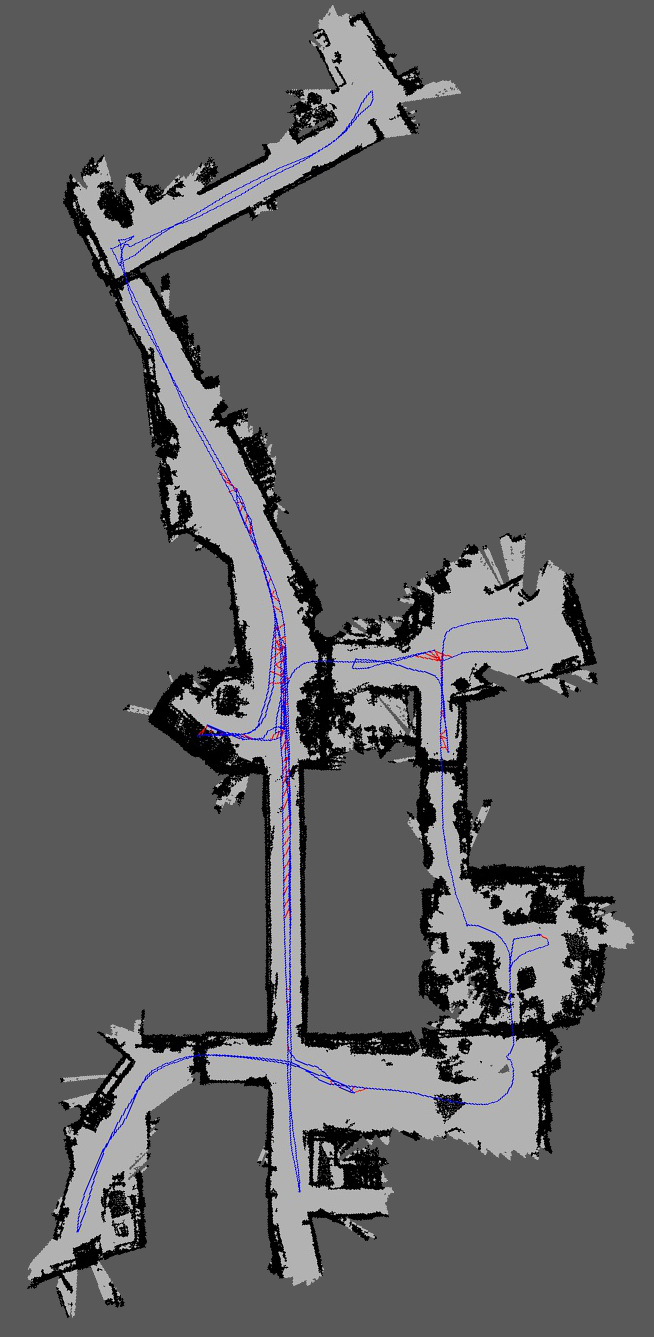}\label{}} \hspace{0.1em} \\
\subfloat[Long-range lidar]{\includegraphics[width=0.25\textwidth]{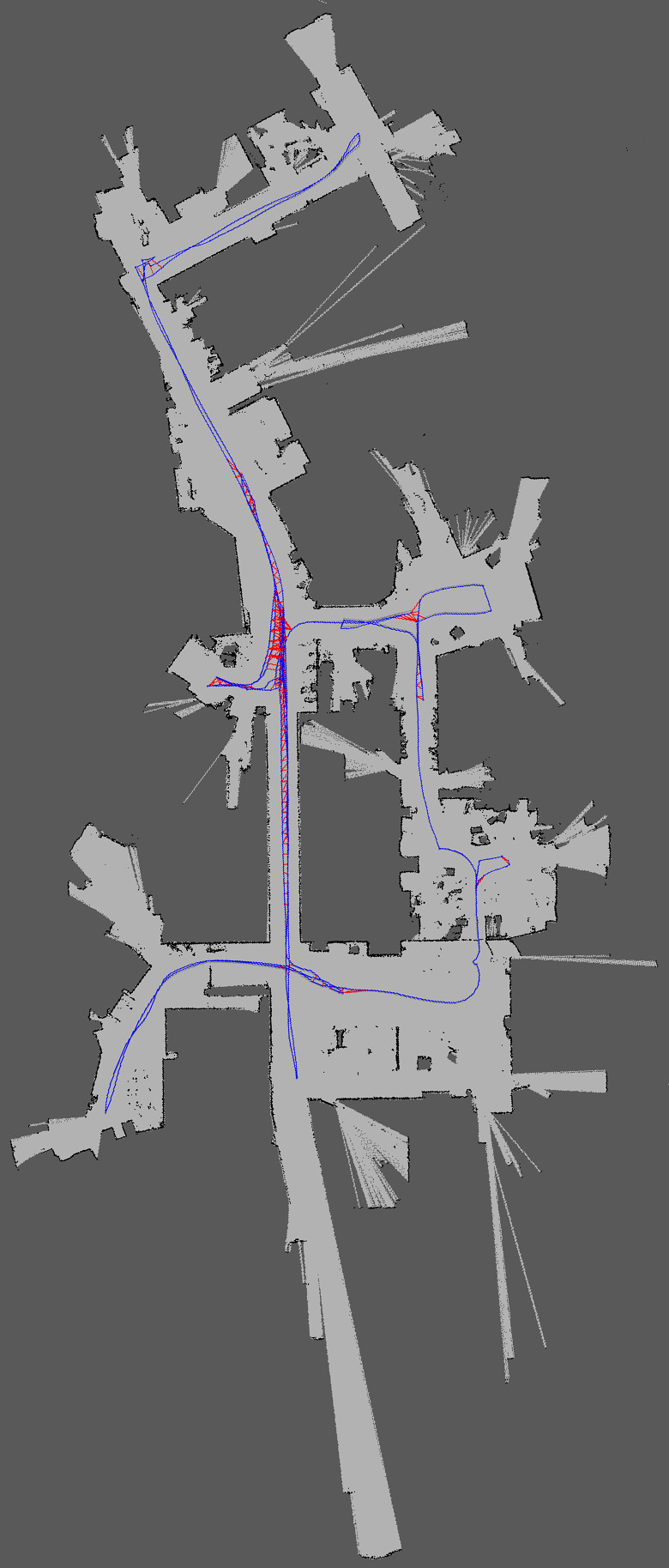}\label{global_grid_long}}  \hspace{0.1em}
\subfloat[Stereo projection without ray tracing]{\includegraphics[width=0.22\textwidth]{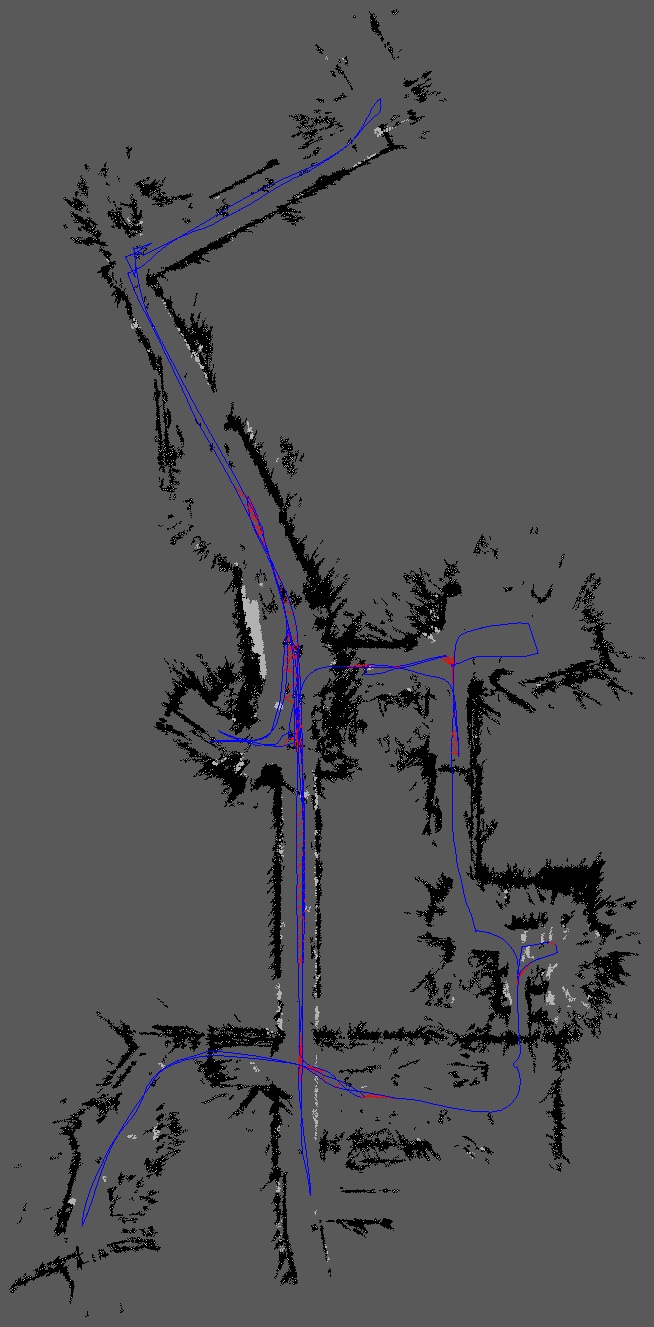}\label{}}  \hspace{0.1em}
\subfloat[Stereo projection with 2D ray tracing]{\includegraphics[width=0.22\textwidth]{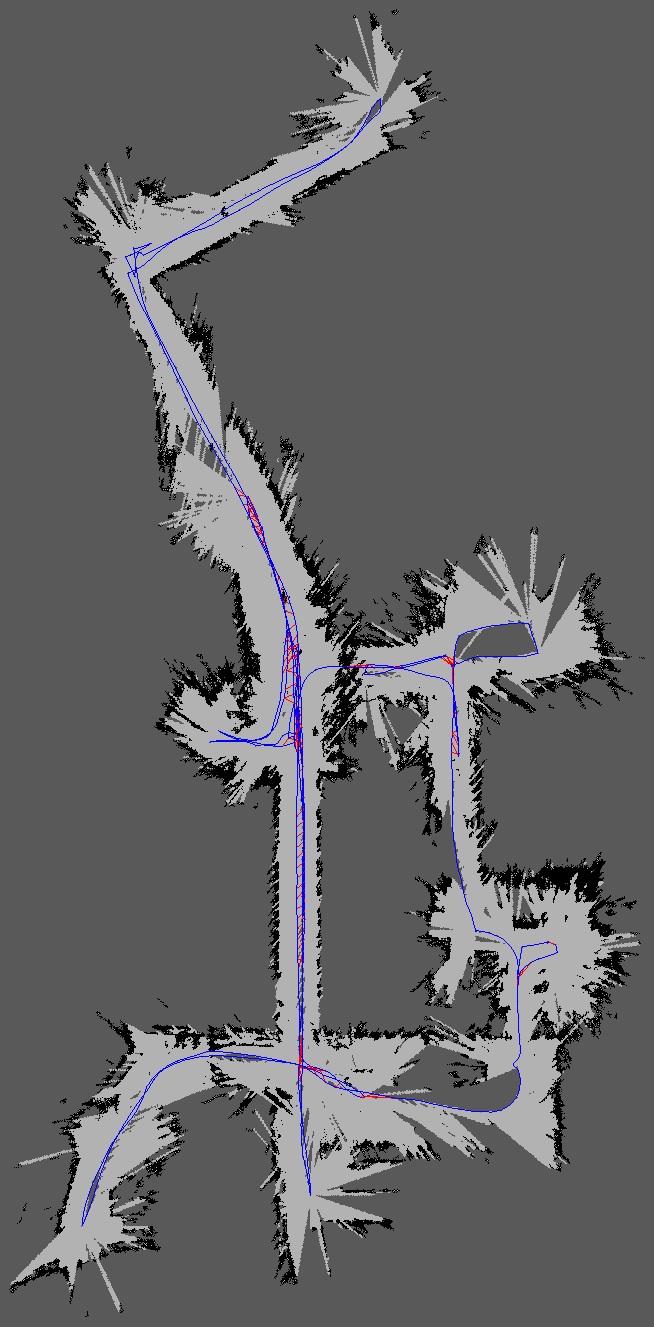}\label{}} \hspace{0.1em}
\subfloat[Stereo OctoMap projection with 3D ray tracing]{\includegraphics[width=0.22\textwidth]{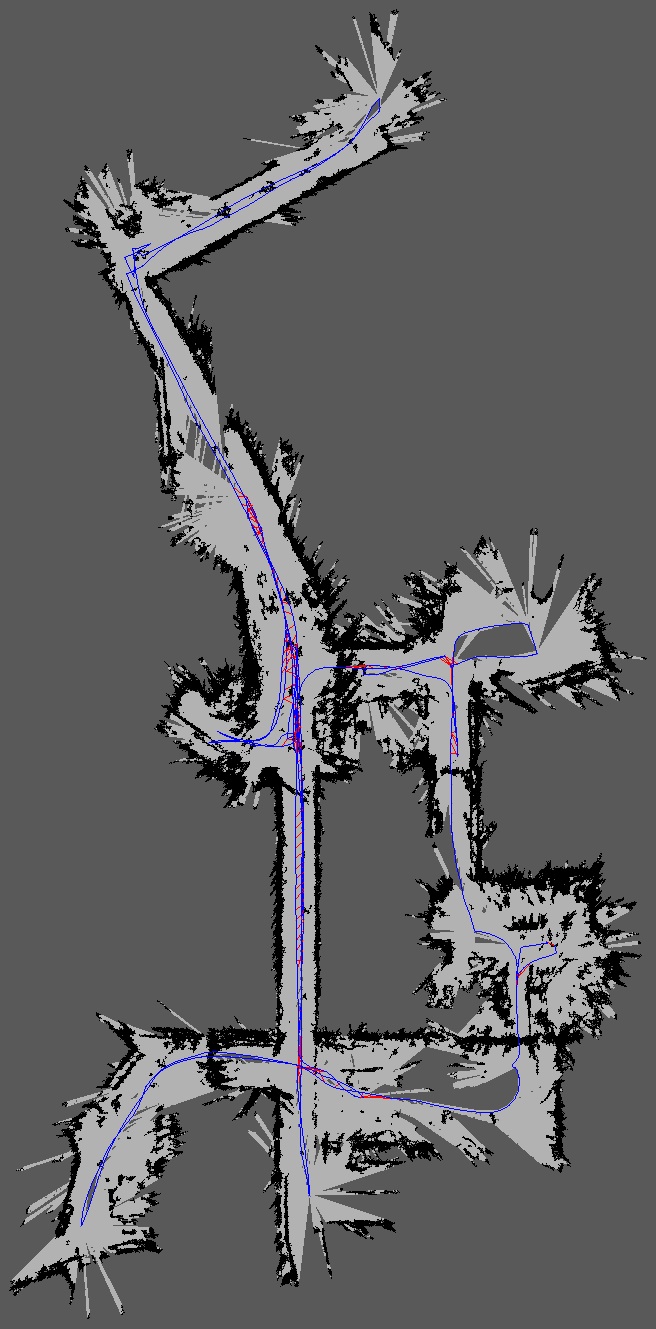}\label{}} \hspace{0.1em}
\end{tabular}
\caption{2D occupancy grid map examples.}
\label{fig:grids}
\end{figure*}

\begin{figure*}[!t]
\centering
\begin{tabular}{cc}
\subfloat[]{\includegraphics[width=0.4\textwidth]{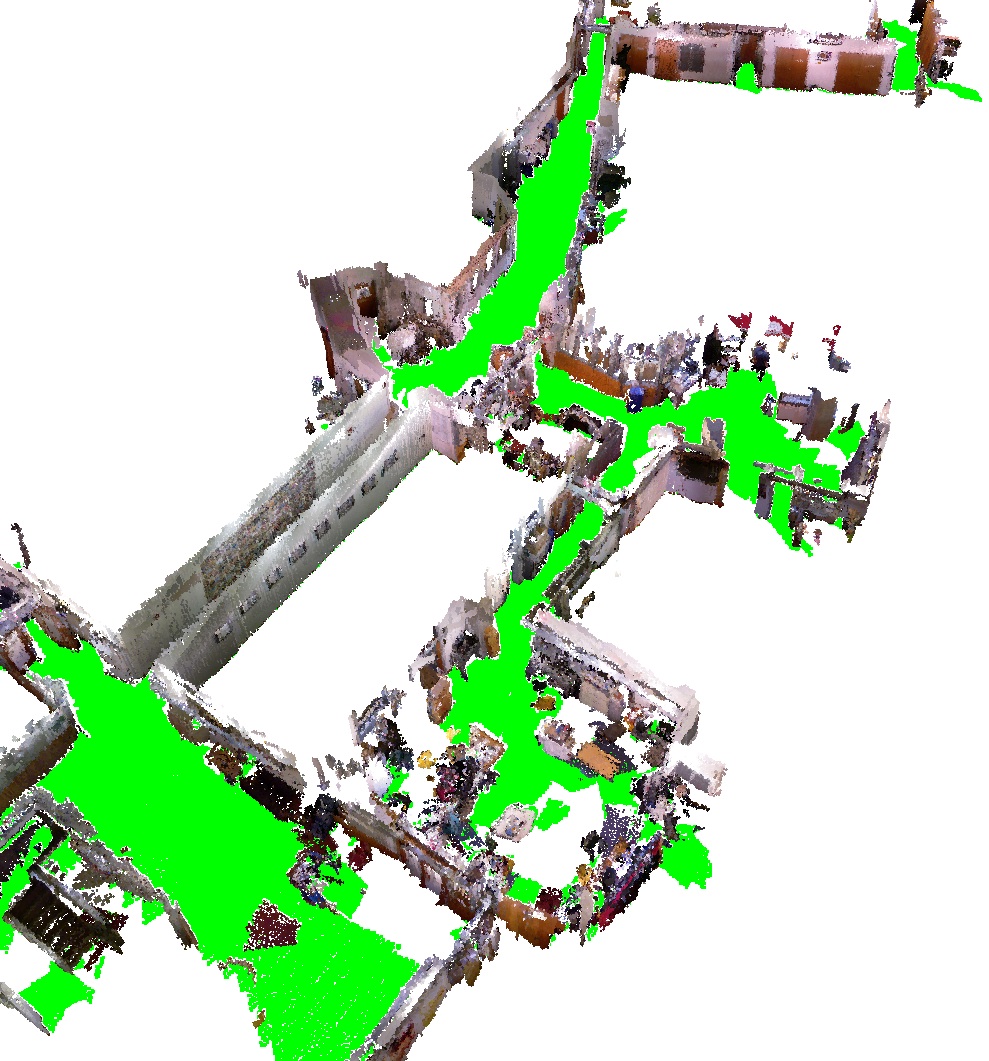}\label{}}
\subfloat[]{\includegraphics[width=0.4\textwidth]{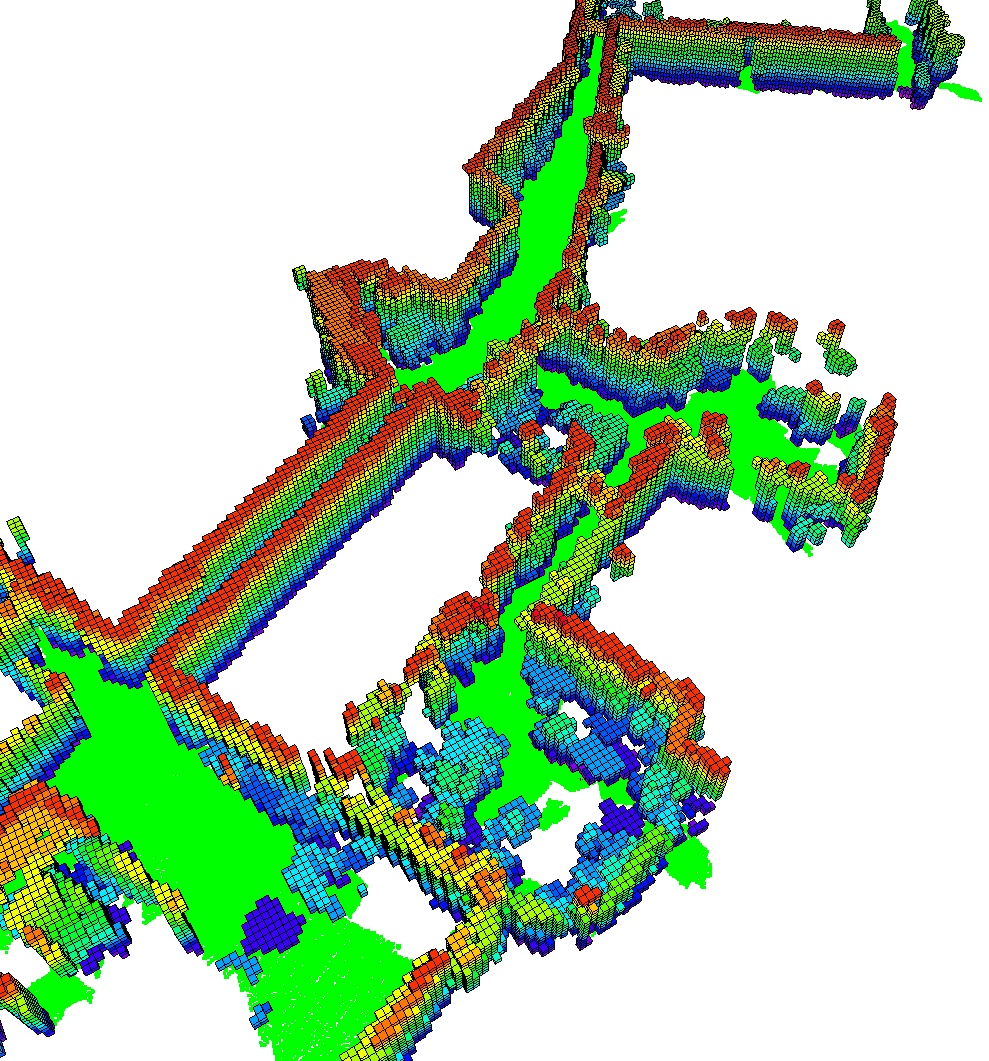}\label{}}
\end{tabular}
\caption{OctoMap of depth a) 16 and b) 14 using the RGB-D camera.}
\label{fig:compare_octomap}
\end{figure*}

Figure \ref{fig:localgrids} presents examples of generation of local 2D occupancy grids depending on the sensor and approach used. 
Lidar-based grid provides a larger field of view, but only obstacles at the height of the lidar can be detected. 
In comparison, RGB-D based grid can detect some obstacles that lidar cannot, like the yellow chair, making navigation safer in this kind of environment. 
However, as shown in the top left of Figure \ref{fig:localgrids:rgbdtop}, obstacles detected by RGB-D camera after 5 m lack accuracy. 
Stereo camera can detect most of the obstacles as long as they are textured. 
It has a larger field of view than a RGB-D camera, detecting the table just in front of the robot while the RGB-D camera cannot see it. 
Note that with this 2D ray tracing approach and because the cameras are not close to ground, if the robot is approaching the table from the front so that the cameras do not see it anymore, the obstacles previously added to global occupancy grid map because of the table will be incorrectly cleared. 
To use 2D ray tracing, it is preferred to lower the camera at the height of the smallest obstacle in the environment. 
If lowering the camera is not possible because of some robot physical constraints, using 2D local occupancy grids without ray tracing would then be safer, and dynamic obstacles would only be cleared if the camera can see the ground where the obstacle was (as in Figure \ref{fig:localgrids:seg}). 
Note that stereo camera cannot see the ground (see Figure \ref{fig:localgridsStereo:seg}), so dynamic obstacles cannot be cleared without ray tracing. 
Another solution to solve the problem of safe obstacle clearing is to use 3D local occupancy grids with ray tracing and OctoMap, at the cost of significantly increasing the processing power required. 
Figure \ref{fig:local3dgrids} illustrates examples of 3D ray tracing with RGB-D and stereo cameras. 
As depth images are more dense than disparity images, ray tracing takes more time to do using RGB-D camera. 
However, as stereo cameras have a larger field of view, more volume would be filled, creating larger local occupancy grids.  
This explains why updating 3D global local occupancy grids from stereo camera takes more time than from RGB-D cameras. 
If a 3D occupancy grid is required and the environment is static (no dynamic obstacles to clear), avoiding ray tracing can help save a lot of computing resources.

Figure \ref{fig:grids} presents the corresponding 2D global occupancy grids created for the different sensors and local occupancy grid approaches of Table \ref{stata_occupancy}. 
Lidar-based maps give the most accurate geometry of the environment (at the height of the lidar), followed by maps created from a RGB-D camera. 
Without ray tracing, stereo-based map contains almost no empty cells because the disparity approach cannot see texture-less ground. 
Some walls are also not detected or very noisy. 
Using ray tracing, empty space can be filled. 
When comparing 2D ray tracing against 3D ray tracing, it is possible to see that some obstacles were incorrectly cleared using 2D ray tracing: beside the doors, the environment is considered static so obstacles should not have been cleared. 
For 3D ray tracing, when opening the door, if the field of view of the camera cannot see the whole opening (e.g., camera cannot see the bottom of the door), it will be able to clear only the volume of the door it can see, leaving the bottom as obstacle.
Figure \ref{fig:compare_octomap} presents the 3D occupancy grid of the OctoMap at tree depth 16 and 14 using the RGB-D camera. 
Tree depth 16 corresponds to cell size of 5 cm and shown with RGB color. 
Generating the OctoMap at lower tree depth increases cell size (lower resolution), which can be useful for faster path planning.

\subsection{Examining the Use of RTAB-Map's Memory Management Mechanism}
\label{sec:mem}

For large-scale and long-term SLAM where the graph is constantly adding new nodes, these previous solutions to adjust computation load based on occupancy grid type may not be sufficient. 
In all previously described experiments, RTAB-Map's memory management mechanism was disabled to always have access to global map for trajectory accuracy and occupancy grid comparisons. 
To outline how much time is required for each module of RTAB-Map's WM in Figure \ref{fig:rtabmap} in a large scale environment, the two MIT Stata Center sequences were played back to back, creating a long mapping experiment containing two mapping sessions linked together. 
Both sequences start and finish at the same location, so a loop closure between the end of the first sequence and the beginning of the second sequence can be detected when playing the second bag, merging automatically the two maps. 
As the ground truth coordinates between the two bags are slightly off, the ground truth of the second bag is transformed in the same coordinates than the ground truth of the first bag. 
To do so, we assembled the scans for each sequence separately using their respective ground truths, then using \textit{pcl\_icp} tool\footnote{\url{https://github.com/PointCloudLibrary/pcl/blob/master/tools/icp.cpp}}, the two point clouds are registered and the resulting transformation $(x,y,\theta)=(0.006236$ m, $-0.351500$ m, $-0.017832$ rad$)$ can be applied to the ground truth of the second bag.
RTAB-Map's update rate ``Rtabmap/DetectionRate" is also increased to 2 Hz to add twice the nodes to the graph, with ``Mem/STMSize" set back to 30 so that nodes stay the same time in STM than at 1 Hz. 
WM is limited to a maximum size ``Rtabmap/MemoryThr" of 300 nodes to better observe the effect of memory management when a low number of nodes is kept in WM. 
RTAB-Map's odometry configuration used is WheelIMU\textrightarrow S2M with short-range lidar. 
From Table \ref{stata_rmse}, while accuracy results are slightly worst than with long-range version, using short-range lidar requires significantly less time to regenerate the global occupancy grid as shown in Table \ref{stata_occupancy} for a similar quality (Figure \ref{global_grid_short} versus Figure \ref{global_grid_long}). 

\begin{figure*}[!t]
\centering
\begin{tabular}{cc}
\subfloat[]{\includegraphics[width=0.5\textwidth]{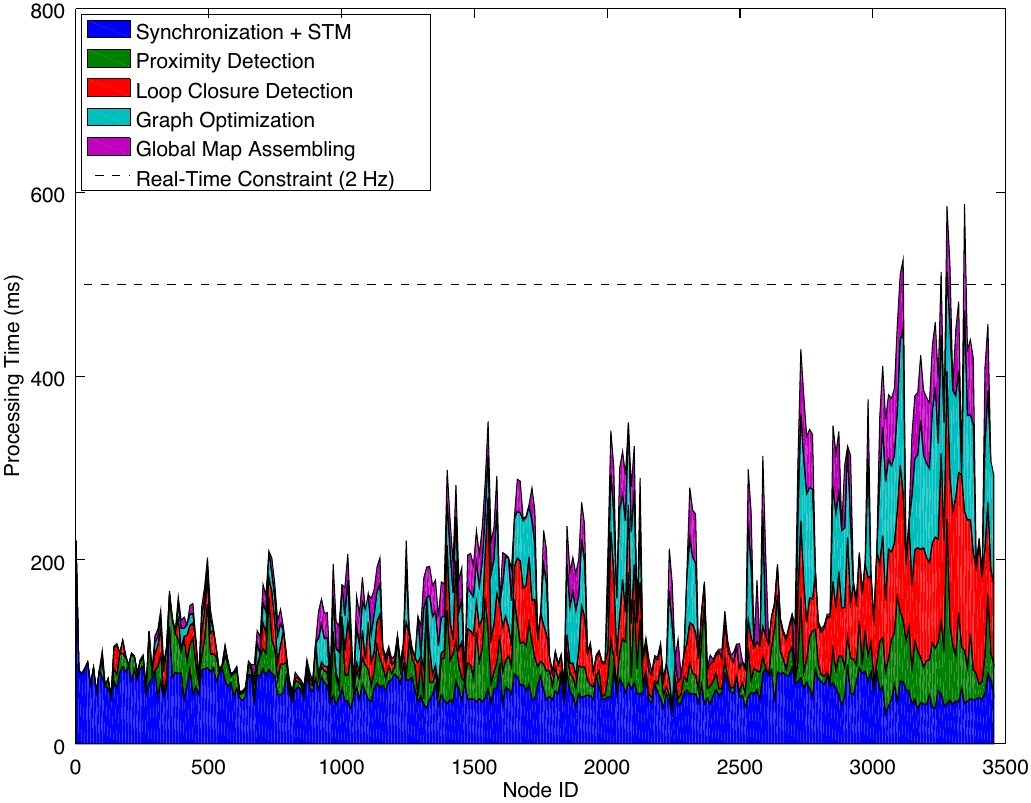}\label{}}
\subfloat[]{\includegraphics[width=0.5\textwidth]{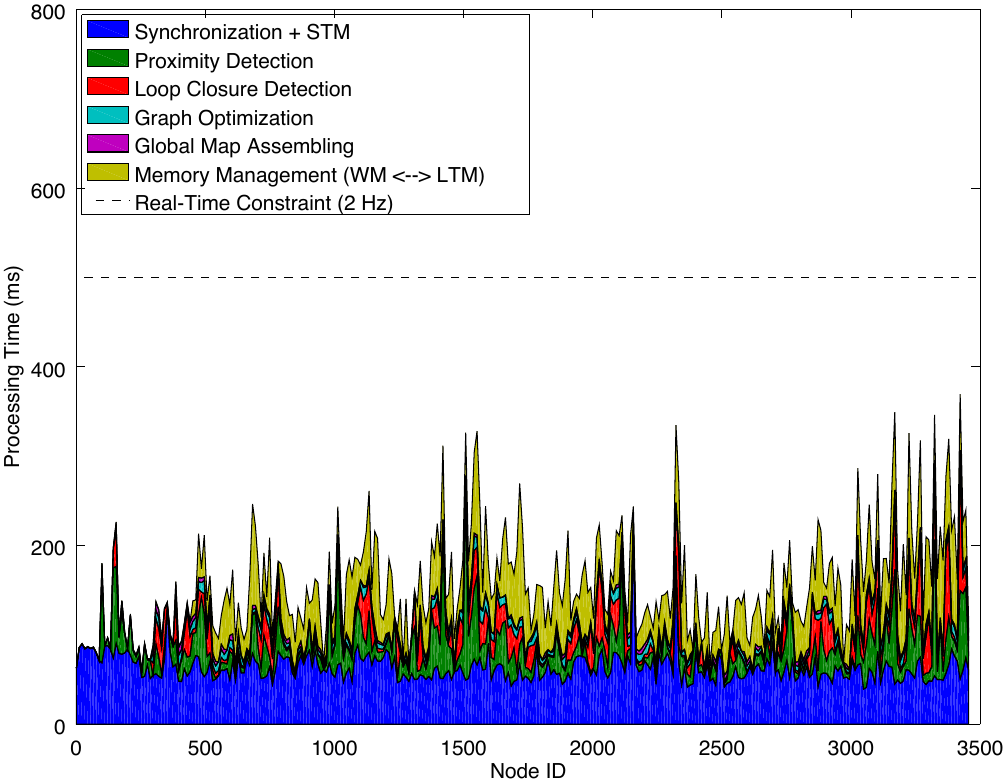}\label{fig:timeb}}
\end{tabular}
\caption{Processing time required for each module inside \textit{rtabmap} ROS node without (a) and with (b) memory management, for a map update rate of 2 Hz using the combined sessions of the MIT Stata Center sequences.}
\label{fig:time}
\end{figure*}

\begin{figure*}[!t]
\centering
\begin{tabular}{c}
\subfloat[]{\includegraphics[width=0.79\textwidth]{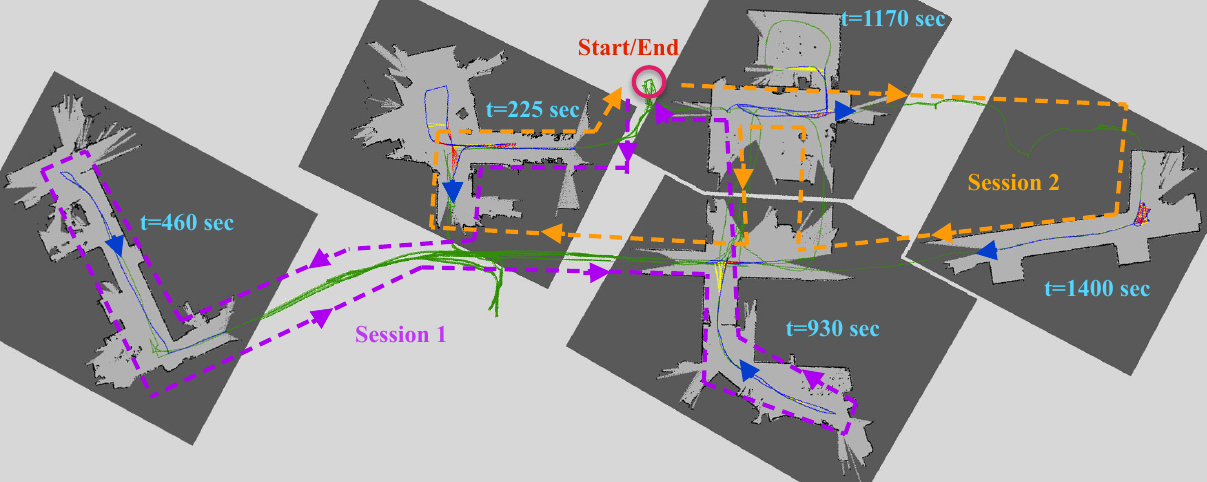}\label{fig:globalmaps:memsplit}}  \\
\subfloat[]{\includegraphics[width=0.79\textwidth]{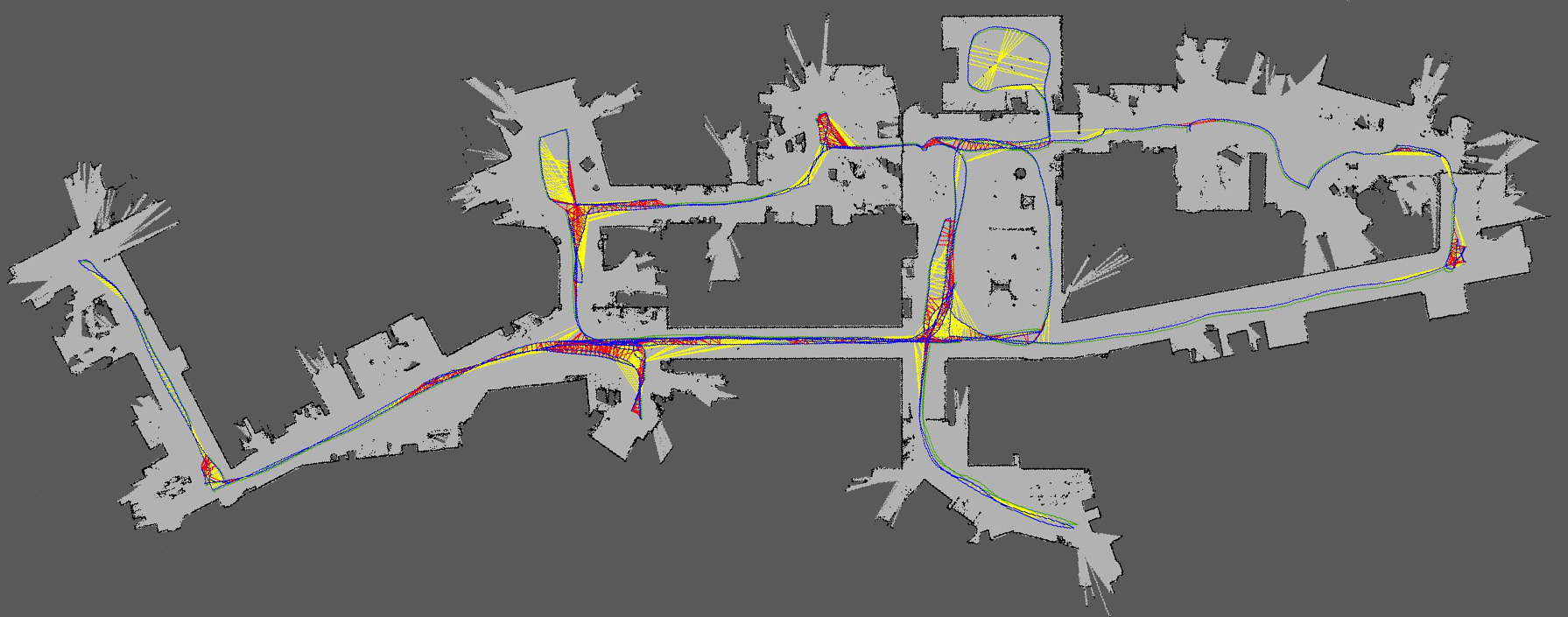}\label{fig:globalmaps:fullmem}}  \\
\subfloat[]{\includegraphics[width=0.79\textwidth]{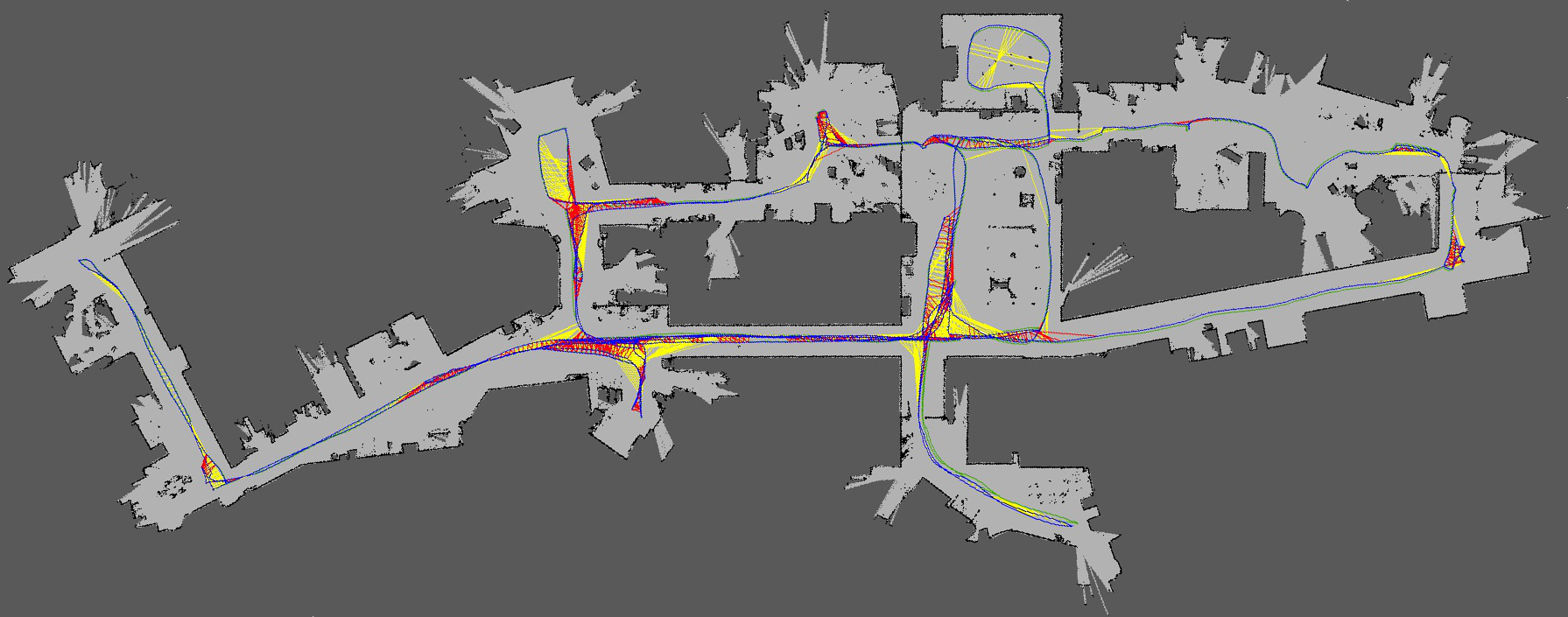}\label{fig:globalmaps:fullnomem}} 
\end{tabular}
\caption{Global maps created with (a,b) and without (c) memory management using the combined sessions of the MIT Stata Center sequences. 
In (a), five examples of the biggest map created online using available nodes in WM at the specified time are presented, with the blue triangle as the robot pose at that time. 
The global trajectories done during each session are shown in purple and orange, respectively. 
Neighbor, loop closure and proximity links are shown in blue, red and yellow, respectively. 
The green path is the ground truth, which is almost not visible in (b) and (c) as the blue line is superposed.}
\label{fig:globalmaps}
\end{figure*}

Figure \ref{fig:time} presents the timing results without (a) and with (b) memory management for RTAB-Map WM modules presented in Figure \ref{fig:rtabmap}. 
The horizontal line is the real-time constraint, i.e., the maximum time allowed for addition of new nodes to map at 2 Hz. 
In that case, RTAB-Map's WheelIMU\textrightarrow S2M without memory management is not satisfying real-time constraints, because some updates require more than 500 msec. 
Between nodes 3000 and 3500, the robot is revisiting an area which has been already visited multiple times (e.g., the beginning and ending areas of each sequence), triggering many more loop closure and proximity detections with previously mapped paths. 
As the map increases in size, the update time increases, which creates large holes in the map if there are no new nodes added during a number of seconds if the robot is moving. 
With memory management, RTAB-Map's WheelIMU\textrightarrow S2M is able to satisfy real-time constraints for the whole experiment. 
Memory management adds a small overhead (average of 52 msec) when moving nodes between WM and LTM, but it greatly reduces the processing time required for other modules depending on graph size. 
However, the global occupancy grid map does not always represent the full environment visited. 
Figure \ref{fig:globalmaps:memsplit} shows five global occupancy grid examples at different time in the experiment using memory management. 
The blue triangle indicates the robot pose at the referred time. 
At $t=225$ sec, $t=1170$ sec and $t=1400$ sec, the robot is moving to a new area. 
At $t=460$ sec and $t=930$ sec, the robot is revisiting a previously mapped area, explaining why there is an area of the map in front of the robot. 
When revisiting a previously visited area, memory management retrieves nodes from LTM to WM to expand the current map with old locations. 
While the online global occupancy grid map is limited around the current pose of the robot, using its memory management mechanism, RTAB-Map is still able to detect most of the loop closures needed to correctly merge the two sessions. Without memory management, there are 2048 loop closure links and 1129 proximity links, in comparison to 1256 loop closure links and 774 proximity links with memory management. For instance, Figure \ref {fig:globalmaps:fullmem} and Figure \ref{fig:globalmaps:fullnomem} are the global occupancy grid maps created from the experiments with and without memory management, respectively. 
The map in (b) is generated after online mapping using all links saved in LTM. 
Both experiments give the same final ATE of 12 cm.

%% file: 5_discussion.tex
Trajectory performance evaluation presented in Section \ref{sec:slamperf} demonstrates that integrating odometry approaches in our extended version of RTAB-Map can lead to results comparable to state-of-the-art visual-based and lidar-based SLAM approaches, making it a powerful library for the design and prototyping SLAM with different sensors. 
To our knowledge, using our extended version of RTAB-Map, this paper is the first to report such experimental comparison of lidar versus visual-based SLAM configurations on the same system.  
While being also ROS ready for 2D and 3D online autonomous navigation, this makes the approach easy to integrate to a custom robot in order to compare live the differences between visual and lidar SLAM configurations. It is often difficult to compare these SLAM configurations when they have been tested on datasets that only work with their sensor type and derived simply by teleoperating the robot or by having a human positioning the sensor. Some configurations also require that the sensor moves in a specific way to compensate for their limitations. 
These live comparisons can then help to reveal flaws and limitations inherent to the sensor chosen when combined with standard navigation approaches like ROS' navigation stack \cite{officemarathon}.

Consequently, RTAB-Map can be used to conduct trials with different sensors and identify early on if a sensor is suitable for the targeted application. 
Based on the results presented in this paper, guidelines can be derived regarding when using SLAM (without external global localization) in an indoor environment. 
Unless a long-range lidar is used, having odometry input from proprioceptive sensors (e.g., IMU, wheel encoders) is mandatory for robust autonomous navigation. 
When relying only on short-range sensors, it is very likely that the robot will end in an area where the sensor cannot see enough features to be able to localize itself in its map. 
For cameras, seeing a white wall, a textureless or a dark area would result in loosing localization. 
For short-range lidar, a large empty space or a long corridor with low geometry complexity can be also problematic. 
Both kind of sensors have then their issues depending on the environment. 

In RTAB-Map, motion estimation during localization or loop closure detection is done primarily visually, then optionally refined by geometry if a lidar is available. 
This means that if the visual motion estimation fails, lidar motion estimation cannot be done. 
In future work, tighter coupling of visual and geometry estimations could be evaluated so that if one fails, the other can still be used to get an estimation of the position. 
This also applies to odometry, where a visual-lidar approach could be more robust when environments are textureless or lacking geometry. 
For loop closure detection, the current bag-of-words approach is dependent on a camera, meaning that a camera is always required even if lidar SLAM is done. 
As a solution, it is possible to feed a fake empty image to RTAB-Map if the robot does not have a camera, relying only on proximity detection for map corrections. 
As long as the robot is not drifting too much and the environment is relatively small (e.g., a single building), proximity detection could detect most of the loop closures without needing a camera. 
For large-scale loop closure detection where pose estimation cannot be used robustly, a lidar-based loop closure detection could be integrated (similar to \cite{Bosse08} and \cite{hess2016real}) so that the robot can detect very large loop closures only using a lidar (even in completely dark environments).

Based on the results, general observations can be made regarding sensor choice for indoor navigation. 
While stereo cameras give slightly better localization accuracy, RGB-D cameras are preferred because textureless surfaces can be detected for obstacle avoidance. 
For all exteroceptive sensors based on light, navigation in environments filled with glass and reflective objects can be unsafe, as obstacles cannot be detected or false obstacles will be added to the map. 
In term of cost, using a RGB-D camera can be beneficial compared to a lidar. 
However, the extra field of view of the lidar is a huge advantage over a single RGB-D camera when it comes to low-drift navigation, as shown in Section \ref{sec:mit} with lower ATE\textsubscript{max}. 
One could add more RGB-D cameras to get a field of view similar to the lidar, but the increase in multi-camera calibration complexity and computational load do not justify the cost for 2D navigation robustness unless 3D obstacles must be detected, as a low-cost lidar can do the job using less computing resources. 
However for 3D navigation (e.g., drone), having a multi-camera setup is beneficial to get a larger field of view compared with expensive 3D lidars (e.g., Velodyne) in term of cost.
In our use of RTAB-Map on real robots, depending of the projects and the target cost, we had to deal with the above limitations of the sensors used. 
One example is a patrolling robot doing continuously SLAM while autonomously navigate in the environment (\cite{labbe2017}). 
The robot was equipped with a RGB-D camera, a short-range lidar and wheel odometry. 
The environment was mostly long textureless corridors, making it difficult to localize visually and also geometrically. 
Visual loop closures could only be found at the end of corridors or in the rooms. 
The use of wheel odometry was then mandatory. 
Proximity detection helped alignment with the corridor using geometry in almost any directions (the lidar had more than 180$^\circ$ of field of view). 
The lidar's large field of view also helped during navigation when the robot had to avoid an obstacle, allowing it to localize even if the robot was oriented differently during the mapping.
Another example is its use on a low cost autonomous wheelchair using only a RGB-D camera for SLAM and navigation \cite{burhanpurkar-2017}. 
To handle textureless corridor environments, wheel odometry was used instead of visual odometry (or visual inertial odometry) to avoid getting lost as soon as entering a corridor. 
The limited field of view of the front facing RGB-D camera was also a problem during navigation. 
If the robot did not follow a very similar path than the one done previously when mapping the environment (e.g., to avoid someone passing by), the robot could get lost as loop closures or localizations could not be detected afterward. 
The retrofitted wheel odometry was drifting more on this platform than the previous robot (in particular if the planner was sending many commands making often the robot turn in place), increasing the problem of relocalizing after the robot avoided an obstacle. 
This justifies why the ATE\textsubscript{max} metric presented in this paper is important for navigation: the lower the odometry drifts, the faster the localization recovery happens after the robot  changes course for some reasons and has to come back to follow the original planned path. 
In these application examples using short-range sensors, clearing dynamic obstacles (after they moved) would not always be possible using the current ray tracing approach if the sensor rays could not ``hit'' something behind where the obstacles were in order to clear the space, keeping some fake obstacles on the map that can affect planning afterward. 
Similar to \cite{Barsan2018ICRA}, a smarter understanding of the images or laser scans could be implemented to segment dynamic objects from the static environment.

%% file: 6_conclusion.tex
This paper presents the extended version RTAB-Map, which provides a full integration with ROS to handle robot's \textit{tf}, to synchronize RGB-D, stereo, laser scan and point cloud topics, and the ability to generate occupancy grids for all sensors. 
As a result, RTAB-Map is now a multi-purpose graph-based SLAM approach that can be used out-of-the-box by novice SLAM users and for prototyping on robot platforms with different sensor configurations and processing capabilities. 
It can be used to compare performance over datasets and to conduct online evaluations. 
Sensors required for SLAM, whether they are low cost or expensive, all have limitations that influence localization accuracy, map quality and computing resources. 
RTAB-Map's flexibility is demonstrated in this paper by making meaningful comparisons between visual and lidar-based SLAM configurations, allowing to analyze which robot sensor configuration is best for indoor autonomous navigation. 
RTAB-Map is distributed as an open-source library and is already available to the community. 
RTAB-Map is currently one of the top ROS packages actively used (over 1600 questions across its forum\footnote{\url{http://official-rtab-map-forum.67519.x6.nabble.com/}}, github repositories\footnote{\url{https://github.com/introlab/rtabmap}, \url{https://github.com/introlab/rtabmap_ros}} and ROS Answers\footnote{\url{https://answers.ros.org}}) by the community, for low-cost SLAM with RGB-D and stereo cameras.
Our goal with RTAB-Map is to continue integrating new odometry approaches lacking proper ROS integration, to facilitate comparison of SLAM configurations for autonomous navigation of mobile robot platforms.